\documentclass{article}

\PassOptionsToPackage{sort}{natbib}
\usepackage[accepted]{mlsys2021}

\usepackage[utf8]{inputenc} % allow utf-8 input
\usepackage[T1]{fontenc}    % use 8-bit T1 fonts
\usepackage{url}            % simple URL typesetting

\usepackage{algorithm}
\renewcommand{\algorithmiccomment}[1]{ $\quad\triangleright$ #1}

\usepackage[export]{adjustbox}
\usepackage{subcaption}
\usepackage[percent]{overpic}
\usepackage{wrapfig}
\usepackage{graphicx}
\usepackage{svg}

\usepackage{pifont}% http://ctan.org/pkg/pifont
\usepackage{float}

\usepackage{amsmath,amsfonts,amsthm,amssymb, nccmath}
\usepackage{nicefrac}       % compact symbols for 1/2, etc.
\allowdisplaybreaks % allow page breaks of equations
\usepackage{bbm}
\usepackage{framed}
\usepackage{enumerate}
\usepackage{mathtools}
\usepackage{scalefnt}
\usepackage{changepage}

\usepackage{comment}
\usepackage{setspace}
\usepackage{verbatim}
\usepackage{xcolor}
\usepackage{scalefnt}
\usepackage{xr}

\usepackage[shortlabels]{enumitem}
\setlist{nolistsep, noitemsep, leftmargin=16pt, labelwidth=3pt}

\usepackage{placeins}
\usepackage{color}
\usepackage{footmisc}
\usepackage{xspace}

\usepackage{thmtools}
\usepackage{thm-restate}

\usepackage{multirow}
\usepackage{booktabs}
\usepackage{array}
\usepackage{tabularx}

\usepackage{microtype}

\usepackage{hyperref}

\newcommand{\paragraphsmall}[1]{\textbf{{#1} $\;$}}

\makeatletter
\newcommand{\settitle}{\@maketitle}
\makeatother

\DeclarePairedDelimiterX{\dotp}[2]{\langle}{\rangle}{#1, #2}

\DeclareMathOperator*{\argmin}{\arg\!\min}
\DeclareMathOperator*{\argmax}{\arg\!\max}

\makeatletter
\newcommand*\bigcdot{\mathpalette\bigcdot@{.5}}
\newcommand*\bigcdot@[2]{\mathbin{\vcenter{\hbox{\scalebox{#2}{$\m@th#1\bullet$}}}}}
\makeatother

\newcommand{\be}{\begin{equation}}
\newcommand{\ee}{\end{equation}}

\newcommand{\norm}[1]{\left\| #1\right\|}                               %

\newcommand{\abs}[1]        {\left| #1 \right|}

\newcommand{\eps}{\ensuremath{\varepsilon}}                       % Epsilon
\renewcommand{\epsilon}{\varepsilon}

\declaretheorem[name=Definition]{definition}

\newcommand{\REAL}{\ensuremath{\mathbb{R}}}
\newcommand{\Reals}{\REAL}

\DeclareMathOperator*{\E}{\mathbb{E} \,}

\newcommand\PP{\mathcal{P}}
\newcommand\DD{\mathcal{D}}

\renewcommand\SS{\mathcal{S}}

\newcommand\UU{\mathcal{U}}

\newcommand{\xmark}{\ding{55}}%

\newcommand*{\ood}{o.o.d.\@\xspace}

\newcommand{\includenoisegraphics}[2]{ \adjincludegraphics[width=#1\textwidth, trim={0 0 {0.26\width} 0}, clip]{{#2}}%
}

\newcommand{\includenoiselegend}[2][0.29]{ \adjincludegraphics[width=\textwidth, trim={{0.76\width} {#1\height} 0 {0.07\height}}, clip]{fig/#2_WT_noise_matching}%
}

\newcommand{\includenoisefigure}[5][]{
\begin{figure*}[htb]
\centering
\begin{minipage}[c]{0.82\textwidth}
    \begin{minipage}[t]{0.49\textwidth}
        \includenoisegraphics{1.0}{fig/#2_#3_noise_matching}
        \subcaption{#3, Labels}
    \end{minipage}%
    \hfill
    \begin{minipage}[t]{0.49\textwidth}
        \includenoisegraphics{1.0}{fig/#2_#3_noise_diff}
        \subcaption{#3, Difference}
    \end{minipage}%
    \vspace{1ex}
    \begin{minipage}[t]{0.49\textwidth}
        \includenoisegraphics{1.0}{fig/#2_#4_noise_matching}
        \subcaption{#4, Labels}
    \end{minipage}%
    \hfill
    \begin{minipage}[t]{0.49\textwidth}
        \includenoisegraphics{1.0}{fig/#2_#4_noise_diff}
        \subcaption{#4, Difference}
    \end{minipage}
\end{minipage}%
\hfill
\begin{minipage}[c]{0.16\textwidth}
    \includenoiselegend{#2}
    \hfill
\end{minipage}
\caption{The functional similarities between pruned \textbf{{#5}} models and their unpruned parent. {#1}}
\label{fig:#2_#3_#4_noise}
\end{figure*}%
}

\newcommand{\includeprunecurvegraphics}[2]{ \adjincludegraphics[width=#1\textwidth, trim={0 0 {0.23\width} 0}, clip]{fig/#2_Jpeg_Speckle_Gauss_generalization_prune_pot}%
}

\newcommand{\includeprunecurvelegend}[2][0.36]{ \adjincludegraphics[width=\textwidth, trim={{0.77\width} {#1\height} 0 {0.26\height}}, clip]{fig/#2_Jpeg_Speckle_Gauss_generalization_prune_pot}%
}

\newcommand{\includeppgraphics}[2]{ \adjincludegraphics[width=#1\textwidth, trim={0 0 {0.21\width} 0}, clip]{fig/#2_generalization_prune_pot}%
}

\newcommand{\includepplegend}[2][0.48]{ \adjincludegraphics[width=\textwidth, trim={{0.79\width} {#1\height} 0 {0.29\height}}, clip]{fig/#2_generalization_prune_pot}%
}

\newcommand{\includeppfulllegend}[1][0.48]{
\adjincludegraphics[width=\textwidth, trim={{0.79\width} {0.49\height} 0 {0.29\height}}, clip]{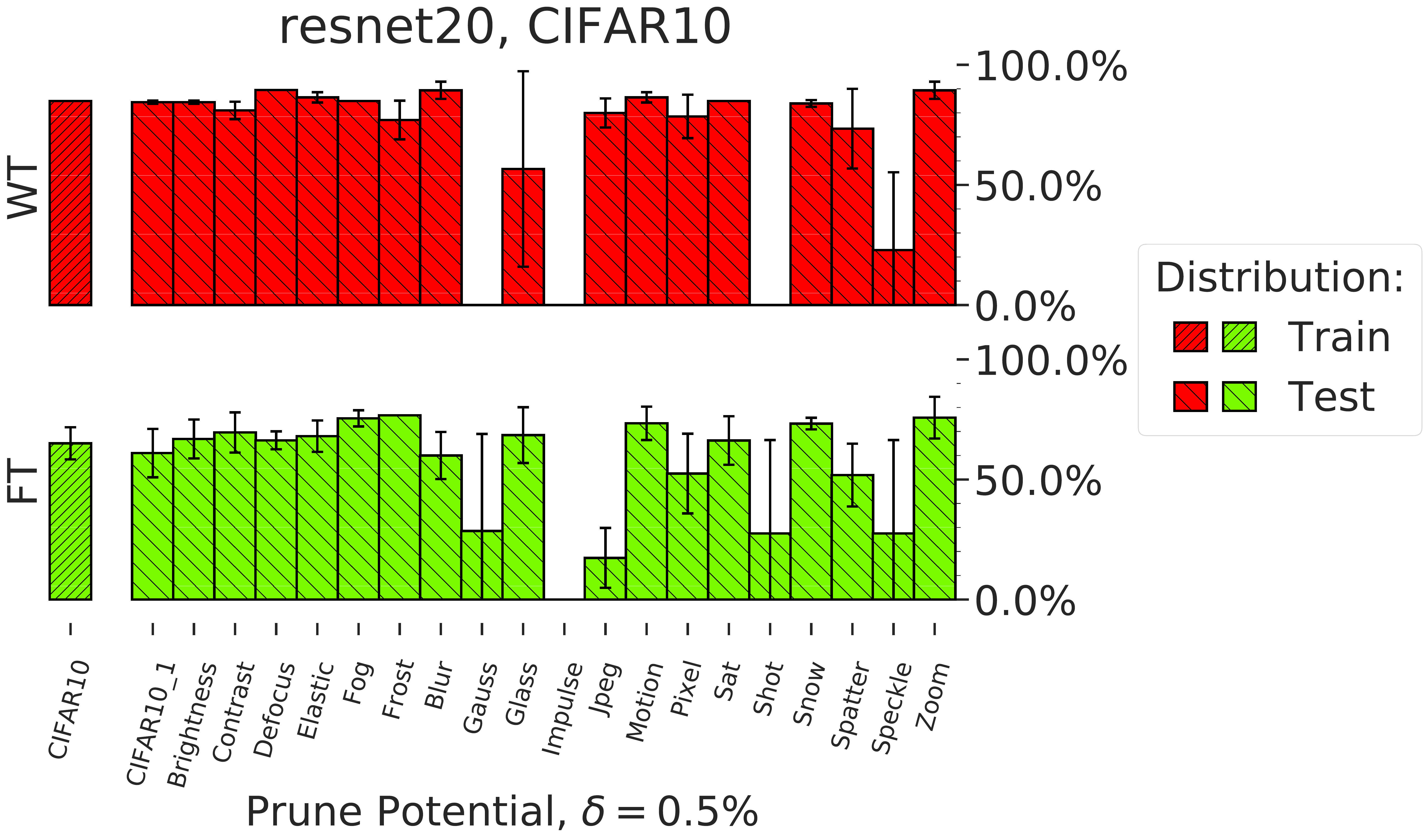}
\adjincludegraphics[width=\textwidth, trim={{0.79\width} {#1\height} 0 {0.35\height}}, clip]{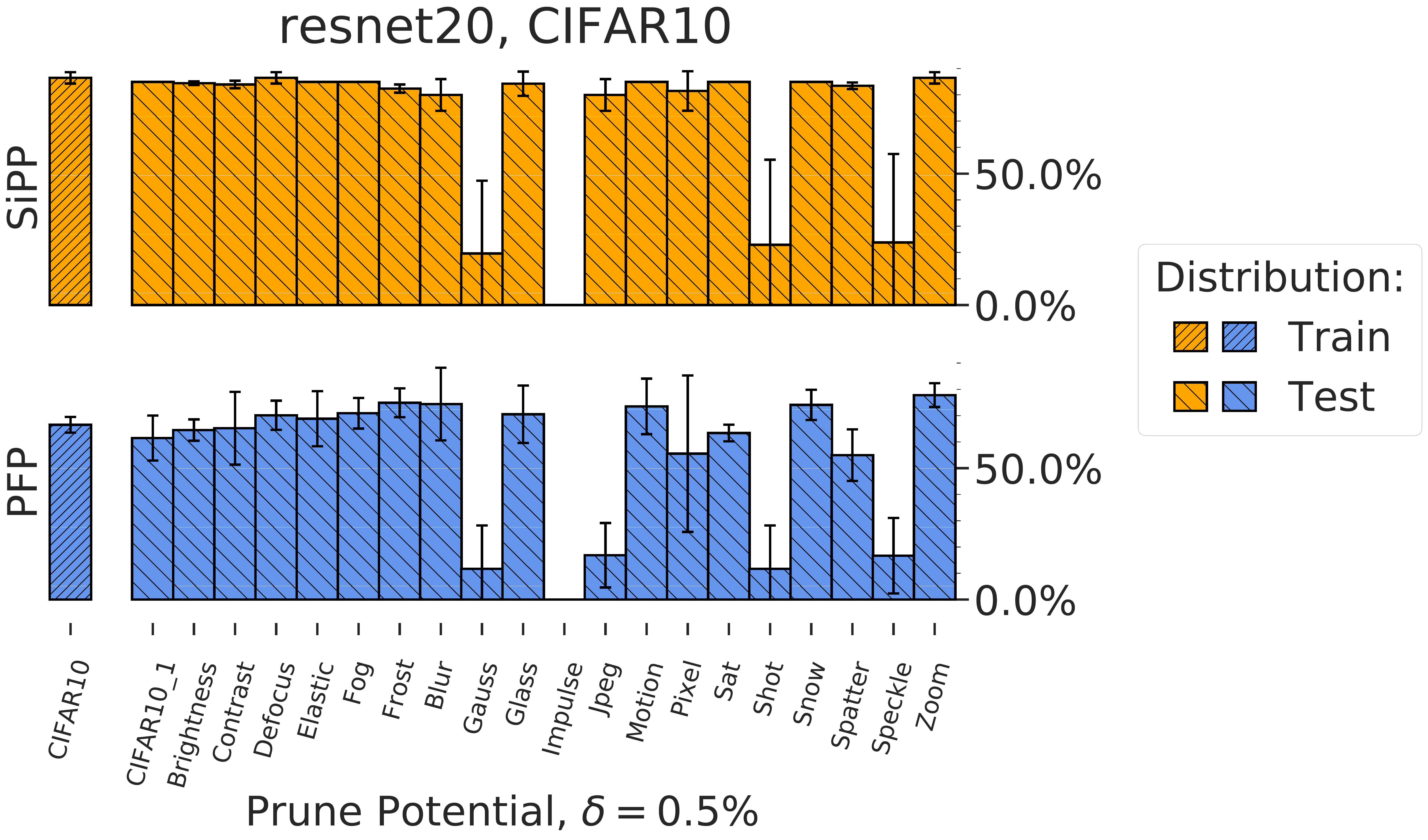}%
}

\newcommand{\includeppfigure}[4]{
\begin{figure*}[htb]
\centering
\begin{minipage}[t]{0.4\textwidth}\vspace{0pt}%
    \includeppgraphics{1.0}{#1_WT_SiPP}
    \subcaption{WT, SiPP}
\end{minipage}%
\begin{minipage}[t]{0.4\textwidth}\vspace{0pt}%
    \includeppgraphics{1.0}{#1_FT_PFP}
    \subcaption{FT, PFP}
\end{minipage}%
\begin{minipage}[t]{0.125\textwidth}\vspace{0pt}%
    \vspace{0.5\textwidth}
    \includeppfulllegend
\end{minipage}
\begin{minipage}[t]{0.4\textwidth}
    \includeprunecurvegraphics{1.0}{WT_#1_#2}
    \subcaption{WT, prune-test curve}
\end{minipage}%
\begin{minipage}[t]{0.4\textwidth}
    \includeprunecurvegraphics{1.0}{FT_#1_#2}
    \subcaption{FT, prune-test curve}
\end{minipage}%
\begin{minipage}[t]{0.125\textwidth}
    \includeprunecurvelegend[0.0]{WT_#1_#2}
\end{minipage}
\caption{The prune potential of a \textbf{{#3}} achievable for \textbf{{#4}} out-of-distribution data sets.}
\label{fig:#1_generalization_prune_pot}
\end{figure*}%
}

\newcommand{\includecifarppfigure}[2]{
\includeppfigure{#1_CIFAR10}{CIFAR10}{#2}{CIFAR10}
}

\newcommand{\includeimagenetppfigure}[2]{
\includeppfigure{#1_ImageNet}{ImageNet}{#2}{ImageNet}
}

\newcommand{\includevocppfigure}[2]{
\includeppfigure{#1_VOCSegmentation2011}{VOCSegmentation2011}{#2}{Pascal VOC}
}

\newcommand{\includecifarmixppfigure}[2]{
\includeppfigure{#1_CIFAR10_C_Mix1}{CIFAR10}{#2}{CIFAR10}
}

\newcommand{\includeexcessgraphics}[2]{ \adjincludegraphics[width=#1\textwidth, trim={0 0 {0.171\width} 0}, clip]{fig/#2_generalization_excess_err}%
}

\newcommand{\includeexcesslegend}[2][0.54]{ \adjincludegraphics[width=\textwidth, trim={{0.82\width} {#1\height} 0 {0.26\height}}, clip]{fig/#2_generalization_excess_err}%
}

\newcommand{\includeexcessfulllegend}[1][0.365]{
\adjincludegraphics[width=\textwidth, trim={{0.82\width} {#1\height} 0 {0.26\height}}, clip]{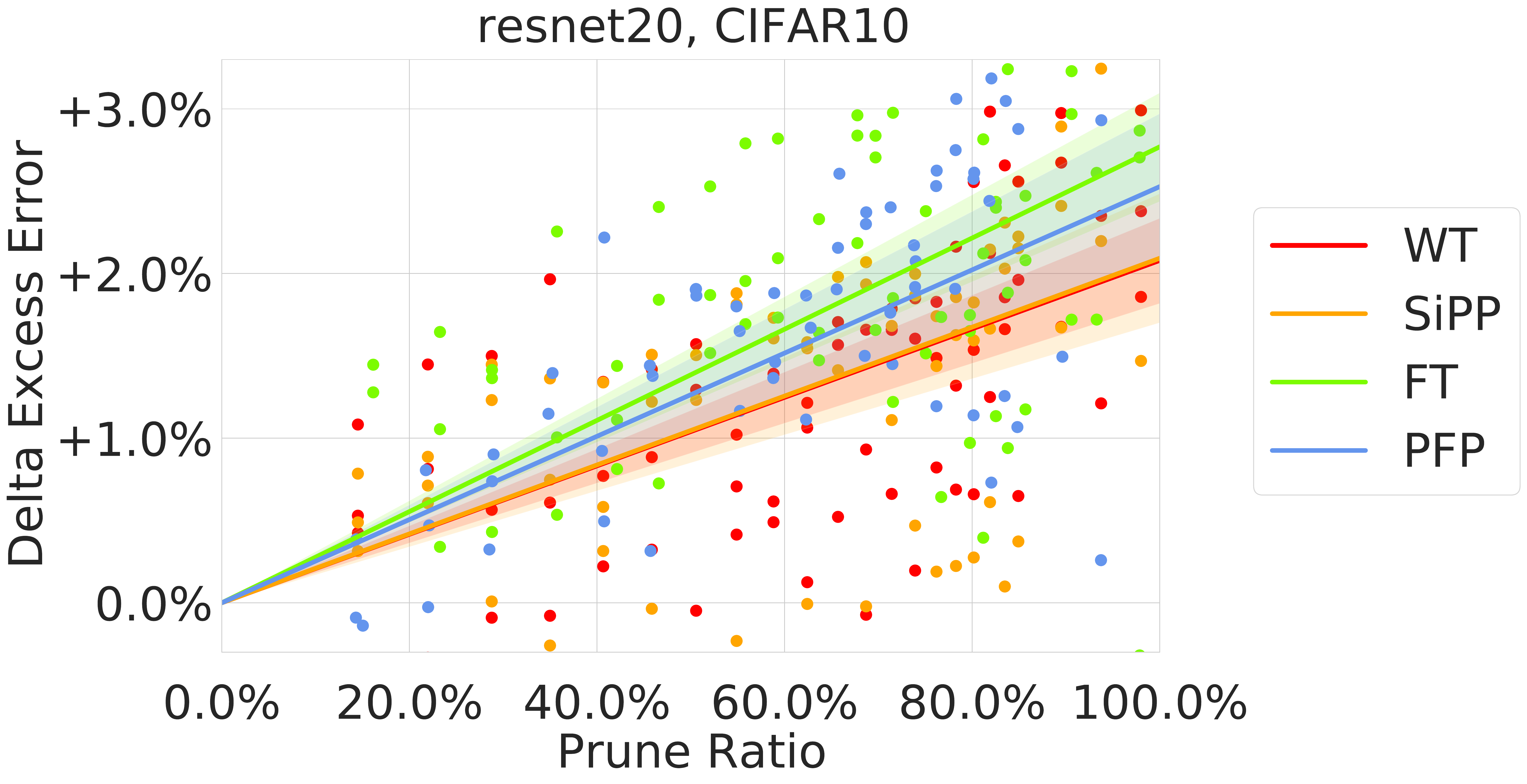}%
}

\newcommand{\includeexcessfigure}[3]{

\begin{figure*}[htb]
    \centering
    \begin{minipage}[t]{0.44\textwidth}
        \includeexcessgraphics{1.0}{#1_WT_SiPP}
    \end{minipage}%
    \begin{minipage}[t]{0.44\textwidth}
        \includeexcessgraphics{1.0}{#1_FT_PFP}
    \end{minipage}%
    \begin{minipage}[t]{0.11\textwidth}
        \includeexcessfulllegend[0.05]
    \end{minipage}
    \caption{The difference in excess error for a \textbf{#2} trained on \textbf{#3}.}
    \label{fig:#1_excess_error}
\end{figure*}%
}

\newcommand{\includecifarexcessfigure}[2]{
\includeexcessfigure{#1_CIFAR10}{#2}{CIFAR10}
}

\newcommand{\includeimagenetexcessfigure}[2]{
\includeexcessfigure{#1_ImageNet}{#2}{ImageNet}
}

\newcommand{\includevocexcessfigure}[2]{
\includeexcessfigure{#1_VOCSegmentation2011}{#2}{Pascal VOC}
}

\newcommand{\includecifarmixexcessfigure}[2]{
\includeexcessfigure{#1_CIFAR10_C_Mix1}{#2}{CIFAR10}
}

\renewcommand{\paragraph}{\paragraphsmall}
\mlsystitlerunning{Lost in Pruning: The Effects of Pruning Neural Networks beyond Test Accuracy}

\begin{document}
%%%%%%%%%%%%%%%%%%%%%%%%%%%%%%%%%%%%%%%%
%%%%%%% TITLE, ABSTRACT, AUTHORS %%%%%%%
%%%%%%%%%%%%%%%%%%%%%%%%%%%%%%%%%%%%%%%%
\twocolumn[
\mlsystitle{Lost in Pruning: The Effects of Pruning Neural Networks beyond Test Accuracy}

\mlsyssetsymbol{equal}{*}
\begin{mlsysauthorlist}
\mlsysauthor{Lucas Liebenwein}{csail}
\mlsysauthor{Cenk Baykal}{csail}
\mlsysauthor{Brandon Carter}{csail}
\mlsysauthor{David Gifford}{csail}
\mlsysauthor{Daniela Rus}{csail}
\end{mlsysauthorlist}

\mlsysaffiliation{csail}{Computer Science and Artificial Intelligence Lab, Massachusetts Institute of Technology, Cambridge, MA, USA}

\mlsyscorrespondingauthor{Lucas Liebenwein}{lucasl@mit.edu}

\mlsyskeywords{neural networks, sparsity, pruning, compression, performance, robustness}

% one-sentence summary
% The performance of pruned neural networks is disproportionally more affected by out-of-distribution data compared to their unpruned counterpart.

\vskip 0.3in
\begin{abstract}
%%%%%%%%%%%%%%%%%%%%%%%%%%%%%%%%%%%%%%%%%%%%
%%% DON'T CHANGE BEFORE FINAL SUBMISSION %%%
%%%%%%%%%%%%%%%%%%%%%%%%%%%%%%%%%%%%%%%%%%%%
Neural network pruning is a popular technique used to reduce the inference costs of modern, potentially overparameterized, networks. Starting from a pre-trained network, the  process is as follows: remove redundant parameters, retrain, and repeat while maintaining the same test accuracy. The result is a model that is a fraction of the size of the original with comparable predictive performance (test accuracy). Here, we reassess and evaluate whether the use of test accuracy alone in the terminating condition is sufficient to ensure that the resulting model performs well across a wide spectrum of "harder" metrics such as generalization to out-of-distribution data and resilience to noise. Across evaluations on varying architectures and data sets, we find that pruned networks effectively approximate the unpruned model, however, the prune ratio at which pruned networks achieve commensurate performance varies significantly across tasks. These results call into question the extent of \emph{genuine} overparameterization in deep learning and raise concerns about the practicability of deploying pruned networks, specifically in the context of safety-critical systems, unless they are widely evaluated beyond test accuracy to reliably predict their performance.
Our code is available at \url{https://github.com/lucaslie/torchprune}.
%%%%%%%%%%%%%%%%%%%%%%%%%%%%%%%%%%%%%%%%%%%%
%%%%%%%%%%%%%%%%%%%%%%%%%%%%%%%%%%%%%%%%%%%%
%%%%%%%%%%%%%%%%%%%%%%%%%%%%%%%%%%%%%%%%%%%%
\end{abstract}

]

\printAffiliationsAndNotice{}
% \printAffiliationsAndNotice{\mlsysEqualContribution}

%%%%%%%%%%%%%%%%%%%%%%%%%%%%%%%%%%%%%%%%
%%%%%%%%%%%%%% MAIN PAPER %%%%%%%%%%%%%%
%%%%%%%%%%%%%%%%%%%%%%%%%%%%%%%%%%%%%%%%

% sections
\section{Introduction}
\label{sec:introduction}

% The importance of NN pruning 
Deep neural networks~\cite{ILSVRC15, you2019large} tend to contain millions or billions of parameters, necessitating costly computational resources in order to train and deploy them in practice, and motivating the need to develop compact networks with fewer parameters~\cite{liebenwein2020provable, Han15}. Such a reduction in parameter count alleviates the computational burden on training and inference, making it easier to deploy high-capacity models to small devices and use them in resource-constrained environments~\cite{frankle2018lottery,  baykal2018datadependent, renda2020comparing}.

To this end, network pruning is a popular technique used to obtain compact networks that maintain the accuracy of the original network with orders of magnitude of fewer parameters~\cite{gale2019state, blalock2020state}. Typical pruning algorithms either proceed by gradually pruning the network during training~\cite{zhu2017prune, gale2019state, he2018soft} or by pruning the network after training followed by a retraining period~\cite{Han15, sipp2019, renda2020comparing, liebenwein2020provable}.

A prune pipeline with retraining usually consists of the following steps:
\begin{enumerate}
    \item 
    Prune weights (``unstructured pruning'') or filters/neurons (``structured pruning'') from the trained network according to some criterion of importance;
    \item
    Retrain the resulting network to regain the full accuracy;
    \item
    Iteratively repeat steps 1 \& 2 to further reduce the size.
\end{enumerate}

These approaches are both simple and network-agnostic, and have been shown to yield state-of-the-art pruning results for both unstructured~\cite{renda2020comparing} and structured pruning~\cite{liebenwein2020provable}. For example, these techniques enable pruning of 89\% of the weights~\cite{renda2020comparing} or 84\% of the filters~\cite{liebenwein2020provable} of a ResNet56 trained on CIFAR10.

While the ability to prune large portions of a network is not obvious at first sight, common wisdom attributes the apparent overparameterization~\cite{arora2018stronger, baykal2018datadependent, frankle2018lottery, du2018gradient} of modern deep learning architectures as one of the key reasons why pruning such large portions of the network is possible without harming the performance of the network.

In other words, successfully pruning a network entails identifying and removing the redundant parameters. Moreover, starting from a pre-trained, overparameterized network with good performance (as opposed to a small, randomly initialized network) ensures that the pruned network maintains the same level of performance as its uncompressed counterpart.

In this paper, we revisit these common assumptions and rigorously assess how pruning using state-of-the-art prune-retrain techniques~\cite{renda2020comparing, sipp2019, liebenwein2020provable} affects the function represented by a neural network, including the similarities and disparities exhibited by a pruned network with respect to its unpruned counterpart. 

We formalize the notion of (functional) similarities between networks by introducing novel types of classification-based functional distance metrics. Using these metrics, we test the hypothesis that pruned models are (functionally) similar to their (unpruned) parent network and can be reliably distinguished from separately trained networks.
We term the network's ability to be pruned for a particular task without performance decrease its \emph{prune potential}, i.e., the maximal prune ratio for which the pruned network maintains its original performance, and test a network's prune potential under various tasks. The prune potential provides insights into the amount of overparameterization the network exhibits for a particular task and thus serves as a useful indicator of how much of the network can be safely pruned.

\paragraph{Our findings.}
We find that the pruned models are functionally similar to the uncompressed parent model, which enables us to distinguish the parent of a pruned network for a range of prune ratios.
Despite the similarity between the pruned network and its parent, we observe that the prune potential of the network varies significantly for a large number of tasks. 
That is, a pruned model may be of similar predictive power as the original one when it comes to test accuracy, but may be much more brittle when faced with out of distribution data points.   
This raises concerns about deploying pruned models on the basis of accuracy alone, in particular for safety-critical applications such as autonomous driving~\cite{schwarting2020deep}, where unforeseen, out-of-distribution, or noisy data points commonly arise. Our insights, which hold even when considering robust training objectives, underscore the need to consider task-specific evaluation metrics during pruning, prior to the deployment of a pruned network to, e.g., safety-critical systems. 
These results also question the common assumption that there exists a significant amount of ``redundant'' parameters to begin with and provide a robust framework to measure the amount of genuine overparameterization in networks.

\begin{figure}[t!]
\begin{center}
\includegraphics[width=0.8\columnwidth]{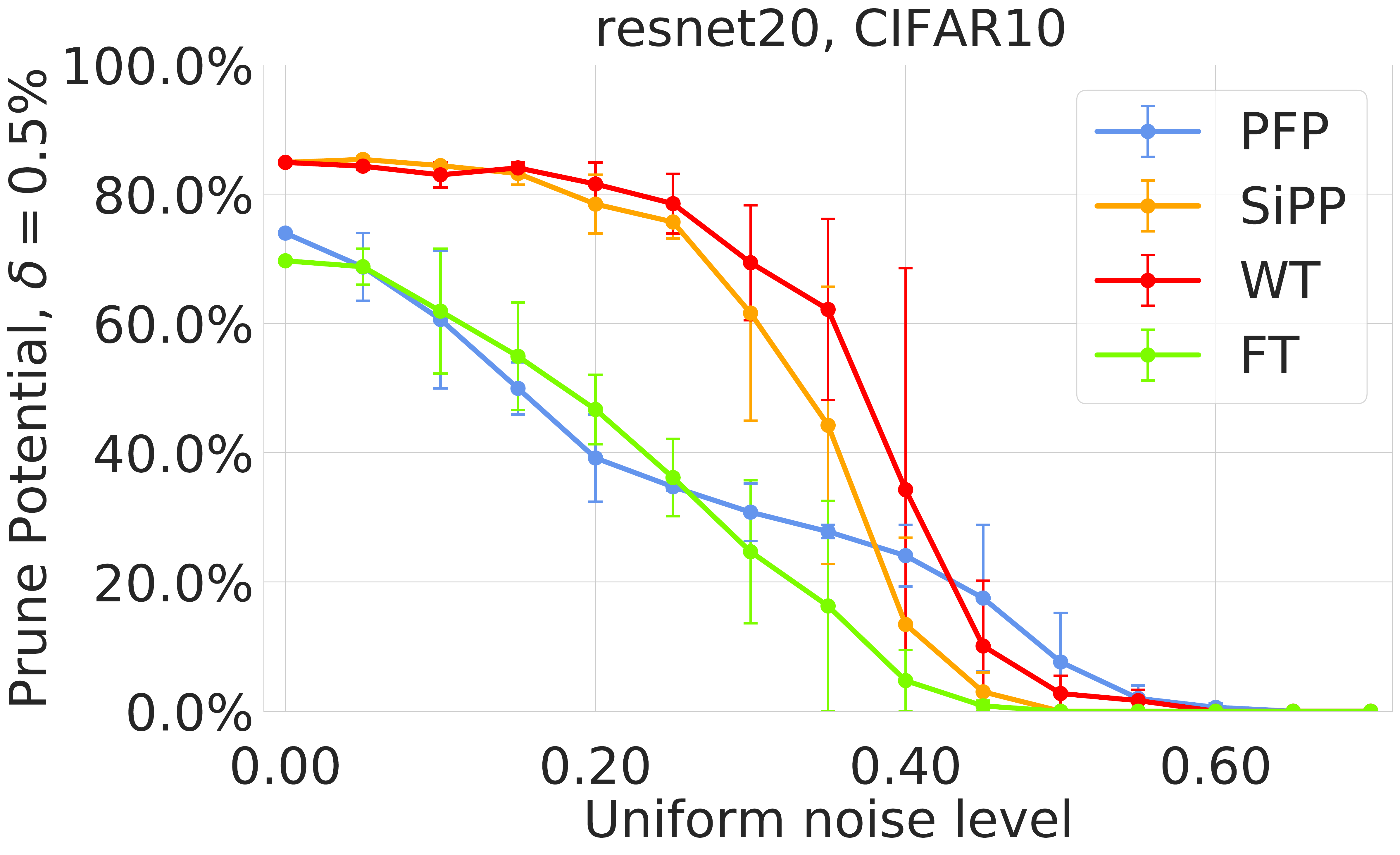}
\vspace{-1ex}
\caption{A network's ability to be pruned without loss of accuracy, i.e., its ``prune potential'', can be significantly affected by small changes in the input data distribution.}
\label{fig:prune_potential_noise}
\vspace{-4ex}
\end{center}
\end{figure}

\paragraph{Guidelines.}
Based on these observations we formulate a set of easy-to-follow guidelines to pruning in practice: 
\begin{enumerate}
    \item 
    Don't prune if unexpected shifts in the data distribution may occur during deployment.
    \item
    Prune moderately if you have partial knowledge of the distribution shifts during training and pruning.
    \item
    Prune to the full extent if you can account for all shifts in the data distribution during training and pruning.
    \item
    Maximize the prune potential by explicitly considering data augmentation during retraining.
    
\end{enumerate}

\paragraph{Contributions.}
\begin{itemize}
    \item 
    We propose novel functional distance metrics for classification-based neural networks and investigate the functional similarities between the pruned network and its unpruned counterpart.
    \item
    We propose the notion of \emph{prune potential}, i.e., the maximal prune ratio (model sparsity) at which the pruned network can achieve commensurate performance, as a quantifiable means to estimate the overparameterization of a network and show that it is significantly lower on challenging inference tasks.
    \item
    We provide a unified framework to establish task-specific guidelines that help practitioners assess the effects of pruning during the design and deployment of neural networks in practice.
    \item
    We conduct experiments across multiple data sets, architectures, and pruning methods showing that our observations hold across common pruning benchmarks and real-world scenarios.
\end{itemize}
\section{Related Work}
\paragraph{Pruning.}
Generally, pruning is categorized into unstructured~\cite{Han15} and structured~\cite{he2018soft, li2019learning} pruning approaches. While the former is beneficial for research, it provides little computational speed-ups compared to structured pruning~\cite{luo2018autopruner}. Moreover, pruning can be performed before~\cite{lee2018snip, tanaka2020pruning, Wang2020Picking}, during~\cite{zhu2017prune,yu2018slimmable, kusupati2020soft}, or after training~\cite{singh2020woodfisher,sipp2019, liebenwein2020provable}, and repeated iteratively~\cite{renda2020comparing}. 
In this work, we focus on iterative pruning with retraining after training. While computationally more expensive than other approaches, this pruning pipeline produces state-of-the-art results, is usually architecture-agnostic, and requires little hyperparameter tuning, thus providing a simple and effective baseline for our experiments.
A thorough overview of recent pruning approaches is, e.g., provided by \citet{blalock2020state, gale2019state, liu2018rethinking}.

\paragraph{Robustness.}
Our work builds on and extends previous work that investigates the robustness of pruned networks. The works of~\citet{zhao2018compress, ye2019adversarial, gamboa2020campfire, NIPS2018_7308, wang2018adversarial} have investigated the effect of adversarial input on pruned networks, however, the resulting evidence is inconclusive. While~\citet{NIPS2018_7308, gamboa2020campfire} report that adversarial robustness may improve or remain the same for pruned networks, \citet{zhao2018compress, ye2019adversarial, wang2018adversarial} report decreased robustness for pruned networks. Recently, the work of~\citet{hooker2019selective} investigated at the individual image level whether certain class accuracies are more affected than others.
In contrast to prior work, we investigate both the functional similarities of pruned networks and the task-specific prune potential. Our work highlights the need to assess pruned networks across a wide variety of tasks to safely deploy networks due to the unpredictable nature of a network's prune potential.
% , i.e., the prune potential may improve for some tasks while significantly decrease for others.

\paragraph{Robust training and pruning.}
Recent works~\cite{dhillon2018stochastic, wijayanto2019towards} have investigated pipelines that incorporate pruning and robust training to obtain simultaneously sparse and robust networks. \citet{gui2019model, sehwag2019towards} use magnitude-based pruning~\cite{Han15} to train sparse, robust networks, while the work of~\citet{sehwag2020hydra} incorporates the robustness objective into an optimization-based pruning procedure.
Our work offers a complementary viewpoint to the findings of prior work in that we can indeed efficiently train pruned networks that are (adversarially) robust. However, for the first time we make the crucial observation that pruned networks are disproportionally more affected by distributional changes in the input regardless of the training procedure.

\paragraph{Generalization via pruning.}
Among others \citet{zhou2018compressibility, baykal2018datadependent, arora2018stronger, nagarajan2018deterministic} have demonstrated that (provable) pruning can facilitate tighter generalization bounds of networks.
Pruning is hereby viewed as a form of noise injection into the network and by quantifying the (unpruned) network's ability to withstand random noise, they characterize the prunability of the network to establish generalization bounds. 
However, these works do not capture the generalization ability of pruned networks under distributional changes. Here, we show that as a result of pruning, the network's ability to withstand noise and other types of data corruption is diminished. In other words, the network's robustness is ``traded'' in exchange for compactness. 

\paragraph{Implicit regularization via overparameterization.}
Our work is also related to the beneficial role of overparameterization in deep learning~\cite{ allen2019learning, zhang2016understanding}.
Conventional wisdom~\cite{NeyshaburTS14, neyshabur2018towards,du2018gradient, neyshabur2018the} states that stochastic gradient methods used for training implicitly regularize the network, in effect ensuring it generalizes well despite the potential to severely overfit. Moreover, the works of~\citet{belkin2019reconciling, Nakkiran2020Deep} note that the implicit regularization potential increases with the parameter count. Our work, for the first time, thoroughly establishes that pruned networks distinctly suffer more from small shifts in the input data distribution compared to unpruned networks -- possibly due to the decreased \emph{implicit} regularization potential as a result of the lower parameter count. Our findings concretely highlight that \emph{explicit} regularization in the form of robust training can help regain some of the robustness properties that would otherwise be lost.
\section{Methodology}
\label{sec:methods}

\begin{table*}[t!]
\centering
\small
\begin{tabular}{c|rl|c|l|c}
    Type
    & \multicolumn{2}{c|}{Method}
    & Data-informed & \multicolumn{1}{c|}{Sensitivity} & Scope \\ \hline
    Unstructured
    & WT: & Weight Thresholding~\cite{renda2020comparing}
    & \xmark & $ \hphantom{\propto}$ $\abs{W_{ij}}$ 
    & Global \\
    (Weights)
    & SiPP: & Sensitivity-informed Pruning~\cite{sipp2019}
    & $\checkmark$ & $\propto$ $\abs{W_{ij} a_j(x)}$ 
    & Global\\
    \hline
    Structured
    & FT: & Filter Thresholding~\cite{renda2020comparing} 
    & \xmark & $ \hphantom{\propto}$ $\norm{W_{:j}}_1$ 
    & Local \\
    (Neurons/Filters)
    & PFP: & Provable Filter Pruning~\cite{liebenwein2020provable} 
    & $\checkmark$ & $\propto$ $\norm{W_{: j} a(x)}_\infty$ 
    & Local \\
\end{tabular}
\caption{Overview of the pruning methods evaluated. Here, $a(x)$ denotes the activation of the corresponding layer with respect to a sample input $x$ to the network.}
\label{tab:prune-algs}
\vspace{-3ex}
\end{table*}    

\subsection{Pruning Setup}
For our experiments, we consider a variety of network architectures, data sets, and pruning methods as outlined below. Our pruning pipeline, see Algorithm~\ref{alg:prune}, is based on iterative pruning and retraining following~\citet{renda2020comparing}. It is simple, network-agnostic, and widely used; hence we opted to choose it as representative pruning pipeline.

\paragraph{Data sets and network architectures.}
We consider CIFAR10~\cite{torralba200880}, ImageNet~\cite{ILSVRC15}, and Pascal VOC segmentation data~\cite{everingham2015pascal} as data sets. We consider ResNets 18/56/110~\cite{he2016deep}, WRN16-8~\cite{zagoruyko2016wide}, DenseNet22~\cite{huang2017densely}, and VGG16~\cite{Simonyan14} on CIFAR10; ResNet18 and 101~\cite{he2016deep} on ImageNet; and a DeeplabV3-ResNet50~\cite{chen2017rethinking} on VOC. 

\paragraph{Training.}
For all networks, we apply the standard training parameters as indicated in the respective papers. We apply the linear scaling rule of~\citet{goyal2017accurate} when training on multiple GPUs in parallel including warm-up. All hyperparameter settings with their numerical values are listed in Appendix~\ref{sec:supp-results}. All networks are trained once to completion before pruning (Line 2 of Algorithm~\ref{alg:prune}).

\paragraph{Pruning.}
We consider multiple unstructured and structured pruning methods, where we prune individual weights and filters/neurons, respectively, see Table~\ref{tab:prune-algs} for an overview.
We perform pruning by updating a binary mask indicating whether the corresponding weight is active or pruned (Line~5 of Algorithm~\ref{alg:prune}).

\paragraph{Unstructured pruning.}
The weight pruning approaches we consider follow a global pruning strategy: (1) globally sort the weights according to their relative importance, i.e., sensitivity, and (2) prune $r_\text{prune}\%$ of the weights with the lowest sensitivity.
In particular, we study two methods to compute the sensitivity of weights, weight thresholding~\cite{renda2020comparing} and SiPP~\cite{sipp2019}. Weight thresholding (WT) is a simple heuristic, originally introduced by~\citet{Han15} and re-purposed by~\citet{renda2020comparing}, that defines the sensitivity of a weight as the magnitude of the weight. SiPP, on the other hand, is a data-informed approach with provable guarantees to computing weight sensitivities~\cite{sipp2019}. The approach uses a small batch of input points $\SS \subseteq \PP$, e.g., from the validation set, to evaluate the saliency of each network parameter. This is done by incorporating the corresponding (sample) activations, $a(x), \, x \in \SS$, along with the weight into the importance computation (see Table~\ref{tab:prune-algs}).

\begin{algorithm}[htb]
\small
\caption{\textsc{PruneRetrain}($n_\text{cycles}$, $r_\text{prune}$, $n_\text{train}$, $\rho_\text{train}$)}
\label{alg:prune}
\textbf{Input:} $n_\text{cycles}$: number of prune-retrain cycles; $r_\text{prune}$: relative prune ratio; $n_\text{train}$: number of train epochs; $\rho_\text{train}$: training hyper-parameters \\
\textbf{Output:} $c$: pruning mask, $\theta$: parameters of the pruned network
\begin{spacing}{1.1}
\begin{algorithmic}[1]
\small
\STATE $\theta_0 \gets \textsc{RandomInit()}$ \label{lin:randominit} \\
\STATE $\theta \gets \textsc{Train}(\theta_0,  n_\text{train}, \rho_\text{train})$ \label{lin:train} \\
\STATE $c \gets 1^{|\theta_0|}$ \COMMENT{binary mask for the parameters} \\
\FOR{$i \in [n_\text{cycles}]$} \label{lin:prune-start}
    \STATE $c \gets \textsc{Prune}(c \odot \theta, r_\text{prune})$ \COMMENT{Prune $r_\text{prune}\%$ of the remaining parameters.} \label{lin:prune} \\
    \STATE $\theta \gets \textsc{Train}(c \odot \theta, n_\text{train}, \rho_\text{train})$ \label{lin:retrain}
\ENDFOR \label{lin:prune-end}
\STATE \textbf{return} $c, \theta$
\end{algorithmic}
\end{spacing}
\end{algorithm}

\paragraph{Structured pruning.}
The filter/neuron pruning approaches we consider follow a two-step strategy: (1) allocate a per-layer prune ratio satisfying the overall prune ratio and (2) prune the filters with lowest sensitivity in each layer.
We study Filter Tresholding (FT) as used by~\citet{renda2020comparing} and PFP~\cite{liebenwein2020provable}. FT, as originally introduced by~\citet{he2018soft, li2016pruning} and used here analogous to~\citet{renda2020comparing}, uses the filter norm to evaluate its sensitivity. Layer allocation is performed manually and we deploy a uniform prune ratio across layers to avoid further hyperparameters. PFP~\cite{liebenwein2020provable} is an extension of SiPP that evaluates filter sensitivity as the maximum sensitivity of the channel in the next layer ($\ell_\infty$-norm of the corresponding weight sensitivity). PFP uses the associated theoretical error guarantees to optimally allocate the layer-wise budget.

\paragraph{Retraining.}
We retrain the network with the exact training hyperparameters as is common~\cite{sipp2019, renda2020comparing, liebenwein2020provable}. Specifically, we re-use the same learning rate schedule and retrain for the same amount of epochs (Line~6 of Algorithm~\ref{alg:prune}). 
After retraining, we iteratively repeat the pruning procedure to obtain even smaller networks (Lines~4-7 of Algorithm~\ref{alg:prune}).

\begin{figure}[t!]
  \centering
    \includegraphics[width=0.75\columnwidth]{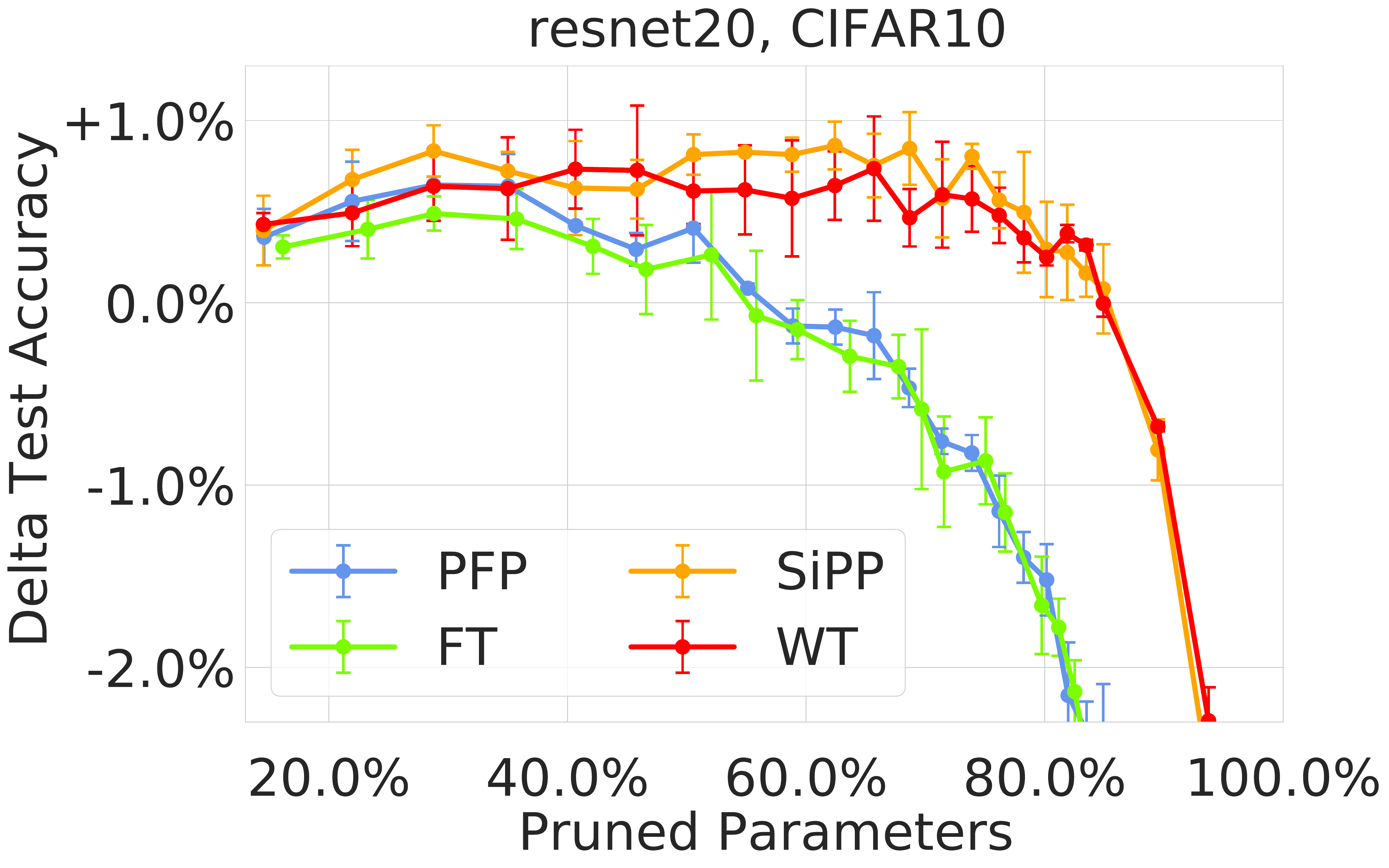}
    \caption{The accuracy of the generated pruned models for the evaluated pruning schemes for various target prune ratios using iterative fine-tuning.}
    \label{fig:cifar_resnet20_prune}
    \vspace{-3ex}
\end{figure}

\paragraph{Prune results.}
Figure~\ref{fig:cifar_resnet20_prune} shows an exemplary test accuracy curve of a Resnet20 (CIFAR10) across different target prune ratios for iterative pruning. The remaining prune results are summarized in the supplementary material. We note while \textsc{PruneRetrain} may be more computationally expensive than other pruning pipelines, it is network-agnostic and produces state-of-the-art pruning results~\cite{renda2020comparing}.

\subsection{Experiments Roadmap}
Our observations stem from multiple experiments that can be clustered into one set of experiments pertaining to understanding the functional similarities (i.e., functional distance) of pruned networks and one set pertaining to the prune potential on a variety of image-based classification tasks. 
First, we compare subsets of pixels that are sufficiently informative for driving the network's decision. For this, we use the feature-wise (pixel-wise) selection mechanism of~\citet{carter2020overinterpretation, carter2019made}.
% that identifies sparse subsets of input features on which the network still makes high-confidence predictions despite the missing information. 
We also investigate how pruned networks behave under $\ell_\infty$-bounded random noise.
Second, we assess the ability of pruned networks to generalize to out-of-distribution (\ood) test data sets that contain random noise ($\ell_\infty$-bounded) and common corruptions~\cite{recht2018cifar10.1, recht2019imagenet, hendrycks2019robustness} including weather, contrast, and brightness changes.
In all experiments, we compare the performance, i.e., the accuracy on the various test sets, of pruned networks with those of their unpruned counterparts as well as a separately trained unpruned network. Each experiment is repeated 3 times and we report mean and standard deviation (error bars). In the main part of the paper, we focus on a subset of representative results to highlight the key findings.
We refer the interested reader to the appendix for a complete exposition of our results.
\section{Function Distance}
\label{sec:distance}
Given a pruned model with commensurate test accuracy relative to the parent (uncompressed) network, can we conclude that the \emph{function} represented by the pruned model is similar to the parent network for unforeseen data points? In this section, we investigate the extent to which the pruned and parent model are functionally similar under two distinct metrics: informative features and noise resilience. Our findings show that pruned networks are more functionally similar to their original network than a separately trained, unpruned network underscoring the intuition that pruned networks remain functionally similar to their unpruned counterpart.

\subsection{Methodology}
\label{sec:distance-methods}
\paragraph{Comparison of informative features.}
We compare features (pixels) that are informative for the decision-making of each model. Specifically, for a network $f_\theta(x)$ with parameters $\theta$ and input $x \in \Reals^n$ we want to find an input mask $m \in \{0, 1\}^n$ such that $f_\theta(x) \approx f_\theta(m \odot x)$, i.e.,
\begin{equation}
\label{eq:informative-mask}
m = \argmin_{\norm{m}_0 \leq (1-B)n} \norm{f_\theta(x) - f_\theta(m \odot x)}
\end{equation}
for some sparsity level $B$.
To approximately solve equation~\ref{eq:informative-mask} we use the greedy backward selection algorithm (\verb|BackSelect|) of~\citet{carter2019made, carter2020overinterpretation}. The procedure iteratively masks the least informative pixel (i.e., the pixel which if masked would reduce the confidence of the prediction of the correct label by the smallest amount) to obtain a sorting of the pixels in order of increasing importance. After sorting, we can remove the bottom $B\%$ of pixels.

Given two networks $f_{\theta}(\cdot)$ and $f_{\hat\theta}(\cdot)$, we can then measure the difference between the functions by switching up the respective input masks $m$ and $\hat m$ to see the change in the output, i.e., $\norm{f_{\theta}(\hat m \odot x) - f_{\theta}(m \odot x)}$ and vice versa. If one model can make a confident and correct prediction on the pixels that were informative to another model, the models may have similar decision-making strategies. We apply this strategy to identify subsets of informative pixels across a sample of 2000 CIFAR-10 test images. For each image, we compute the subset of informative pixels, i.e., the input mask $m$, for an unpruned network, five pruned networks (of increasing prune ratio) derived from that network, and a separate, unpruned network of the same type. We probe whether the informative pixels from one model are also informative to the other models for a sparsity level of $90\%$.

\paragraph{Noise similarities.}
We consider injecting $l_\infty$-bounded, random noise into the test data and we compare the predicted labels between the pruned networks and their unpruned counterpart to investigate the behavior in local neighborhood of points. Specifically, for two networks $f_\theta(\cdot)$ and $f_{\hat\theta}(\cdot)$ with parameters $\theta$ and $\hat\theta$, respectively, we consider the expected number of matching label prediction and the expected norm difference of the output with noise $\eps$, i.e.,
\begin{equation*}
\textstyle
\E_{x' \sim \DD + \UU^n(-\eps, \eps)}\big[ \argmax{f_\theta(x')} = \argmax{f_{\hat\theta}(x')} \big]
\end{equation*}
and
\begin{equation*}
\textstyle
\E_{x' \sim \DD + \UU^n(-\eps, \eps)} \norm{f_\theta(x') - f_{\hat\theta}(x')}_2,
\end{equation*}
respectively. 
We test the noise similarity of networks for a random subset of 1000 test images for 100 repetitions of random noise injection and average over the results.

\begin{figure}[t!]
    \centering
    \includegraphics[width=0.93\columnwidth]{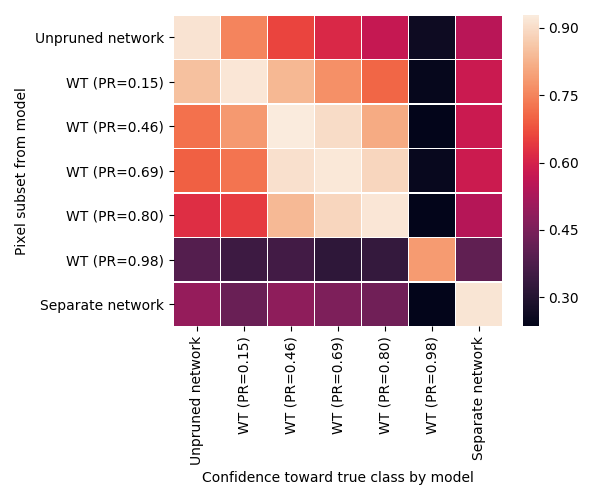}
    \caption{Confidence heatmaps on informative pixels from WT-pruned ResNet20s. Y-axis is the model used to generate the informative pixel subset toward the predicted class, x-axis describes the models evaluated with the subset, cells indicate mean confidence toward the true class.}
    \label{fig:sis-confidence-heatmap}
\end{figure}

\begin{figure}[t!]
\centering
\begin{minipage}[t]{0.42\columnwidth}
    \includenoisegraphics{1.0}{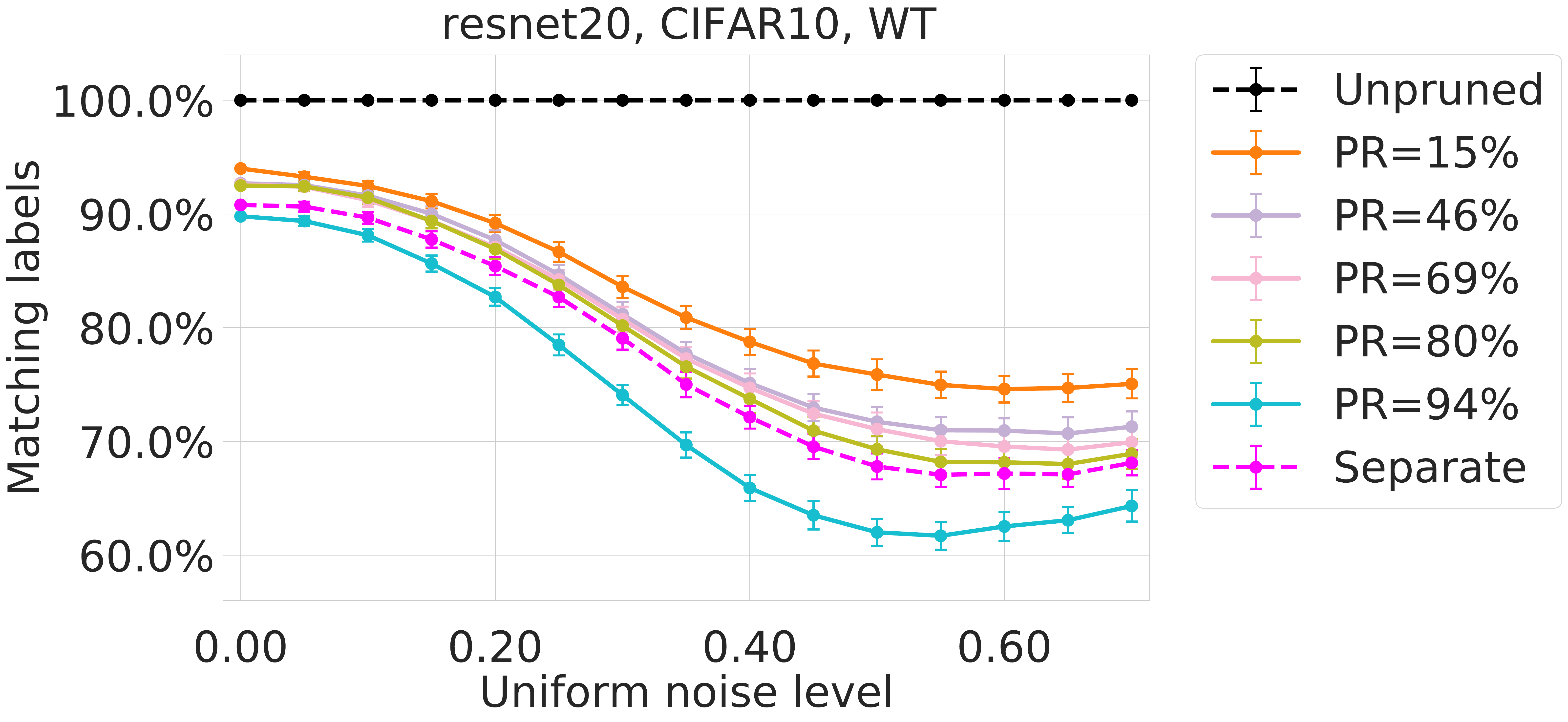}
    \subcaption{WT, Labels}
    \label{fig:noise_label_wt}
\end{minipage}%
\hfill
\begin{minipage}[t]{0.42\columnwidth}
    \includenoisegraphics{1.0}{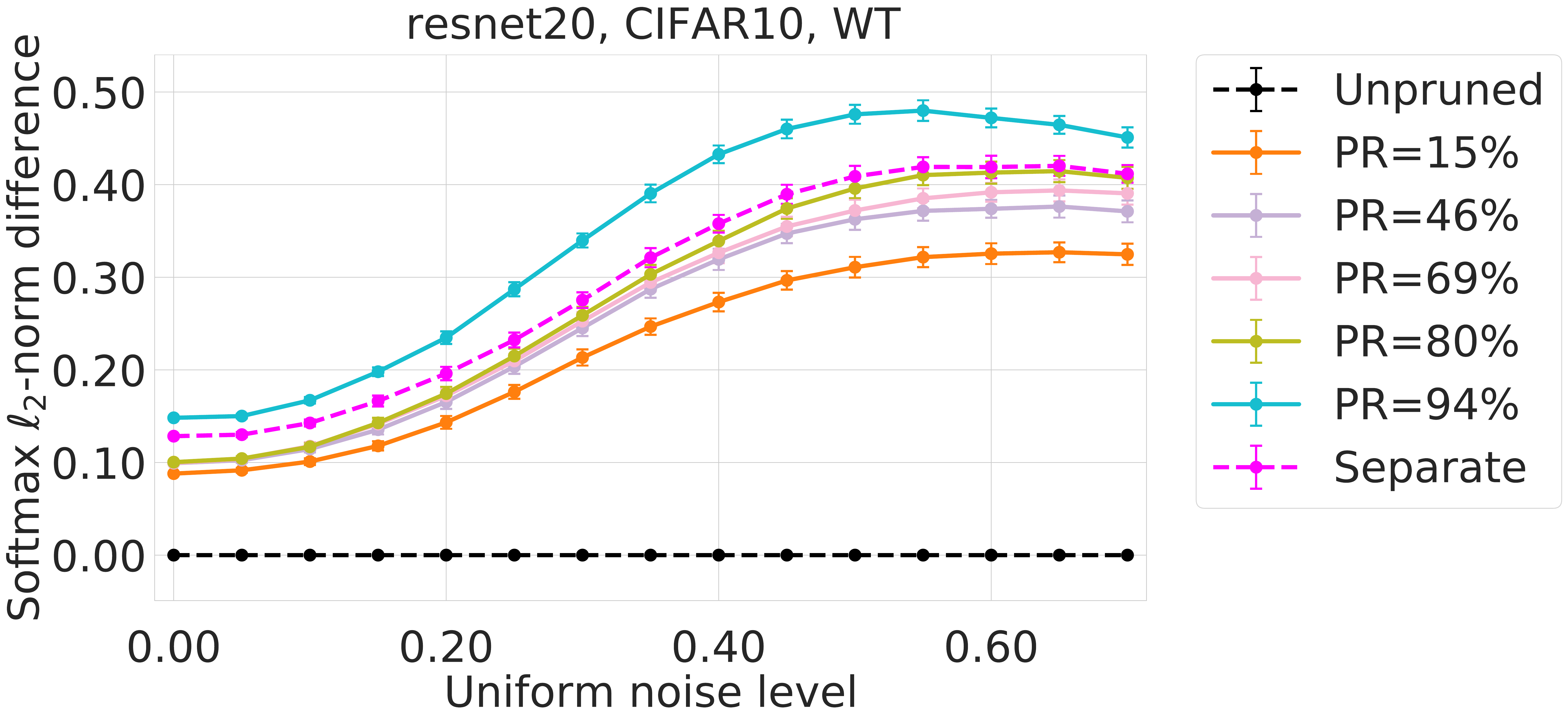}
    \subcaption{WT, Norm}
    \label{fig:noise_norm_wt}
\end{minipage}%
\hfill
\begin{minipage}[t]{0.15\columnwidth}
    \includenoiselegend[0.08]{resnet20_CIFAR10}
\end{minipage}
\caption{
The functional similarities between WT-pruned ResNet20 models measured by considering (a) the percentage of matching predictions, (b) the difference in the softmax output after injecting noise into the input. Pruned networks are more similar.
}
\label{fig:noise_similarities}
\vspace{-3ex}
\end{figure}

\subsection{Results}
\label{sec:distance-results}

\paragraph{Comparison of informative features.}
Figure~\ref{fig:sis-confidence-heatmap} shows heatmaps of mean confidence on masked images containing only the $10\%$ most informative features as ordered by \verb|BackSelect| from an unpruned network (ResNet20 on CIFAR-10), five WT-pruned networks, and a separate, unpruned network.
For each masked image containing only the informative features (found with respect to the predicted class), we evaluate the confidence toward the true class for all models to reveal whether such features are informative to the other models.
We find features informative to the unpruned network suffice for confident predictions by the pruned networks derived from it, but do not suffice for prediction by the separate, unpruned network.
We also find that informative features from pruned networks can be used for prediction by the original network, suggesting these models employ a similar decision-making process.
In general, our results suggest weight-pruned networks maintain higher confidence on parent features than do filter-pruned networks (see Appendix~\ref{sec:supp-distance-informative}).
For models pruned beyond commensurate accuracy (PR = 0.98 in Figure~\ref{fig:sis-confidence-heatmap}), the informative features are no longer predictive under any other model.

\paragraph{Noise similarities.}
In Figure~\ref{fig:noise_label_wt} the percentage of matching label predictions of WT-pruned ResNet20 networks are shown with respect to their unpruned counterpart for multiple noise levels. We can conclude that the predictions of the pruned networks tend to correlate with the predictions of the unpruned parent -- clearly more than the predictions from a separately trained, unpruned network. We also consider the overall difference in the $\ell_2$-norm between pruned and unpruned networks of the softmax output where we observe similar trends, see Figure~\ref{fig:noise_norm_wt}.
These results indicate that the decision boundaries of the pruned network tend to remain close to those of the unpruned network implying that during retraining properties of the original function are maintained. We note that the correlation decreases as we prune more corroborating our intuitive understanding of pruning.
\section{Pruning under Distribution Changes}
\label{sec:potential}
We highlighted that pruned networks behave functionally similarly, however, ultimately the performance is measured in terms of the loss or accuracy on previously unseen data. 
In this section, we investigate how pruned networks behave in the presence of shifts in the data distribution, including noise, weather, and other corruptions. 
While it is commonly known that out-of-distribution (\ood) data can harm the performance of neural networks~\cite{madry2018towards}, we specifically investigate whether pruned network suffer \emph{disproportionately more} from \ood data compared to their parent network. Answering this question affirmatively has profound implications on the practical deployment of pruned networks, specifically for safety-critical systems. 

To this end, we define a network's \emph{prune potential} to be the maximal prune ratio for which the pruned network achieves similar loss (up to margin $\delta$) compared to the unpruned one for data sampled from distribution $\DD$.
\begin{definition}[Prune Potential]
Given a neural network $f_\theta(x)$ with parameters $\theta$, input-label pair $(x,y) \sim \DD$, and loss function $\ell(\cdot, \cdot)$ the \emph{prune potential} $P(\theta, \DD)$ for some margin $\delta$ is given by
\begin{gather}
P(\theta, \DD)=
\textstyle \max_{c \in \{0, 1\}^{\abs{\theta}}} 
1 - \nicefrac{\norm{c}_0}{\norm{\theta}_0} \nonumber \\
\text{subject to} \label{eq:prune_potential} \\
\textstyle \E_{(x, y) \sim \DD} \left[\ell (y, f_{c \odot \hat\theta}(x)) - \ell (y, f_\theta(x))\right] \leq \delta, \nonumber
\end{gather}
where $\norm{\cdot}_0$ denotes the number of nonzero elements, and $c$ and $\hat\theta$ denote the prune mask and parameters, respectively, obtained from \textsc{PruneRetrain} (Algorithm~\ref{alg:prune}).
\end{definition}

The prune potential $\mathcal P (\theta, \DD)$ thus indicates how much of the network can be safely pruned with minimal additional loss incurred. In other words, it indicates to what degree the pruned network can \emph{maintain} the performance of the parent network. As an additional benefit the prune potential may act as a robust measure to gauge the \emph{overparameterization} of a network in the presence of distribution shifts. 

Moreover, we define a network's \emph{excess loss} to be the additional loss incurred under distributional changes of the input. 

\begin{definition}[Excess Loss]
Given a neural network $f_\theta(\cdot)$ with parameters $\theta$, training distribution $\DD$ from which we can sample input-label pairs $(x,y) \sim \DD$, test distribution $\DD'$ from which we can also sample input-label pairs, and loss function $\ell(\cdot, \cdot)$, the \emph{excess loss} $e(\theta, \DD')$ is given by
$$
e(\theta, \DD') = \E_{(x', y') \sim \DD'} \ell (y', f_\theta(x')) - \E_{(x, y) \sim \DD} \ell (y, f_\theta(x)).
$$
\end{definition}

The excess loss hereby indicates the expected performance drop of the network for distribution changes, which we can evaluate for various unpruned and pruned parameter sets for a given network architecture to understand to what extend the excess loss varies. 

\subsection{Methodology}
We choose test error (indicator loss function) to evaluate the prune potential and excess loss (excess error). We evaluate the constraint of~\eqref{eq:prune_potential} for a margin of $\delta = 0.5\%$.

We compare the prune potentials $p = P(\theta, \DD)$ and $p' = P(\theta, \DD')$ for two distributions $\DD$ and $\DD'$ to assess whether pruning up to the prune potential $p$ implies that we can also safely prune up to $p$ for $\DD'$. Specifically, the difference $p - p'$ in prune potential can indicate how much the prune potential varies and thus whether it is safe to prune the network up to its full potential $p$ when the input is instead drawn from $\DD'$. We note that in practice we may only have access to $\DD$ but not $\DD'$. Thus in order to safely prune a network up to some prune ratio $p$ it is crucial to understand to what degree the prune potential may vary for shifts in the distribution.

We also compare the excess error $e = e(\theta, \DD')$ and $\hat e = e(c \odot \hat \theta, \DD')$ for an unpruned and pruned network with parameters $\theta$ and $c \odot \hat \theta$, respectively. Note that the difference in excess error, $\hat e - e$, quantifies the additional error incurred by a pruned network under distribution changes compared to the additional error incurred by an unpruned network. 
Ideally, the difference $\hat e - e$ should be zero across all prune ratios, which would imply that the prune-accuracy trade-off for nominal data is indicative of the trade-off for \ood data.

We evaluate the prune potential and excess error using nominal test data (train distribution $\DD$) and \ood test data (test distribution $\DD'$). Specifically, we consider \ood data with random noise following  Section~\ref{sec:distance-methods}, and \ood data corrupted using state-of-the-art corruption techniques, i.e., CIFAR10.1~\cite{recht2018cifar10.1}, CIFAR10-C~\cite{hendrycks2019robustness} for CIFAR10; ObjectNet~\cite{barbu2019objectnet}, ImageNet-C~\cite{hendrycks2019robustness} for ImageNet; and VOC-C~\cite{michaelis2019dragon} for VOC. For CIFAR10-C, ImageNet-C, VOC-C we choose severity level 3 out of 5. The prune potential is evaluated separately for each corruption while the excess error is evaluated by averaging over all corruptions (test distribution).

\subsection{Results}

\begin{figure}[t!]
\begin{center}
\includegraphics[width=1.0\columnwidth]{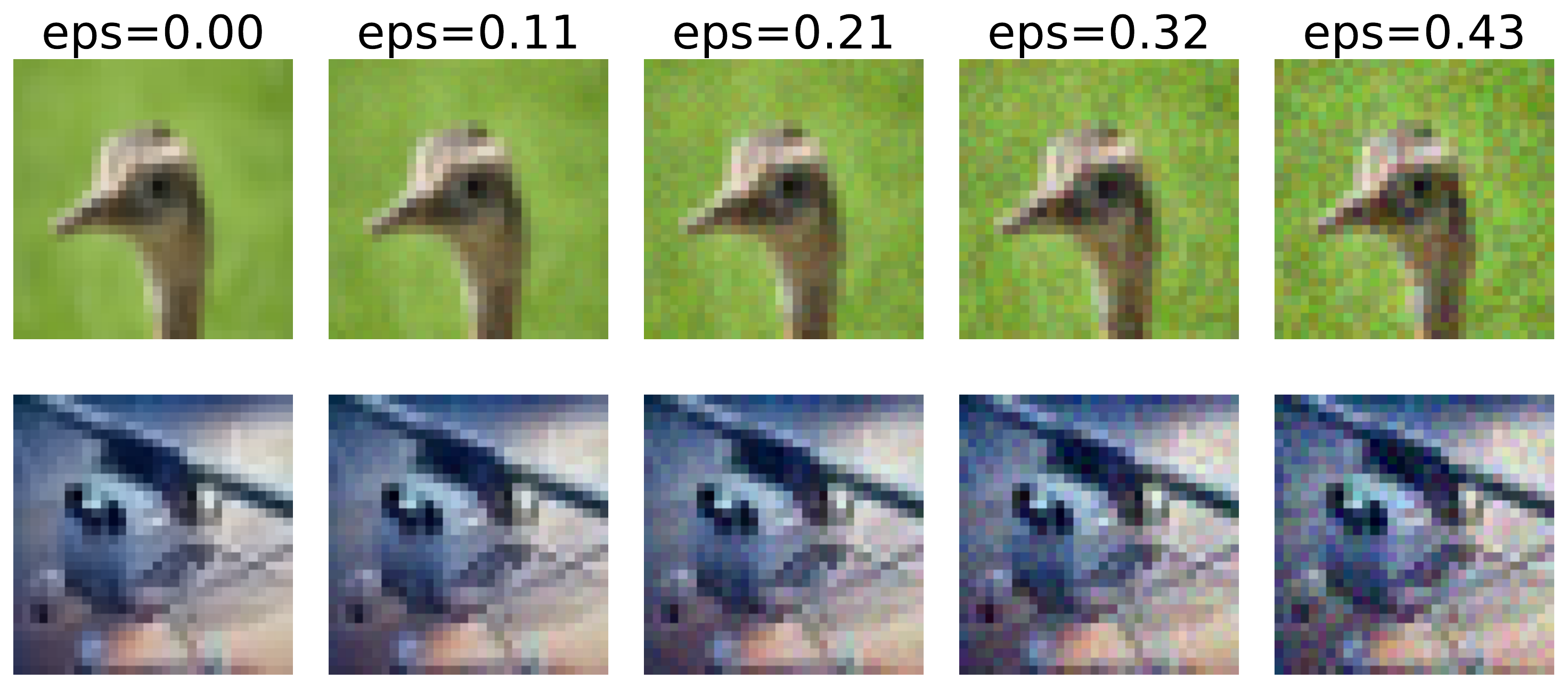}
\vspace{-4ex}
\caption{Example images from the CIFAR10 test dataset that were used in this study with various levels of noise injected. A human test subject can classify the images equally well despite the noise present.}
\label{fig:prune_potential_noisy_images}
\end{center}
\vspace{-4ex}
\end{figure}

\paragraph{Noise.}
We evaluated the prune potential of a ResNet20 (CIFAR10) for various noise levels, the results of which are shown in Figure~\ref{fig:prune_potential_noise}.
Initially, the network exhibits high prune potential, similar to the prune potential on the original test data (noise level $0.0$). However, as we increase the noise injected into the image the prune potential rapidly drops to $0\%$. As shown in Appendix~\ref{sec:supp-potential-noise} most networks' prune potential based on noise exhibit similar properties. This is particularly discomforting as the noise does not deteriorate a human's ability to classify the images correctly as can be seen from Figure~\ref{fig:prune_potential_noisy_images}. These results highlight we may not be able to significantly prune networks if maintaining performance on slightly harder data is the goal.

\begin{figure*}[t!]
\centering
\begin{minipage}[t]{0.33\textwidth}\vspace{0pt}%
    \centering
    \includegraphics[width=0.95\textwidth]{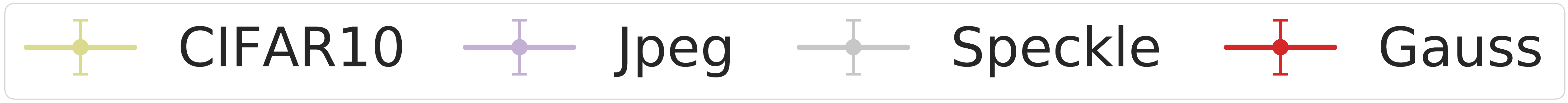}
\end{minipage}%
\hfill
\begin{minipage}[t]{0.33\textwidth}\vspace{0pt}%
    \includegraphics[width=1.0\textwidth]{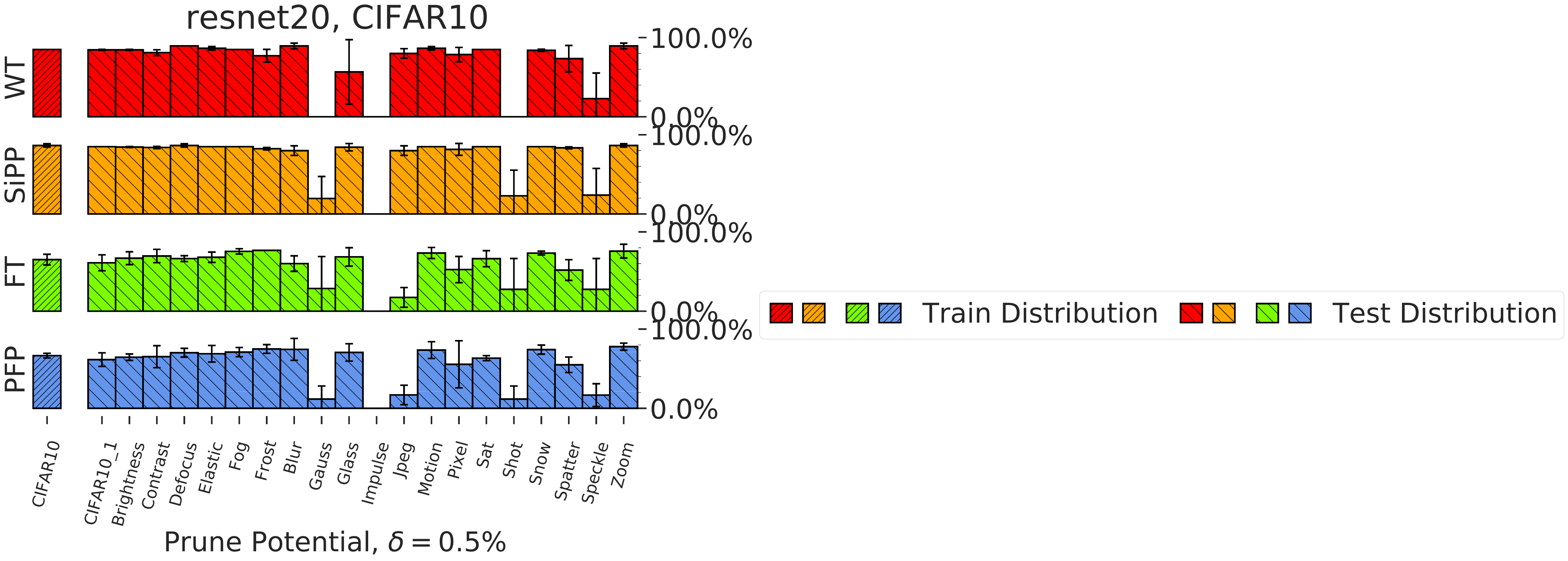}
\end{minipage}%
\hfill
\begin{minipage}[t]{0.33\textwidth}\vspace{0pt}%
    \centering
    \includegraphics[width=0.83\textwidth]{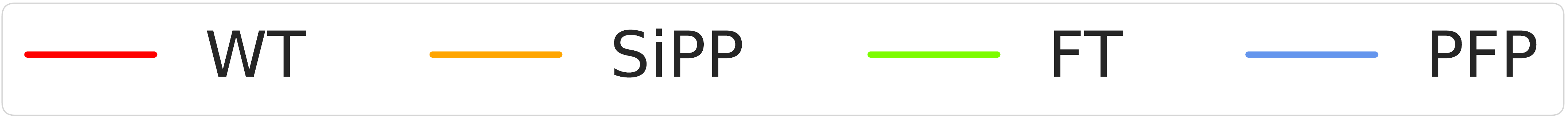}
\end{minipage}
\begin{minipage}[t]{0.33\textwidth}\vspace{0pt}%
    \includeprunecurvegraphics{1.0}{WT_resnet20_CIFAR10_CIFAR10}
    \vspace{0.9ex}
    \subcaption{Prune-test curve (WT)}
    \label{fig:comm_pr_corruption_wt}
\end{minipage}%
\hfill
\begin{minipage}[t]{0.33\textwidth}\vspace{0pt}%
    \includeppgraphics{1.0}{resnet20_CIFAR10_WT_SiPP}
    \vspace{-3.5ex}
    \subcaption{Weight prune potential}
    \label{fig:comm_pr_corruption_weight}
\end{minipage}%
\hfill
\begin{minipage}[t]{0.33\textwidth}\vspace{0pt}%
    \includeexcessgraphics{1.0}{resnet20_CIFAR10_WT_SiPP}
    \vspace{0.8ex}
    \subcaption{Excess error (Weight)}
    \label{fig:excess_error_weight}
\end{minipage}
\begin{minipage}[t]{1.0\textwidth}\vspace{0pt}%
\tiny
\vphantom{3}
\end{minipage}
\begin{minipage}[t]{0.33\textwidth}\vspace{0pt}%
    \includeprunecurvegraphics{1.0}{FT_resnet20_CIFAR10_CIFAR10}
    \vspace{0.9ex}
    \subcaption{Prune-test curve (FT)}
    \label{fig:comm_pr_corruption_ft}
\end{minipage}%
\hfill
\begin{minipage}[t]{0.33\textwidth}\vspace{0pt}%
    \includeppgraphics{1.0}{resnet20_CIFAR10_FT_PFP}
    \vspace{-3.5ex}
    \subcaption{Filter prune potential}
    \label{fig:comm_pr_corruption_filter}
\end{minipage}%
\hfill
\begin{minipage}[t]{0.33\textwidth}\vspace{0pt}%
    \includeexcessgraphics{1.0}{resnet20_CIFAR10_FT_PFP}
    \vspace{0.9ex}
    \subcaption{Excess error (Filter)}
    \label{fig:excess_error_filter}
\end{minipage}
\caption{
The prune potential for a ResNet20 on CIFAR10-C test datasets. We observe that depending on the type of corruption the network has significantly less prune potential than when measured w.r.t.\ the nominal CIFAR10 test accuracy.}
\label{fig:prune_potential_corruption}
\vspace{-2ex}
\end{figure*}

\paragraph{Prune-accuracy curves for corruptions.}
Separately, we investigated the prune potential for image corruptions based on the CIFAR10-C, ImageNet-C. 
In Figures~\ref{fig:comm_pr_corruption_wt} and~\ref{fig:comm_pr_corruption_ft} we show the test accuracy of pruned networks across various target prune ratios for a subset of CIFAR10-C corruptions for a ResNet20 pruned with WT and FT, respectively. In particular, for some simpler corruptions, such as Jpeg, the prune-accuracy curves closely resembles the original CIFAR10 curve, while for  metrics such as Speckle and Gauss the curve indicates a noticeable accuracy drop across all target prune ratios. Moreover, the prune-accuracy curve becomes more unpredictable and less stable as indicated by the significantly higher variance of the resulting accuracy.
We thus conclude that the achievable accuracy of the network depends on the pruning method and the target prune ratio, however, we observe an equally strong dependence on the chosen test metric. Consequently, this affects the prune potential of the network highlighting the sensible trade-off between generalization performance and prune potential.

\begin{figure}[t]
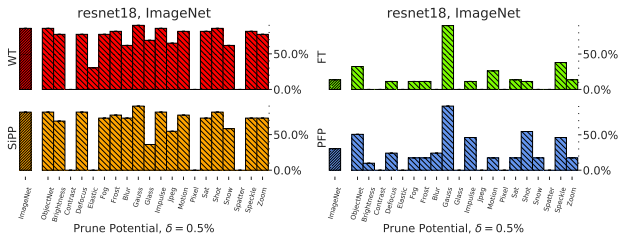

    \centering
    \begin{minipage}[t]{0.24\textwidth}\vspace{0pt}%
        \includeppgraphics{1.0}{resnet18_ImageNet_WT_SiPP}
    \end{minipage}%
    \begin{minipage}[t]{0.24\textwidth}\vspace{0pt}%
        \includeppgraphics{1.0}{resnet18_ImageNet_FT_PFP}
    \end{minipage}
    \caption{Prune potential of a ResNet18 (ImageNet).}
    \label{fig:prune_pot_corruption_imagenet}
    \vspace{0ex}
\end{figure}

\paragraph{Prune potential for corruptions.}
For each corruption we then extracted the resulting prune potential from the prune-accuracy curves, see Figures~\ref{fig:comm_pr_corruption_weight} and~\ref{fig:comm_pr_corruption_filter} for weight pruning and filter pruning, respectively. 
In particular, for corruptions, such as Gauss, Impulse, or Shot, we observe that the network's prune potential hits (almost) $0\%$ implying that any form of pruning may adversely affect the network's performance under such circumstances. 
We repeated the same experiment for a ResNet18 trained on ImageNet and tested on ImageNet-C, see Figure~\ref{fig:prune_pot_corruption_imagenet}. Noticeably, the network exhibits significantly higher variance in the prune potential across different corruptions compared to the networks tested on CIFAR10. This effect is also more pronounced for filter pruning methods. 
In Appendix~\ref{sec:supp-potential-corruptions}, we provide results for additional networks for both CIFAR10 and ImageNet corroborating our findings presented here. 

\paragraph{Choice of $\delta$.}
We additionally investigate how the prune potential is affected by our choice of $\delta$ (see Appendix~\ref{sec:supp-potential-delta}). While the actual value of the prune potential is naturally affected by $\delta$, we find that our observations of the resulting trends remain unaffected by our particular choice of $\delta$. Hence, we simply choose $\delta=0.5\%$ uniformly across all experiments reflecting the requirement that our pruned network should be close in accuracy to the parent network while allowing some slack to increase the prune potential.

\paragraph{Excess error.}
In contrast to the prune potential, the difference in excess error enables us to quantify across multiple prune ratios how much \emph{additional error} is incurred by the pruned network on top of the unpruned network's excess error when tested on \ood data. A non-zero difference thus indicates how the prune-accuracy curve changes under distribution changes. In other words, the difference in excess error quantifies the difference in error between the pruned and unpruned network \emph{on top} of the difference in error that is incurred for nominal test data (train distribution).

We evaluated the difference in excess error between pruned and unpruned ResNet20s trained on CIFAR10 for various prune ratios (see Figures~\ref{fig:excess_error_weight},~\ref{fig:excess_error_filter}). Note that by definition the excess error is $0\%$ for a prune ratio of $0\%$. We observe that the difference in excess error can reach upwards of $2\%$ and $3\%$ for weight and filter pruning, respectively. Moreover, the higher the prune ratio the more variance we can observe indicating that the pruned network's behavior becomes less predictable overall. These observations strongly indicate that pruned networks suffer disproportionally more from \ood data across a wide spectrum of prune ratios and that the additional performance drop on \ood data is positively correlated with the prune ratio. In other words, while current pruning techniques achieve commensurate accuracy for high prune ratios on nominal test data, the same pruning techniques do not maintain commensurate accuracy for even small prune ratios on \ood test data. Additional results are presented in Appendix~\ref{sec:supp-potential-excess}.

\paragraph{Measuring overparameterization.}
The results presented so far in this section highlight that the prune potential on nominal test data does not reliably indicate the overall performance of the pruned network. 
This may lead to novel insights into understanding the amount of overparameterization in deep networks.
In Appendix~\ref{sec:supp-potential-overparameterization}, we summarize the prune potential across all tested data distributions and networks as a way to gauge the amount of overparameterization of a network. 
A subset of the results are shown in Table~\ref{tab:overparameterization}.
While some networks' prune potentials are significantly affected by changes in the distribution, other networks' prune potentials are virtually unaffected.
Take for example the weight prune potential on nominal test data (training distribution) of a VGG16 and a WRN16-8, which is around 98\% for both. However, when both networks are evaluated on \ood test data (test distribution) they exhibit distinctly different behaviors. While the WRN16-8's prune potential remains fairly stable at around 95\% (3\% drop), the VGG16's prune potential falls to 80\% (18\% drop).

Overall, these results illustrate the fact that the prune potential may act as a robust measure of a network's genuine overparameterization in theory, and may also be helpful in informing the practitioner on the extent of pruning that should be conducted prior to deployment in practice.

\begin{table}[t]
\small
\centering
\begin{tabular}{c|c|ccc}
\multirow{2}{*}{Model} & \multirow{2}{*}{Method} & \multicolumn{3}{c}{Prune Potential ($\%$)}  \\ 
& & Train Dist. & Test Dist. & Diff. \\ \hline
\multirow{2}{*}{ResNet20}
& WT
& 84.9 $\pm$ 0.0 & \textbf{66.7 $\pm$ 3.3} 
& -18.2 \\
& FT 
& 65.0 $\pm$ 6.7 & \textbf{55.3 $\pm$ 4.8}
& -9.7 \\ \hline
\multirow{2}{*}{VGG16}
& WT
& 98.0 $\pm$ 0.0 & \textbf{80.9 $\pm$ 2.2}
& -17.1 \\
& FT
& 85.4 $\pm$ 2.4 & \textbf{66.3 $\pm$ 0.5}
& -19.1 \\ \hline
\multirow{2}{*}{WRN16-8}  
& WT
& 98.0 $\pm$ 0.0 & \textbf{95.7 $\pm$ 0.7}
& -2.3 \\
& FT 
& 86.2 $\pm$ 1.3 & \textbf{75.7 $\pm$ 4.1}
& -10.5 \\ \hline\hline
\multirow{2}{*}{ResNet18}  
& WT 
& 85.8 & \textbf{63.6}
& -22.2 \\
& FT 
& 13.7 & \textbf{13.5}
& -0.2 \\
\end{tabular}
\caption{The prune potential of various networks trained on CIFAR10 (upper part) and ImageNet (lower part) evaluated on the train and test distribution, which consist of nominal data and the average over all corruptions, respectively.}
\label{tab:overparameterization}
\vspace{-2ex}
\end{table}
\begin{figure*}[t!]
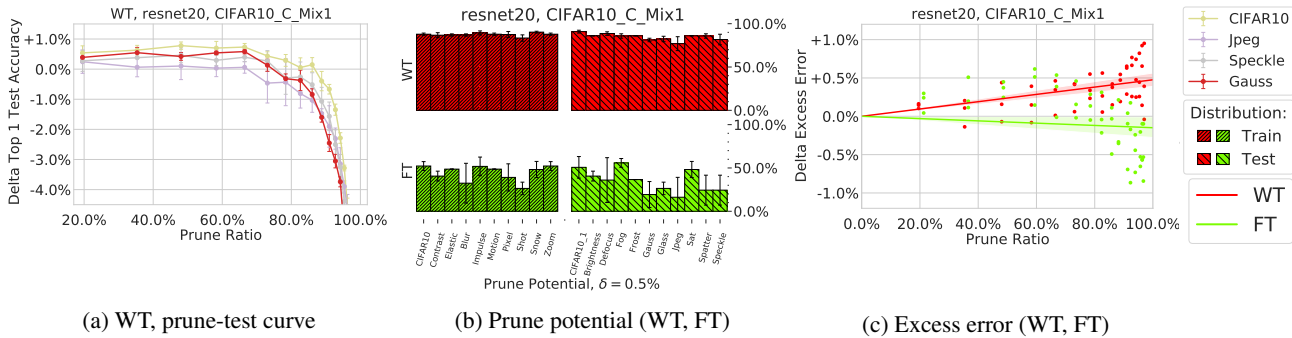

\centering
\begin{minipage}[t]{0.30\textwidth}\vspace{0pt}%
    \includeprunecurvegraphics{1.0}{WT_resnet20_CIFAR10_C_Mix1_CIFAR10}
    \vspace{1.4ex}
    \subcaption{WT, prune-test curve}
    \label{fig:robust_comm_pr_corruption_wt}
\end{minipage}%
\hfill
\begin{minipage}[t]{0.30\textwidth}\vspace{0pt}%
    \includeppgraphics{1.0}{resnet20_CIFAR10_C_Mix1_WT_FT}
    \subcaption{Prune potential (WT, FT)}
    \label{fig:robust_prune_pot}
\end{minipage}%
\hfill
\begin{minipage}[t]{0.30\textwidth}\vspace{0pt}%
    \includeexcessgraphics{1.0}{resnet20_CIFAR10_C_Mix1_WT_FT}
    \vspace{1.4ex}
    \subcaption{Excess error (WT, FT)}
    \label{fig:robust_excess_error}
\end{minipage}%
\begin{minipage}[t]{0.09\textwidth}\vspace{0pt}%
    \includeprunecurvelegend{WT_resnet20_CIFAR10_CIFAR10}
    \includepplegend{resnet20_CIFAR10_C_Mix1_WT_FT}
    \includeexcesslegend{resnet20_CIFAR10_C_Mix1_WT_FT}
\end{minipage}
\caption{
The prune potential of a ResNet20 shown for corruptions that were included (train distribution) and excluded (test distribution) during training. The prune-accuracy curves in (a) are shown for corruptions from the test distribution.}
\label{fig:robust_prune_potential}
\vspace{-2ex}
\end{figure*}

\section{Towards Robust Pruning}
\label{sec:robustness}
Our experiments raise the question whether the decreased performance of pruned networks is a limitation of our current pruning and training techniques or whether it is an  inherent limitation of pruned, i.e., smaller, networks themselves. To this end, we investigated whether training (and retraining) in a robust manner can benefit the pruned network and minimize the effects we observed previously.

\subsection{Methodology}
To test the hypothesis that we can regain some of the robustness properties of the unpruned network we repeated the experiments of Section~\ref{sec:potential} but during (re-)training we incorporated a randomly chosen fixed subset of nine corruptions from CIFAR10-C and ImageNet-C into the data augmentation pipeline. That is, every time we sample an image from the train set during (re-)training we choose an image corruption (or no corruption) uniformly at random to corrupt the image effectively altering the train distribution that the network sees. The remaining corruptions are not used during training and make up the new test distribution. Additional experimental details are provided in Appendix~\ref{sec:supp-robustness}.

\subsection{Results}

\paragraph{Prune-accuracy curves and prune potential.}
In Figure~\ref{fig:robust_comm_pr_corruption_wt} we show the prune-accuracy curves for three corruptions from the test distribution for WT-pruned ResNet20s. Compared to the results in Figure~\ref{fig:comm_pr_corruption_wt} where we did not perform robust (re-)training we observe that the prune-accuracy curves are much more stable and that the prune-accuracy curve on nominal test data (CIFAR10) is more predictive of the others. However, the results on the evaluation of the prune potential as shown in Figure~\ref{fig:robust_prune_pot} for weight and filter pruning reveal that even in this setting the prune potential can be significantly lower (or exhibit high variation over multiple trials) for some of the corruptions from the test distribution. These observations further corroborate our findings but also highlight the beneficial effects of robust training in efficiently alleviating some of the short-comings (see Appendix~\ref{sec:supp-robustness} for a complete exposition of the results).

\paragraph{Excess error.}
Similar trends can also be observed when considering the difference in excess error as shown in Figure~\ref{fig:robust_excess_error}. While we can reduce the correlation between prune ratio and excess error, we can not entirely eliminate it. Moreover, we can still observe high variations in the excess error confirming the sensible trade-off between generalization performance and prune potential.

\paragraph{Implicit regularization.}
In this section, we show that we can regain much of the prune potential even under distribution changes, at least when we can incorporate these additional data points into our training pipeline. For example, the weight prune potential for a ResNet20 for both the nominal and robust training scenario is around $85\%$. Consequently, we argue that both pruned and unpruned networks have sufficient capacity to represent the underlying distribution given that the training is performed using an appropriate optimization pipeline. 

Specifically, previous work noted that overparameterized networks may benefit from implicit regularization when optimized with a stochastic optimizer and that more parameters amplify this effect. We can confirm these observations in the sense that pruned networks suffer disproportionally more from \ood data than unpruned networks with more parameters. That is, unpruned network exhibit more implicit regularization through SGD leading to more robustness. However, we can regain some of the robustness properties by adding \emph{explicit} regularization during training in the form of data augmentation. We can thus ``trade'' the implicit regularization potential which we lose by removing parameters for explicit regularization through data augmentation. 

\paragraph{Choice of test distribution.}
While our results suggest that robust training indeed improves the generalization of pruned networks for \ood data, we would like to emphasize that our conclusion intrinsically hinges upon on the choice of train and test distribution. While we did strictly separate the corruptions used during train and test time, these corruptions can be loosely categorized into four types, i.e. noise, blur, weather, digital, all of which are present in both the train and test distribution. Therefore, we suspect that for significantly different corruption models (or adversarial inputs) we may observe more significant trade-offs resembling the results of Section~\ref{sec:potential} where we performed nominal (re-)training. This is a consequence
of requiring additional explicit regularization since the explicit regularization must be modeled. 

\section{Discussion}
\label{sec:discussion}

\paragraph{Weight vs filter pruning.}
We find that across all tested corruptions filter pruning methods are more error-prone and have  lower prune potential compared to weight pruning. We conjecture this trend stems from the fact that structured pruning is overall a harder problem, implying that structurally pruned networks are less capable of maintaining the properties of the parent network when compared to those generated by weight-based pruning.

\paragraph{Genuine overparameterization.}
In light of our results we conjecture that while the high capacity of modern networks may not be strictly necessary to achieve high test accuracy -- since pruned networks with commensurate accuracy exist --, the "excess" capacity of these networks may be beneficial to maintaining other crucial properties of the network, such as its ability to perform well on unforeseen or out-of-distribution data. This challenges the common wisdom that modern networks are overparameterized and, hence, contain redundant parameters that can be pruned in a straight-forward manner without "loss of performance." 
Unlike prior work that has predominantly pointed to the test accuracy as a gauge for overparameterization (and thus the ability to prune a network), we hypothesize that a more robust and accurate measure of \emph{genuine} overparameterization is one that not only considers test accuracy, but also the minimum (or average) prune potential over a variety of tasks.
Studying the prune potential is thus not only useful to study the ability to safely prune a network but also has the positive side-effect of establishing a robust measure of network overparameterization.

\paragraph{Implicit regularization.}
Our studies reveal that the amount of overparameterization is not only a function of the task and the network size but also a function of the \emph{training procedure}. Specifically, we can prune the network more if we explicitly regularize the network during (re-)training thus increasing the ``genuine'' overparameterization of the network which implies a higher prune potential. However, with fewer parameters (due to pruning more) we trade in some of the implicit regularization potential from SGD. Since implicit regularization is not necessarily model-based we can only regain the robustness of the pruned network for known, i.e. modeled, distribution changes. 

\paragraph{Generalization-aware pruning.}
Based on our results we formulate a set of guidelines for pruning in practice as shown in Section~\ref{sec:introduction}. We argue that in order to reliably and robustly deploy pruned networks especially in the context of safety-critical systems we should not only designate a \emph{hold-out data set} (test set) but also a \emph{hold-out data distribution} (test distribution). By assessing the performance of the pruned network on data from the train and test distribution, we can then quantify the effect of pruning in a way that can unearth some of the short-comings that are \emph{lost in pruning} and are not apparent from a plain prune-accuracy curve on nominal test data. Following our framework, a practitioner will be able to more reliably assess whether the pruned network can be considered as performant (in a robust sense) as the unpruned network. 
\section{Conclusion}
In this work, we have investigated the effects of the pruning process on the functional properties of the pruned network relative to its uncompressed counterpart. Our empirical results suggest that pruned models are functionally similar to the their uncompressed counterparts but that, despite this similarity, the prune potential of the network varies significantly on a task-dependent basis: the prune potential decreases significantly as the difficulty of the inference task increases. 
Our findings underscore the need to consider task-specific evaluation metrics beyond test accuracy prior to deploying a pruned network and provide novel insights into understanding the amount of network overparameterization in deep learning. 
We envision that our framework may invigorate further work towards rigorously understanding the inherent model size-performance trade-off and help practitioners in adequately designing and pruning network architectures in a task-specific manner.

\section*{Acknowledgements}
This research was supported in part by the U.S. National Science Foundation (NSF) under Award 1723943, Office of Naval Research (ONR) Grant N00014-18-1-2830, and JP Morgan Chase. We thank them for their support.
We would like to further thank Siddhartha Jain for the insightful discussions during the project conception phase and for helping with early drafts of the paper.

%%%%%%%%%%%%%%%%%%%%%%%%%%%%%%%%%%%%%%%%
%%%%%%%%%%%%% BIBLIOGRAPHY %%%%%%%%%%%%%
%%%%%%%%%%%%%%%%%%%%%%%%%%%%%%%%%%%%%%%%

\bibliography{misc/references}
\bibliographystyle{mlsys2021}

%%%%%%%%%%%%%%%%%%%%%%%%%%%%%%%%%%%%%%%%
%%%%%%%% SUPPLEMENTARY MATERIAL %%%%%%%%
%%%%%%%%%%%%%%%%%%%%%%%%%%%%%%%%%%%%%%%%

% setup
\clearpage
\newpage
\onecolumn
\appendix

% supplementary sections
\section*{Broader Impact Statement}
\label{sec:impact}
In this paper, we study the impact of pruning when deploying neural networks in real-world conditions. 
This is of particular importance since the main motivation to prune neural networks lies within training neural networks that are simultaneously efficient and accurate. Henceforth, these networks can then be deployed in resource-constrained environments, such as robotics, to achieve tasks that could otherwise only be computed on large-scale compute infrastructure. 

However, we show that pruned neural networks do not necessarily perform on par with their unpruned parent but rather that they exhibit significant performance decreases depending on slight variations within their assigned task. This raises concerns with regards to the ability to find effectively pruned architectures with current network pruning techniques. With the increasing amount of applications for deep learning including on small-scale devices, we have to vigilantly monitor and consider the effects, brittleness, and potential biasedness of small-scale neural networks at an even higher degree than for regular deep networks.
\section{Overview of the Appendix}
\label{sec:supp-overview}
In the main part of the paper, we have focused on representative subsets of each experiment. In the following, we provide additional experimental details and present the complete set of conducted experiments. Specifically, the supplementary material contains the following sections:

\begin{itemize}
    \item 
    \emph{Section~\ref{sec:supp-results}}: additional experimental details and hyperparameters for how we prune networks
    \item
    \emph{Section~\ref{sec:supp-distance}}: experiments pertaining to measuring the functional similarities between pruned and unpruned networks
    \item
    \emph{Section~\ref{sec:supp-potential}}: experiments pertaining to the prune potential and excess error of pruned networks
    \item
    \emph{Section~\ref{sec:supp-robustness}}: experiments pertaining to the prune potential and excess error of pruned networks with robust (re-)training
\end{itemize}

In addition to the results presented in the main part of the paper, we present further experimental evidence for our claims across a variety of architectures and data sets.

We provide the code to reproduce our experiments at \url{https://github.com/lucaslie/torchprune}.
\section{Pruning Results}
\label{sec:supp-results}
Our experimental evaluations are based on a variety of neural network architectures including ResNets~\cite{he2016deep}, VGGs~\cite{simonyan2014very}, DenseNets~\cite{huang2017densely}, and WideResNets~\cite{zagoruyko2016wide} trained on CIFAR10~\cite{torralba200880} and ImageNet~\cite{ILSVRC15}. We also conduct experiments on a DeeplabV3~\cite{chen2017rethinking} with a ResNet50 backbone trained on the Pascal VOC 2011 segmentation data set~\cite{everingham2015pascal}. In the following section we outline the experimental details of the
experiments on which we base our observations. All networks were trained and evaluated on a compute cluster with NVIDIA RTX 2080Ti and NVIDIA Titan RTX, and the experiments were implemented in PyTorch~\cite{paszke2017automatic}.
For each trained network, we summarize the hyperparameters and the resulting prune results on the nominal test data.

\begin{table}
    %\small
    \centering
    \begin{tabular}{cr|cccc}
        & & VGG16 & Resnet20/56/110 & DenseNet22 & WRN-16-8 \\ \hline
        \multirow{9}{*}{Train} 
        & test error    &  7.19          & 8.6/7.19/6.43 
                        &  10.10         & 4.81 \\
        & loss          & cross-entropy  & cross-entropy
                        & cross-entropy  & cross-entropy \\
        & optimizer     & SGD            & SGD 
                        & SGD            & SGD  \\
        & epochs        & 300            & 182
                        & 300            & 200  \\
        & warm-up       & 5              & 5    
                        & 5              & 5    \\
        & batch size    & 256            & 128
                        & 64             & 128   \\
        & LR            & 0.05           & 0.1
                        & 0.1            & 0.1  \\
        & LR decay      & 0.5@\{30, \ldots\} & 0.1@\{91, 136\}
                        & 0.1@\{150, 225\}   & 0.2@\{60, \ldots\}\\
        & momentum      & 0.9            & 0.9
                        & 0.9            & 0.9  \\
        & Nesterov      & \xmark         & \xmark      
                        & $\checkmark$   & $\checkmark$ \\
        & weight decay  & 5.0e-4         & 1.0e-4 
                        & 1.0e-4         & 5.0e-4   \\ \hline
        \multirow{2}{*}{Prune}
        & $\gamma$      & 1.0e-16        & 1.0e-16
                        & 1.0e-16        & 1.0e-16  \\
        & $\alpha$      & 0.85           & 0.85
                        & 0.85           & 0.85
    \end{tabular}
    \vspace{2ex}
    \caption{We report the hyperparameters used during training, pruning, and retraining for various convolutional architectures on CIFAR-10. LR hereby denotes the learning rate and LR decay denotes the learning rate decay that we deploy after a certain number of epochs. During retraining we used the same hyperparameters. $\{30, \ldots\}$ denotes that the learning rate is decayed every 30 epochs.}
    \label{tab:cifar_hyperparameters}
\end{table}

\subsection{Experimental Setup for CIFAR10}
\label{sec:supp-cifar10-hyperparameters}

All hyperparameters for training, retraining, and pruning are outlined in Table~\ref{tab:cifar_hyperparameters}.
For training CIFAR10 networks we used the training hyperparameters outlined in the respective original papers, i.e., as described by \citet{he2016deep}, \citet{simonyan2014very}, \citet{huang2017densely}, and \citet{zagoruyko2016wide} for ResNets, VGGs, DenseNets, and WideResNets, respectively. For retraining, we did not change the hyperparameters and repurposed the training hyperparameters following the approaches of~\citet{renda2020comparing, liebenwein2020provable}. We added a warmup period in the beginning where we linearly scale up the learning rate from 0 to the nominal learning rate.
Iterative pruning is conducted by repeatedly removing the same ratio of parameters (denoted by $\alpha$ in Table~\ref{tab:cifar_hyperparameters}). The prune parameter $\gamma$ describes the failure probability of the (provable) randomized pruning algorithms SiPP and PFP. We refer the reader to the respective papers for more details, see the papers by~\citet{sipp2019} and~\cite{liebenwein2020provable} for SiPP and PFP, respectively.

\subsection{Pruning Performance on CIFAR10}
\label{sec:supp-cifar10-pruning}
\begin{figure}
\centering
\begin{minipage}[t]{0.33\textwidth}
    \includegraphics[width=\textwidth]{fig/resnet20_CIFAR10_e182_re182_cascade_int23_CIFAR10_acc_param.pdf}
    \subcaption{Resnet20}
\end{minipage}%
\begin{minipage}[t]{0.33\textwidth}
    \includegraphics[width=\textwidth]{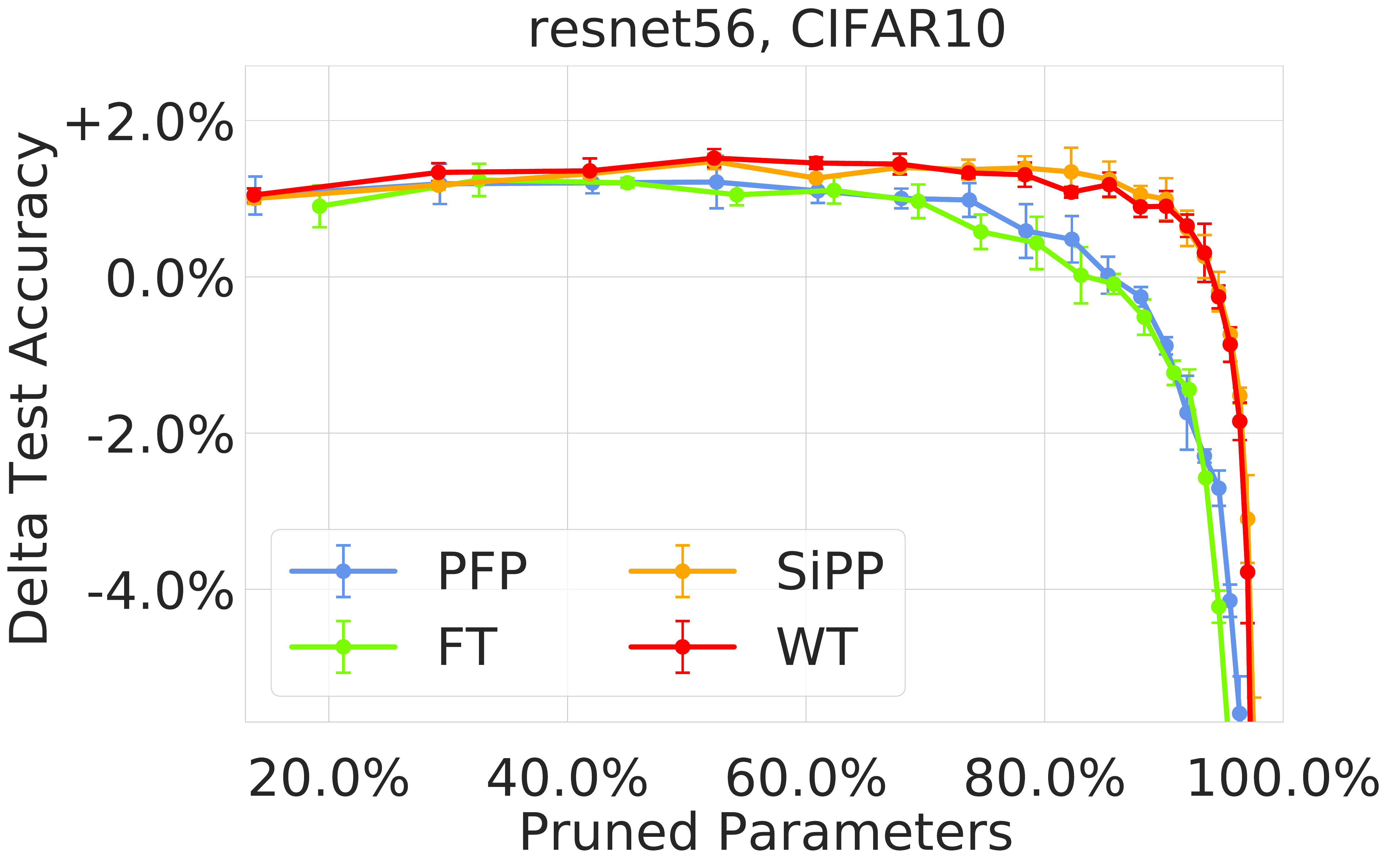}
    \subcaption{Resnet56}
\end{minipage}%
\begin{minipage}[t]{0.33\textwidth}
    \includegraphics[width=\textwidth]{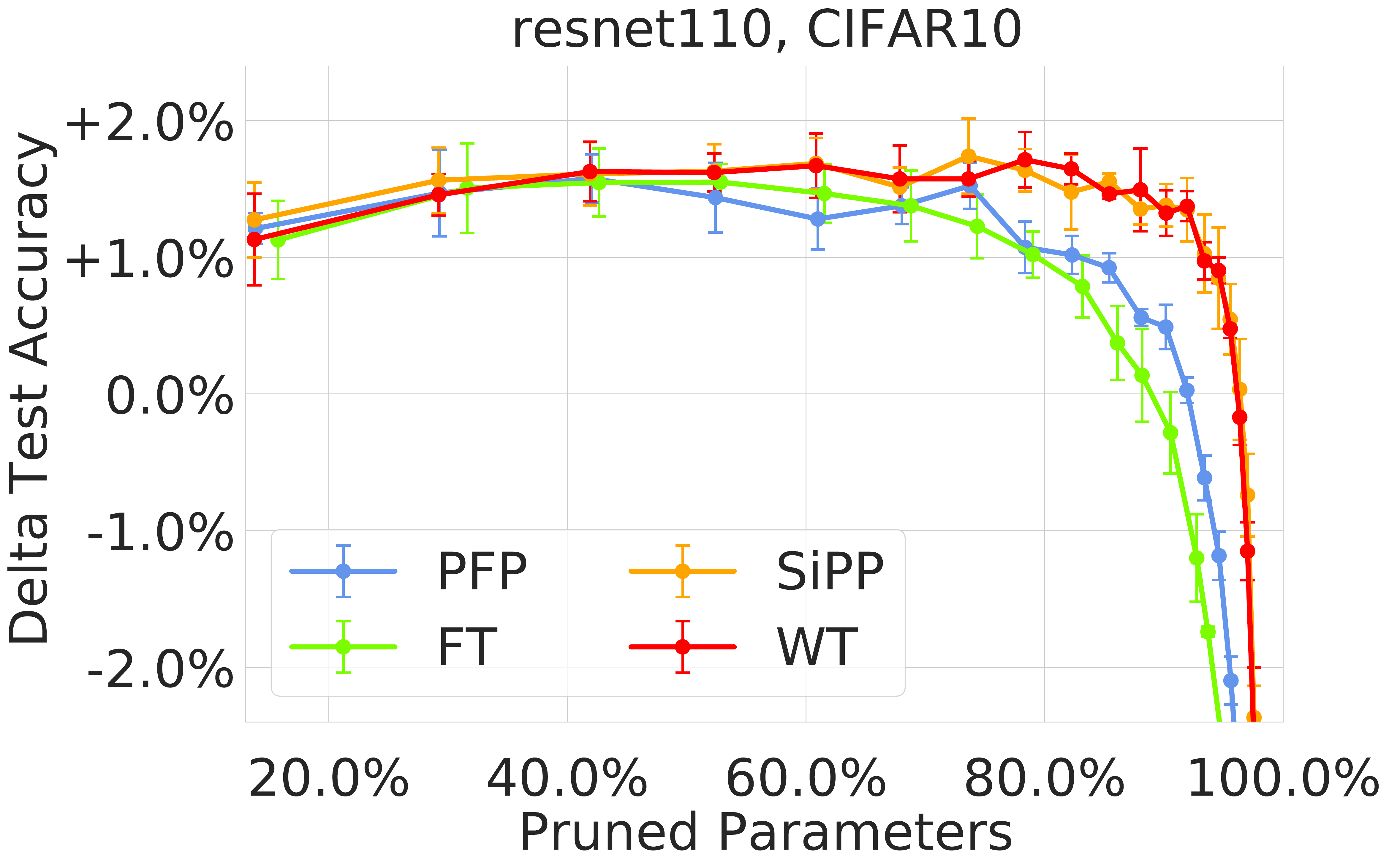}
    \subcaption{Resnet110}
\end{minipage}%
\vspace{2ex}
\begin{minipage}[t]{0.33\textwidth}
    \includegraphics[width=\textwidth]{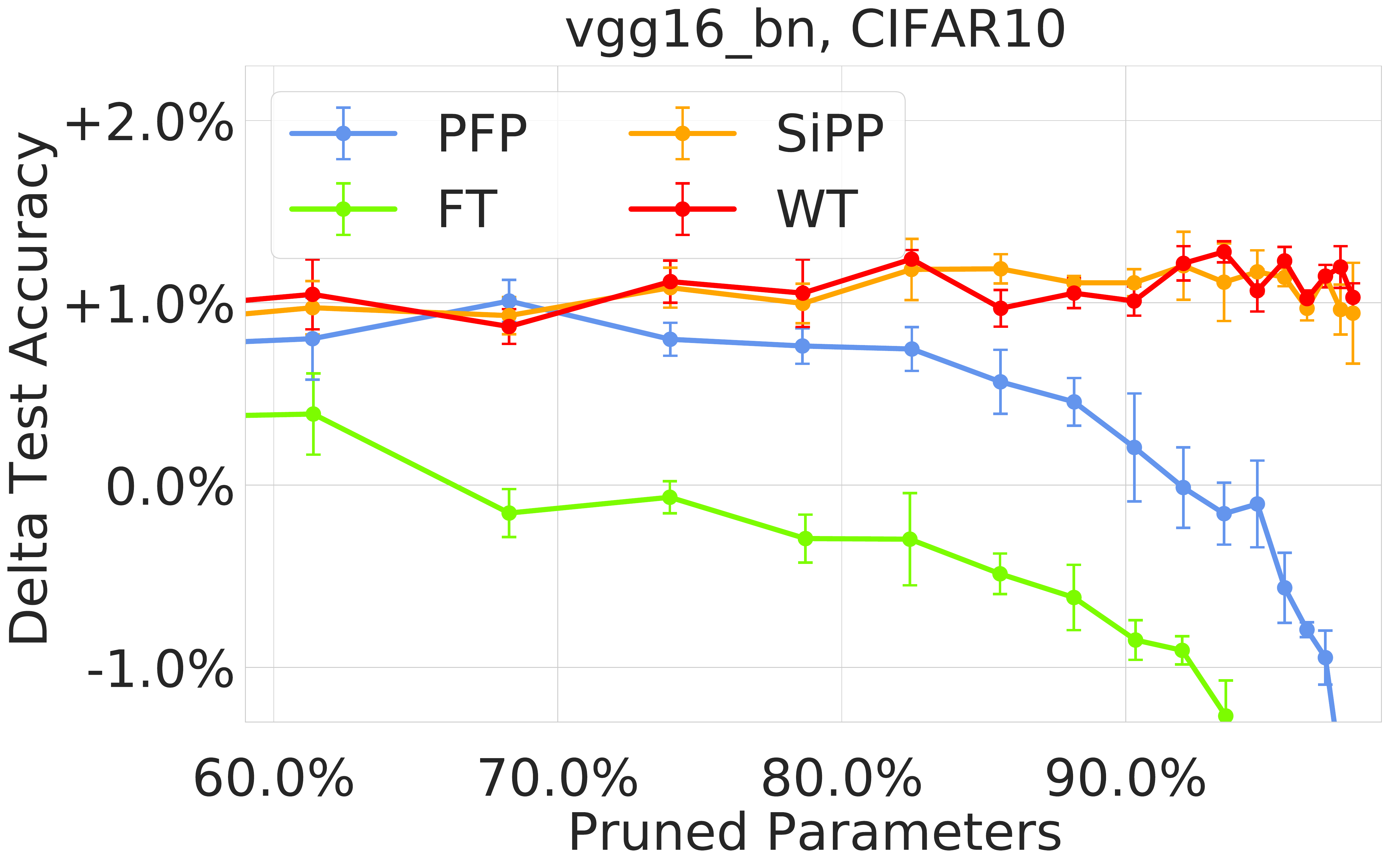}
    \subcaption{VGG16}
\end{minipage}%
\begin{minipage}[t]{0.33\textwidth}
    \includegraphics[width=\textwidth]{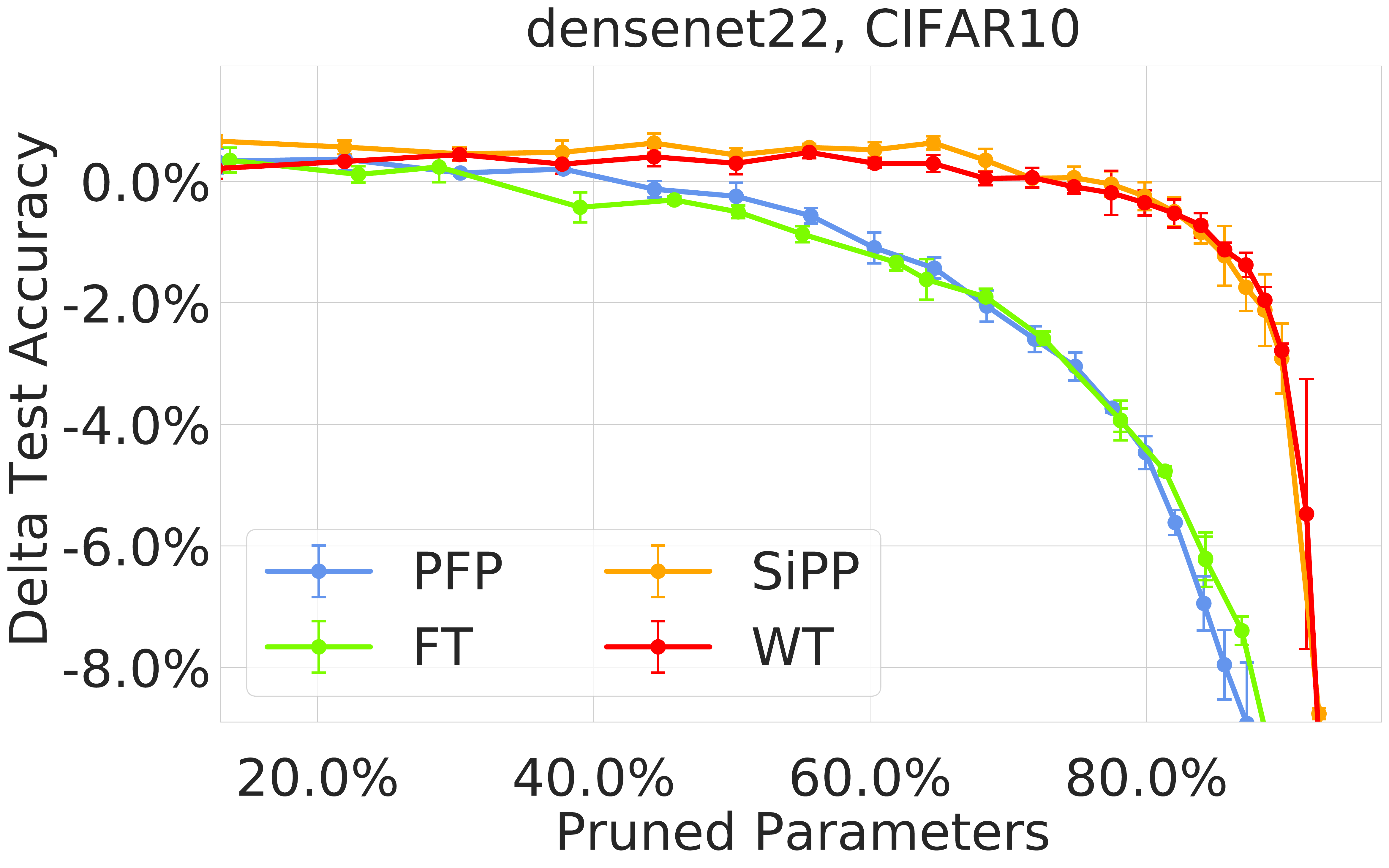}
    \subcaption{DenseNet22}
\end{minipage}%
\begin{minipage}[t]{0.33\textwidth}
    \includegraphics[width=\textwidth]{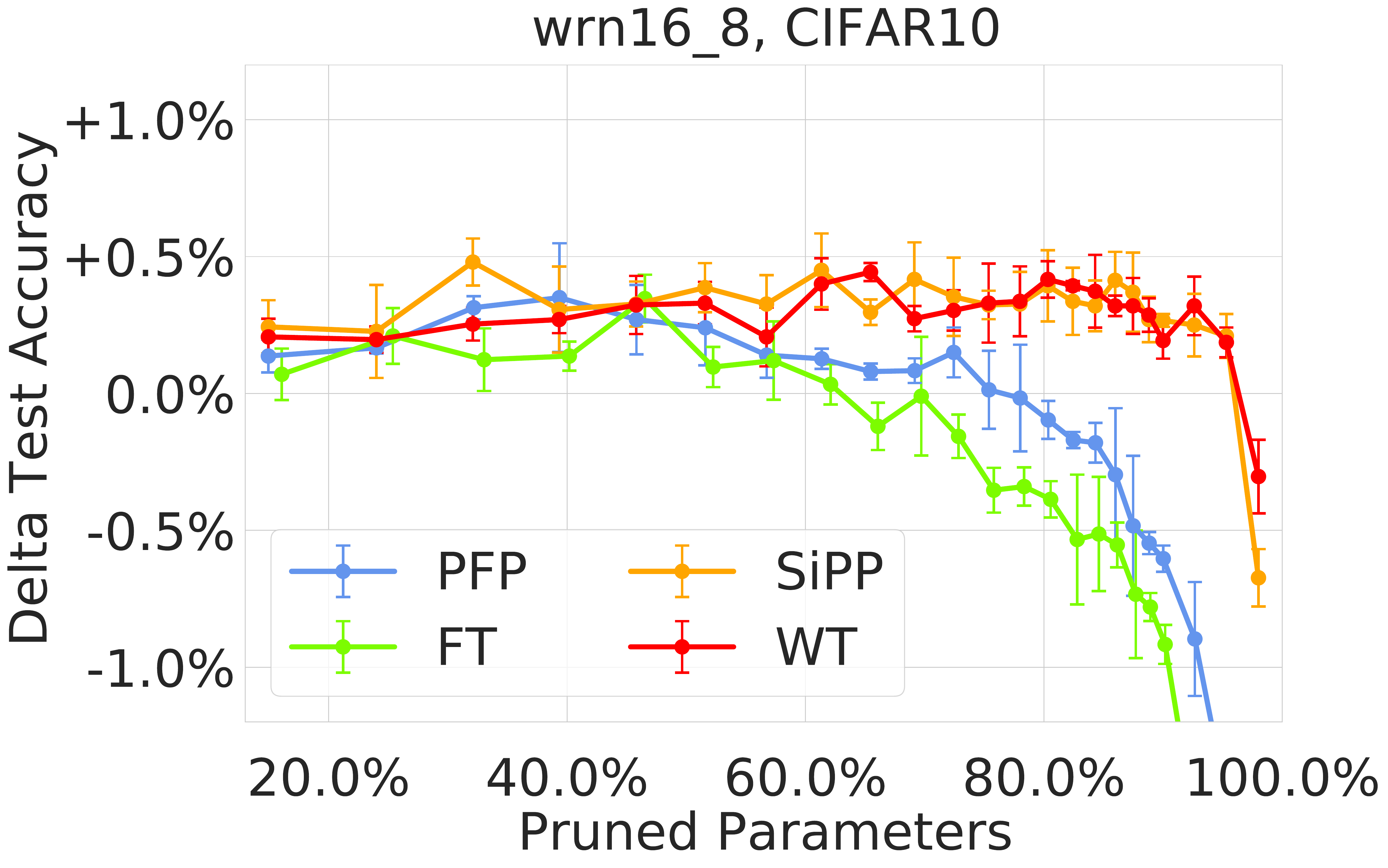}
    \subcaption{WRN16-8}
\end{minipage}%
 
\caption{The difference in test accuracy to the uncompressed network for the generated pruned models trained on CIFAR10 for the evaluated pruning schemes for various target prune ratios.}
\label{fig:cifar_prune_finetune}

\end{figure}

Below we provide the results regarding the achievable test accuracy of pruned networks across multiple target prune ratios. Figure~\ref{fig:cifar_prune_finetune} indices the results for various networks trained on CIFAR10 using an iterative schedule to prune them. In Table~\ref{tab:cifar_results}, we indicate the maximal prune ratio (PR) and the maximal ratio of reduced flops (FR) for which the network achieves commensurate accuracy (within 0.5\% of the original accuracy). We note that the performance of our pruned networks is competitive with state-of-the-art pruning results~\cite{sipp2019, liebenwein2020provable, renda2020comparing, Han15}. For ResNet20 for example, we are able to prune the network to 85\% sparsity while maintaining the original test error (-0.02\% test error), see Table~\ref{tab:cifar_results}.

\begin{table}
    \begingroup
    % \small
    \setlength{\tabcolsep}{3.05pt} % Default value: 6pt
    \renewcommand{\arraystretch}{1.3} % Default value: 1
    \centering
    \begin{tabular}{c|c|ccc|ccc|ccc|ccc|}
        \multirow{2}{*}{Model}
        & Orig.
        & \multicolumn{3}{c|}{WT} 
        & \multicolumn{3}{c|}{SiPP} 
        & \multicolumn{3}{c|}{FT} 
        & \multicolumn{3}{c|}{PFP}  \\%
        & Err.
        & Err. & PR & FR  
        & Err. & PR & FR 
        & Err. & PR & FR 
        & Err. & PR & FR \\ \hline
        Resnet20
        & 8.60
        & -0.02 & \textbf{84.92} & 81.04
        & -0.08 & \textbf{84.92} & 78.78
        & -0.33 & 52.06 & 24.98
        & -0.10 & 44.9 & \textbf{31.27} \\
        Resnet56
        & 7.19
        & -0.53 & 92.91 & 94.31
        & -0.30 & \textbf{93.30} & 93.90
        & -0.38 & 82.71 & 65.50
        & -0.11 & 84.31 & \textbf{73.65} \\
        Resnet110
        & 6.73
        & -0.10 & \textbf{95.76} & 96.73
        & -0.42 & 95.36 & 95.88
        & -0.25 & 86.71 & 72.07
        & -0.25 & 90.27 & \textbf{82.42} \\
        VGG16
        & 7.19 
        & -1.01 & 97.87 & 91.24
        & -0.82 & \textbf{98.00} & 88.45
        & -0.38 & 61.39 & 56.17
        & -0.15 & 90.30 & \textbf{72.03} \\
        DenseNet22
        & 10.10 
        & -0.09 & 71.38 & 76.81
        & -0.20 & \textbf{73.16} & 76.60
        & +0.21 & 43.55 & 42.95
        & -0.04 & 46.18 & \textbf{51.86} \\
        WRN16-8
        & 4.81
        & -0.18 & 95.22 & 92.89
        & -0.21 & \textbf{95.30} & 92.03
        & +0.13 & 76.76 & 71.03
        & -0.08 & 78.79 & \textbf{74.51} \\
    \end{tabular}
    % \vspace{3px}
    \caption{Overview of the pruning performance of each algorithm for various CNN architectures evaluated on the CIFAR data set. For each algorithm and network architecture, the table reports the prune ratio (PR, \%) and the ratio of flop reduction (FR, \%) of pruned models when achieving test accuracy within $\delta=0.5\%$ of the original network's test accuracy (or the closest result when the desired test accuracy was not achieved for the range of tested PRs). The top values for the error and either PR (for weight-based) or FR (for filter-based algorithms) are bolded, respectively.}
    \label{tab:cifar_results}
    \endgroup
\end{table}

% \begin{wraptable}{r}{9cm}
\begin{table}
    % \vspace{-3ex}
    % \small
    \centering
    \begin{tabular}{cr|c}
        &  & ResNet18/101 \\ \hline
        \multirow{10}{*}{Train} 
        & top-1 test error    & 30.26/22.63 \\
        & top-5 test error    & 10.93/6.45 \\
        & loss          & cross-entropy \\
        & optimizer     & SGD \\
        & epochs        & 90 \\
        & warm-up       & 5 \\
        & batch size    & 256 \\
        & LR            & 0.1 \\
        & LR decay      & 0.1@\{30, 60, 80\} \\
        & momentum      & 0.9 \\
        & Nesterov      & \xmark \\
        & weight decay  & 1.0e-4 \\  \hline
        \multirow{2}{*}{Prune}
        & $\gamma$      & 1.0e-16 \\
        & $\alpha$      & 0.90
    \end{tabular}
    \caption{We report the hyperparameters used during training, pruning, and retraining for various convolutional architectures on ImageNet. LR hereby denotes the learning rate and LR decay denotes the learning rate decay that we deploy after a certain number of epochs.}
    % \vspace{-4ex}
    \label{tab:imagenet_hyperparameters}
\end{table}

\subsection{Experimental Setup on ImageNet}
\label{sec:supp-imagenet-hyperparameters}

The hyperparameters for the ImageNet pruning experiments are summarized in Table~\ref{tab:imagenet_hyperparameters}. We consider pruned convolutional neural networks derived from Resnet18 and Resnet101. As in the case of the CIFAR10 experiments, we re-purpose the training schedule from the original ResNet paper~\cite{he2016deep} for both training and retraining. For multi-gpu training we use the linear scaling rule of~\cite{goyal2017accurate} to scale up the learning rate and we use learning rate warm, where we linearly scale up the learning rate from 0 to the nominal learning rate.

\subsection{Pruning Performance on ImageNet}
\label{sec:supp-imagenet-pruning}

\begin{figure}[htb!]
  \centering
    \begin{minipage}[t]{0.45\textwidth}
    \includegraphics[width=\textwidth]{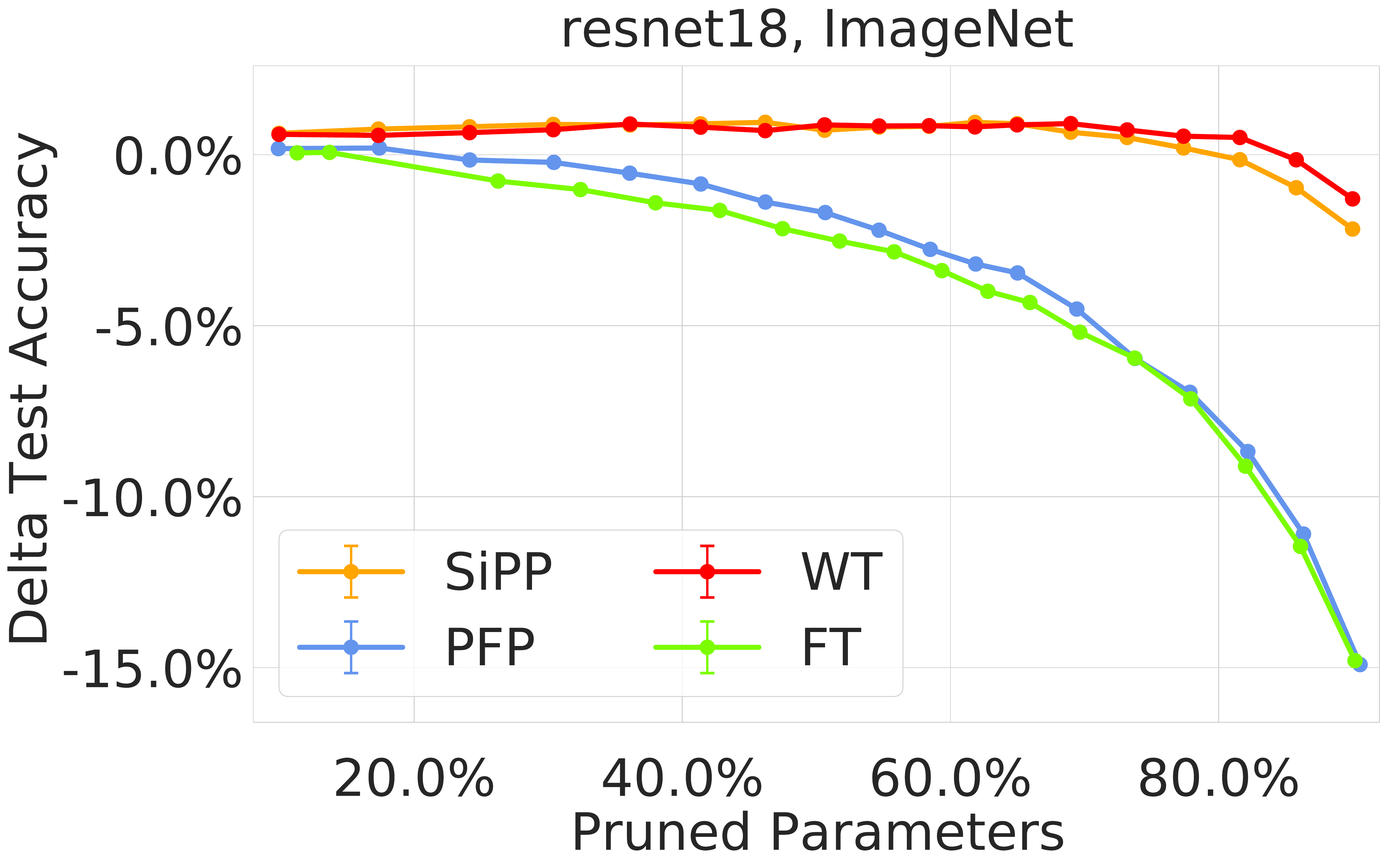}
    \subcaption{Resnet18}
  \end{minipage}%
  \hspace{1ex}
  \begin{minipage}[t]{0.45\textwidth}
    \includegraphics[width=\textwidth]{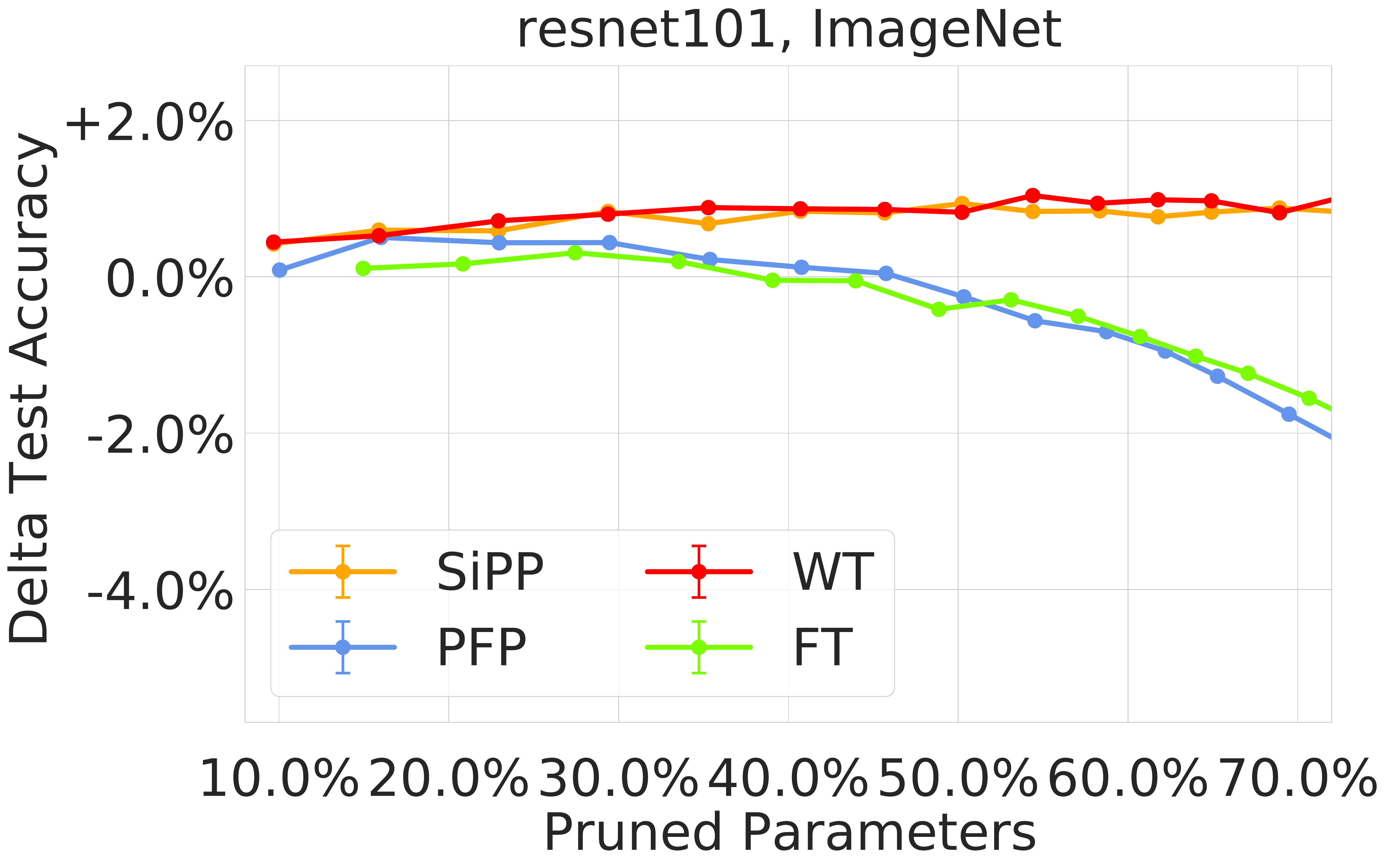}
    \subcaption{Resnet101}
  \end{minipage}%
  \caption{The accuracy of the generated pruned models trained on ImageNet for the evaluated pruning schemes for various target prune ratios.}
  \label{fig:imagenet_prune_finetune}
\end{figure}

The results of our pruning experiments are summarized in Figure~\ref{fig:imagenet_prune_finetune} and Table~\ref{tab:imagenet_results}. Given the computationally expensive nature of ImageNet experiments, we stopped the experiments once the pruned network did not achieve commensurate accuracy anymore (instead of going to extreme prune ratios where the performance decays further). Specifically, we show the achievable test accuracy on nominal ImageNet data for various target prune ratios in Figure~\ref{fig:imagenet_prune_finetune}. In Table~\ref{tab:imagenet_results} we additionally report the maximal prune ratio (PR) and ratio of removed flops (FR) for which the pruned network achieves commensurate accuracy (i.e. within 0.5\% of the unpruned network's accuracy). Just as in the case of CIFAR10, our results are competitive with those reported in state-of-the-art papers~\cite{Han15, renda2020comparing, liebenwein2020provable}.

\begin{table}[htb!]
    \begingroup
    % \small
    \setlength{\tabcolsep}{3.05pt} % Default value: 6pt
    \renewcommand{\arraystretch}{1.3} % Default value: 1
    \centering
    \begin{tabular}{c|c|ccc|ccc|ccc|ccc|}
        \multirow{2}{*}{Model}
        & Orig.
        & \multicolumn{3}{c|}{WT} 
        & \multicolumn{3}{c|}{SiPP} 
        & \multicolumn{3}{c|}{FT} 
        & \multicolumn{3}{c|}{PFP}  \\%
        & Err.
        & Err. & PR & FR  
        & Err. & PR & FR 
        & Err. & PR & FR 
        & Err. & PR & FR \\ \hline
        ResNet18
        & 30.24
        & +0.15 & \textbf{85.79} & 77.14
        & +0.15 & 81.58 & 78.36
        & -0.07 & 13.69 & 10.52
        & +0.22 & 30.42 & \textbf{15.74} \\
        ResNet101
        & 22.63
        & -0.71 & \textbf{81.56} & 83.31
        & -0.59 & \textbf{81.56} & 81.85
        & +0.29 & 53.11 & \textbf{53.28}
        & +0.26 & 50.33 & 44.64 \\
    \end{tabular}
    % \vspace{3px}
    \caption{Overview of the pruning performance of each algorithm for various CNN architectures trained and evaluated on the ImageNet data set. For each algorithm and network architecture, the table reports the prune ratio (PR, \%) and the ratio of flop reduction (FR, \%) of pruned models when achieving test accuracy within $\delta =0.5\%$ of the original network's test accuracy (or the closest result when the desired test accuracy was not achieved for the range of tested PRs). The top values for the error and either PR (for weight-based) or FR (for filter-based algorithms) are bolded, respectively.}
    \label{tab:imagenet_results}
    \endgroup
\end{table}

\subsection{Experimental Setup for Pascal VOC}
\label{sec:supp-voc-hyperparameters}
In addition to CIFAR and ImageNet, we also consider the segmentation task from Pascal VOC 2011~\cite{everingham2015pascal}. We augment the nominal data training data using the extra labels as provided by~\citet{hariharan2011semantic}. As network architecture we consider a DeeplabV3~\cite{chen2017rethinking} with ResNet50 backbone pre-trained on ImageNet. During training we use the following data augmentation pipeline: (1) randomly resize (256x256 to 1024x1024) and crop to 513x513; (2) random horizontal flip; (3) channel-wise normalization. During inference, we resize to 513x513 exactly before the normalization (3) is applied. 
We report both intersection-over-union (IoU) and Top1 test error for each of the pruned and unpruned networks. The experimental hyperparameters are summarized in Table~\ref{tab:voc_hyperparameters}.

\begin{table}
%\small
\centering
\begin{tabular}{cr|c}
& & DeeplabV3-ResNet50 \\ \hline
\multirow{9}{*}{Train} 
&   IoU test error (\%) &  34.78 \\
& top-1 test error (\%) &   7.94 \\
& Loss          & cross-entropy  \\
& Optimizer     & SGD            \\
& Epochs        & 45             \\
& Warm-up       & 0              \\
& Batch size    & 32             \\
& LR            & 0.02           \\
& LR decay      & $\text{(1 - ``step''/``total steps'')} ^ \text{0.9}$ \\
& Momentum      & 0.9            \\
& Nesterov      & \xmark         \\
& Weight decay  & 1.0e-4         \\ \hline
\multirow{2}{*}{Prune}
& $\gamma$ & 1.0e-16             \\
&   $\alpha$      & 0.80         \\ 
\end{tabular}
\caption{
We report the hyperparameters used during training, pruning, and retraining for various architectures on Pascal VOC 2011. LR hereby denotes the learning rate and LR decay denotes the learning rate decay. Note that the learning rate is polynomially decayed after each step.}
\label{tab:voc_hyperparameters}
\end{table}

\subsection{Pruning Performance on VOC}
\label{sec:supp-voc-pruning}

\begin{figure}[htb!]
  \centering
    \begin{minipage}[t]{0.45\textwidth}
    \includegraphics[width=\textwidth]{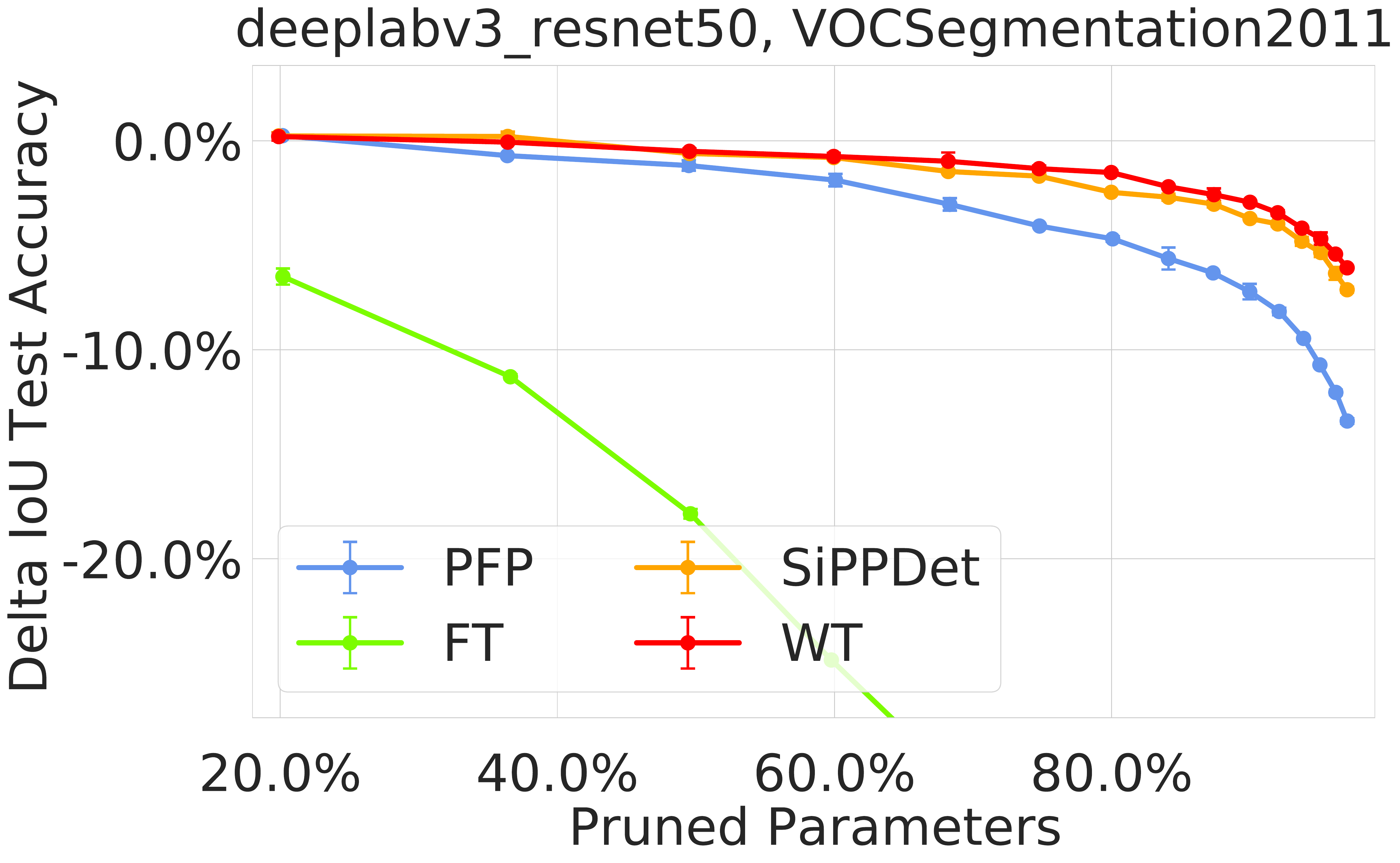}
  \end{minipage}%
  \caption{The accuracy of the generated pruned models trained on VOC for the evaluated pruning schemes and various target prune ratios for a DeeplabV3-ResNet50 architecture.}
  \label{fig:voc_prune_finetune}
\end{figure}

The results of our pruning experiments are summarized in Figure~\ref{fig:voc_prune_finetune} and Table~\ref{tab:voc_results}. Specifically, we show the achievable test accuracy on nominal VOC data for various target prune ratios in Figure~\ref{fig:voc_prune_finetune}. In Table~\ref{tab:voc_results} we report the maximal prune ratio (PR) and ratio of removed flops (FR) for which the pruned network achieves commensurate accuracy (i.e. within 0.5\% of the unpruned network's accuracy).

\begin{table}[htb!]
    \begingroup
    % \small
    \setlength{\tabcolsep}{3.05pt} % Default value: 6pt
    \renewcommand{\arraystretch}{1.3} % Default value: 1
    \centering
    \begin{tabular}{c|c|ccc|ccc|ccc|ccc|}
        \multirow{2}{*}{Model}
        & Orig.
        & \multicolumn{3}{c|}{WT} 
        & \multicolumn{3}{c|}{SiPP} 
        & \multicolumn{3}{c|}{FT} 
        & \multicolumn{3}{c|}{PFP}  \\%
        & Err.
        & Err. & PR & FR  
        & Err. & PR & FR 
        & Err. & PR & FR 
        & Err. & PR & FR \\ \hline
        DeeplabV3
        & 34.78
        & +0.47 & \textbf{58.87} & 58.65
        & +0.29 & 42.98 & 42.70
        & +0.00 &  0.00 &  0.00
        & -0.25 & 20.16 & \textbf{19.14}
    \end{tabular}
    % \vspace{3px}
    \caption{Overview of the pruning performance of each algorithm for DeeplabV3 trained and evaluated on Pascal VOC segmentation data. For each algorithm, the table reports the prune ratio (PR, \%) and the ratio of flop reduction (FR, \%) of pruned models when achieving IoU test accuracy within $\delta =0.5\%$ of the original network's test accuracy (or the closest result when the desired test accuracy was not achieved for the range of tested PRs). The top values for the error and either PR (for weight-based) or FR (for filter-based algorithms) are bolded, respectively.}
    \label{tab:voc_results}
    \endgroup
\end{table}

\section{Additional Results for Function Distance of Pruned Networks}
\label{sec:supp-distance}

In the following, we provide additional empirical evidence for the results presented in Section~\ref{sec:distance} of the main paper. We consider additional CIFAR networks for comparing informative input features and comparing matching predictions when injecting random noise as described in Section~\ref{sec:distance}.

\subsection{Comparison of Informative Features}
\label{sec:supp-distance-informative}

\subsubsection{Informative Features Based on Nominal Test Data}
\begin{figure}[htb]
\vskip 0.2in
\begin{center}
\begin{minipage}{.45\linewidth}
  \includegraphics[width=0.97\columnwidth]{fig/confidence_heatmap_idx-0_bs-frac-0.1.png}%
  \vspace*{-0.07in}
  \subcaption{ResNet20, Pruning by WT}
\end{minipage}%
\begin{minipage}{.45\linewidth}
  \includegraphics[width=0.97\columnwidth]{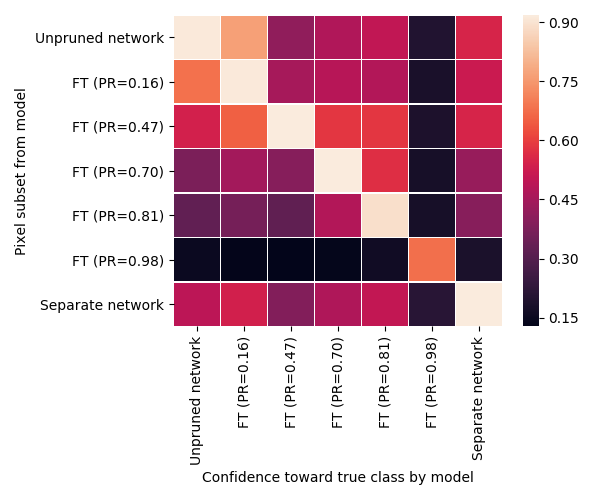}
  \vspace*{-0.07in}
  \subcaption{ResNet20, Pruning by FT}
\end{minipage}
\begin{minipage}{.45\linewidth}
\subfloat[ResNet20, Pruning by SiPP]{
  \includegraphics[width=\columnwidth]{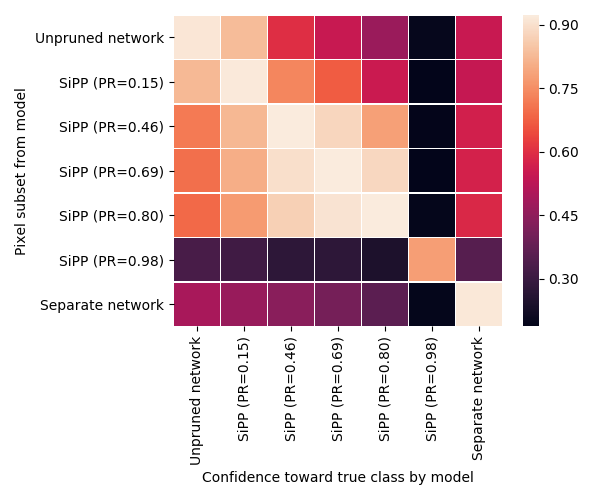}
  \vspace*{-0.07in}
}
\end{minipage}%
\begin{minipage}{.45\linewidth}
\subfloat[ResNet20, Pruning by PFP]{
  \includegraphics[width=\columnwidth]{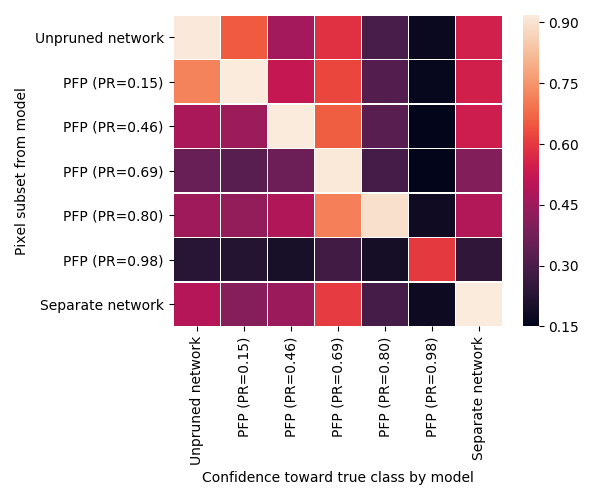}
  \vspace*{-0.07in}
}
\end{minipage}
\caption{Heatmap of confidences on informative pixels from pruned ResNet20 models.  Y-axis is the model used to generate $10\%$ pixel subsets of 2000 sampled CIFAR-10 test images, x-axis describes the models evaluated with each $10\%$ pixel subset, cells indicate mean confidence towards true class of the model from the x-axis on tested data from y-axis.  Pruning by (a) Weight Thresholding (WT),  (b) Filter Thresholding (FT), (c) SiPP, (d) Provable Filter Pruning (PFP).}
\label{fig:supp-resnet20-sis-confidence-heatmaps}
\end{center}
\vskip -0.2in
\end{figure}

\begin{figure}[htb!]
\vskip 0.2in
\begin{center}
\begin{minipage}{.45\linewidth}
\subfloat[VGG16, Pruning by WT]{
  \includegraphics[width=\columnwidth]{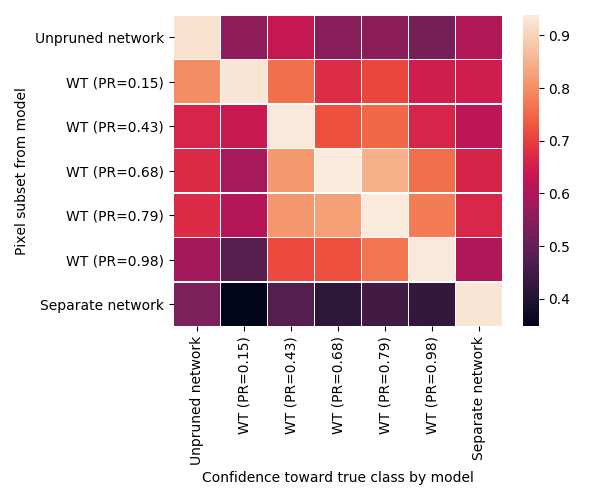}
  \vspace*{-0.07in}
}
\end{minipage}%
\begin{minipage}{.45\linewidth}
\subfloat[VGG16, Pruning by FT]{
  \includegraphics[width=\columnwidth]{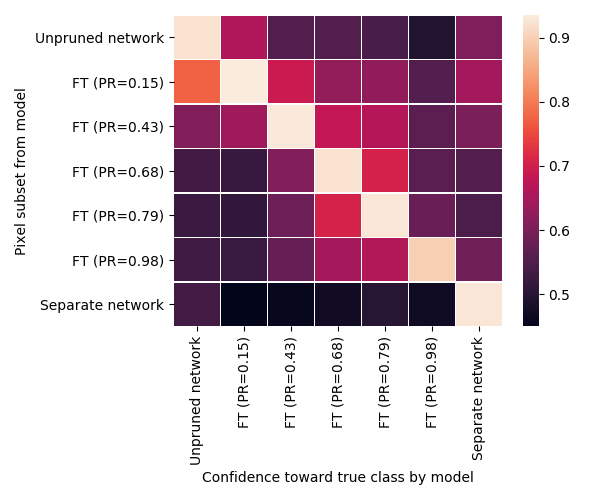}
  \vspace*{-0.07in}
}
\end{minipage}
\begin{minipage}{.45\linewidth}
\subfloat[VGG16, Pruning by SiPP]{
  \includegraphics[width=\columnwidth]{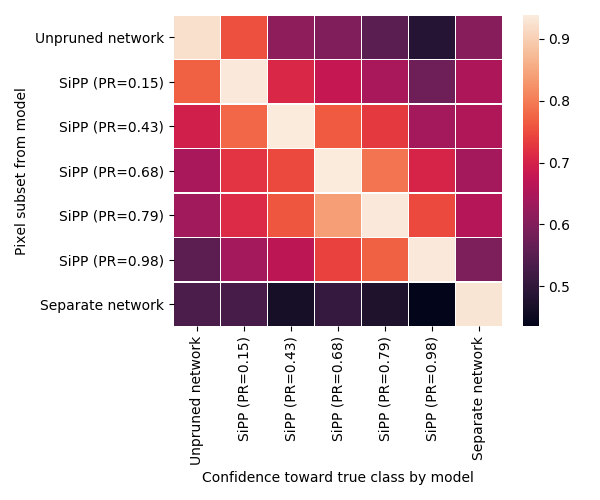}
  \vspace*{-0.07in}
}
\end{minipage}%
\begin{minipage}{.45\linewidth}
\subfloat[VGG16, Pruning by PFP]{
  \includegraphics[width=\columnwidth]{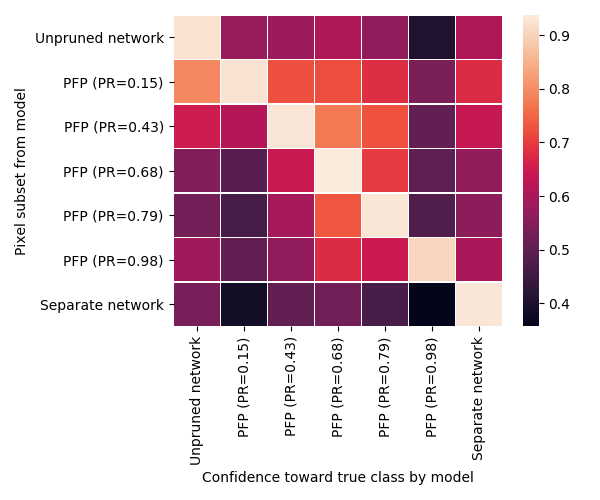}
  \vspace*{-0.07in}
}
\end{minipage}
\caption{Heatmap of confidences on informative pixels from pruned VGG16 models.  Y-axis is the model used to generate $10\%$ pixel subsets of 2000 sampled CIFAR-10 test images, x-axis describes the models evaluated with each $10\%$ pixel subset, cells indicate mean confidence towards true class of the model from the x-axis on tested data from y-axis.  Pruning by (a) Weight Thresholding (WT),  (b) Filter Thresholding (FT), (c) SiPP, (d) Provable Filter Pruning (PFP).}
\label{fig:supp-vgg16-sis-confidence-heatmaps}
\end{center}
\vskip -0.2in
\end{figure}

Figure~\ref{fig:supp-resnet20-sis-confidence-heatmaps} includes results on comparison of informative features for models pruned by other pruning algorithms on ResNet20 (see Section~\ref{sec:distance-results} and Figure~\ref{fig:sis-confidence-heatmap}).
Figure~\ref{fig:supp-vgg16-sis-confidence-heatmaps} shows results for VGG16. The informative features were computed from a random subset of CIFAR10 test data.

\subsubsection{Informative Features Based on Out-of-distribution Test Data}
\begin{figure}[htb]
\vskip 0.2in
\begin{center}
\begin{minipage}{.45\linewidth}
  \includegraphics[width=0.97\columnwidth]{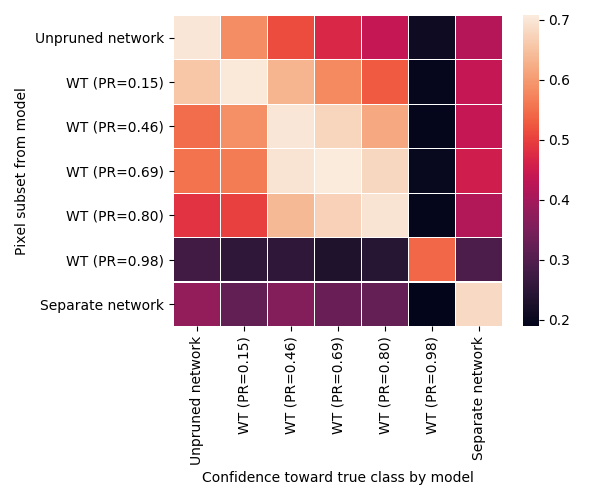}%
  \vspace*{-0.07in}
  \subcaption{ResNet20, Pruning by WT}
\end{minipage}%
\begin{minipage}{.45\linewidth}
  \includegraphics[width=0.97\columnwidth]{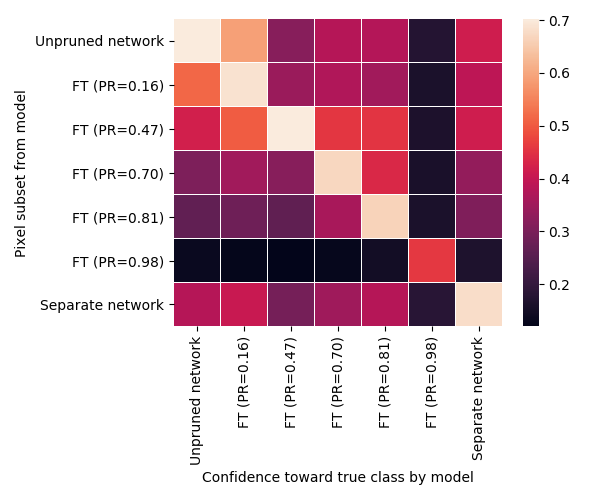}
  \vspace*{-0.07in}
  \subcaption{ResNet20, Pruning by FT}
\end{minipage}
\caption{Heatmap of confidences on informative pixels from pruned ResNet20 models.  Y-axis is the model used to generate $10\%$ pixel subsets of 2000 randomly sampled CIFAR10-C corrupted test images, x-axis describes the models evaluated with each $10\%$ pixel subset, cells indicate mean confidence towards true class of the model from the x-axis on tested data from y-axis.  Pruning by (a) Weight Thresholding (WT),  (b) Filter Thresholding (FT).}
\label{fig:supp-resnet20-sis-confidence-heatmaps-ood}
\end{center}
\vskip -0.2in
\end{figure}

\begin{figure}[htb!]
\vskip 0.2in
\begin{center}
\begin{minipage}{.45\linewidth}
\subfloat[VGG16, Pruning by WT]{
  \includegraphics[width=\columnwidth]{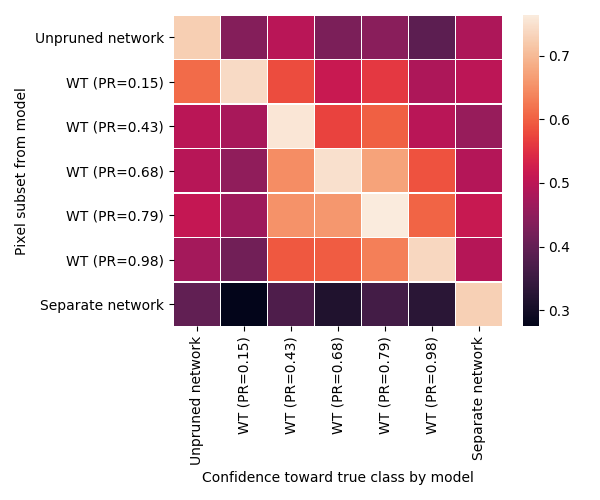}
  \vspace*{-0.07in}
}
\end{minipage}%
\begin{minipage}{.45\linewidth}
\subfloat[VGG16, Pruning by FT]{
  \includegraphics[width=\columnwidth]{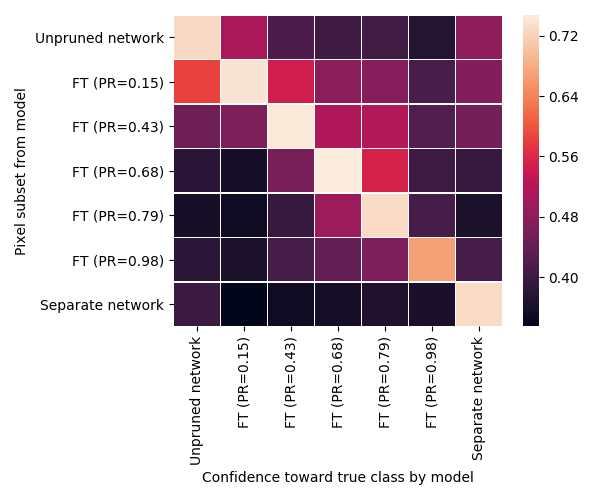}
  \vspace*{-0.07in}
}
\end{minipage}
\caption{Heatmap of confidences on informative pixels from pruned VGG16 models.  Y-axis is the model used to generate $10\%$ pixel subsets of 2000 randomly sampled CIFAR10-C corrupted test images, x-axis describes the models evaluated with each $10\%$ pixel subset, cells indicate mean confidence towards true class of the model from the x-axis on tested data from y-axis.  Pruning by (a) Weight Thresholding (WT),  (b) Filter Thresholding (FT).}
\label{fig:supp-vgg16-sis-confidence-heatmaps-ood}
\end{center}
\vskip -0.2in
\end{figure}

We repeat the experiment with the informative features being computed from a random subset of CIFAR10-C test data (any corruption).
Figure~\ref{fig:supp-resnet20-sis-confidence-heatmaps-ood} includes results on comparison of informative features, c.f. Section~\ref{sec:distance-results}, for models pruned by WT and FT on ResNet20.
Figure~\ref{fig:supp-vgg16-sis-confidence-heatmaps-ood} shows results for VGG16. We note that even when tested with out-of-distribution test data we observe similar trends, i.e., pruned networks in general are more similar in the functional sense to their parent network than a separately trained, unpruned network.

\FloatBarrier
\subsection{Noise Similarities}
\label{sec:supp-distance-noise}
We consider noise properties of networks when feeding perturbed data into the network. In particular, we are interested in understanding the similarities of the output of a pruned network and its unpruned parent as described in Section~\ref{sec:distance} of the main paper. To this end we consider two metrics: (i) percentage of matching predictions (labels) of pruned networks w.r.t. their unpruned parent for various target prune ratios and (ii) the norm-based difference between pruned networks and their unpruned parent. Each result also includes comparisons to a separately trained network of the same architecture with a different random initialization to highlight the functional similarities between unpruned and pruned networks.
Overall, we find that pruned networks functionally approximate their pruned parent more closely than a separately trained network. 
We provide additional empirical evidence for this conclusion below.

\subsubsection{Results for WT and FT on Additional Networks}
We consider the functional similarities between pruned networks and their unpruned parent for the neural network architectures ResNet20, ResNet56, ResNet110, VGG16, DenseNet22, and WideResNet16-8 trained on CIFAR10 as shown in Figures~\ref{fig:resnet20_CIFAR10_WT_FT_noise}, \ref{fig:resnet56_CIFAR10_WT_FT_noise}, \ref{fig:resnet110_CIFAR10_WT_FT_noise}, \ref{fig:vgg16_bn_CIFAR10_WT_FT_noise}, \ref{fig:densenet22_CIFAR10_WT_FT_noise}, and \ref{fig:wrn16_8_CIFAR10_WT_FT_noise}, respectively. All networks shown here were retrained using the same iterative prune schedule. As apparent from the respective figures, the functional similarities are consistent across architectures for the same pruning strategies.

\includenoisefigure[We consider the difference in the output after injecting various amounts of noise into the input, see (a), (b) and (c), (d) for networks weight-pruned with WT and filter-pruned with FT, respectively. The differences between a separately trained network and the unpruned parent is also shown. The plots depict the difference measured as the percentage of matching predictions and as norm-based difference in the output after applying softmax, see (a), (c) and (b), (d), respectively.]{resnet20_CIFAR10}{WT}{FT}{ResNet20}

\includenoisefigure{resnet56_CIFAR10}{WT}{FT}{ResNet56}

\includenoisefigure{resnet110_CIFAR10}{WT}{FT}{ResNet110}

\includenoisefigure{vgg16_bn_CIFAR10}{WT}{FT}{VGG16}

\includenoisefigure{densenet22_CIFAR10}{WT}{FT}{DenseNet22}

\includenoisefigure{wrn16_8_CIFAR10}{WT}{FT}{WRN16-8}

\includenoisefigure{resnet20_CIFAR10}{SiPP}{PFP}{ResNet20}

\FloatBarrier
\subsubsection{Results for Additional Pruning Methods (SiPP and PFP)}
In the following we compare the functional similarities using the alternative weight and filter pruning methods SiPP~\cite{sipp2019} and PFP~\cite{liebenwein2020provable}, respectively. The methods are described in more detail in Section~\ref{sec:methods} of the main paper. Below we present results for ResNet20, ResNet56, ResNet110, VGG16, DenseNet22, and WRN16-8, see Figures~\ref{fig:resnet20_CIFAR10_SiPP_PFP_noise}, \ref{fig:resnet56_CIFAR10_SiPP_PFP_noise}, \ref{fig:resnet110_CIFAR10_SiPP_PFP_noise}, \ref{fig:vgg16_bn_CIFAR10_SiPP_PFP_noise}, \ref{fig:densenet22_CIFAR10_SiPP_PFP_noise},
and~\ref{fig:wrn16_8_CIFAR10_WT_PFP_noise} respectively. We note that the conclusions with regards to the functional similarities remain in essence unaltered for alternative pruning methods. Networks were trained with the same iterative prune-retrain schedule as in the previous subsection.

\includenoisefigure{resnet56_CIFAR10}{SiPP}{PFP}{ResNet56}

\includenoisefigure{resnet110_CIFAR10}{SiPP}{PFP}{ResNet110}

\includenoisefigure{vgg16_bn_CIFAR10}{SiPP}{PFP}{VGG16}

\includenoisefigure{densenet22_CIFAR10}{SiPP}{PFP}{DenseNet22}

\includenoisefigure{wrn16_8_CIFAR10}{WT}{PFP}{WRN16-8}

\FloatBarrier
\section{Additional Results for Prune Potential}
\label{sec:supp-potential}

We present additional experimental evaluation for the results presented in Section~\ref{sec:potential} of the main paper. As in Section~\ref{sec:potential}, we consider the \emph{prune potential} for pruned networks when testing the network under noisy input and under corrupted images. 

\subsection{Additional Results for Prune Potential Based on  Noise}
\label{sec:supp-potential-noise}
Here, we inject randomly sampled, bounded uniform noise into the normalized input and investigate the prune potential under various noise levels. Overall we find that the prune potential significantly decreases as more noise is injected into the input providing further evidence of the results presented in the main paper.

\begin{figure*}[htb]
  \centering
  \begin{minipage}[t]{0.45\textwidth}
    \includegraphics[width=\textwidth]{fig/resnet20_CIFAR10_noise_prune_pot.pdf}
    % \subcaption{ResNet20}
  \end{minipage}
  \begin{minipage}[t]{0.45\textwidth}
    \includegraphics[width=\textwidth]{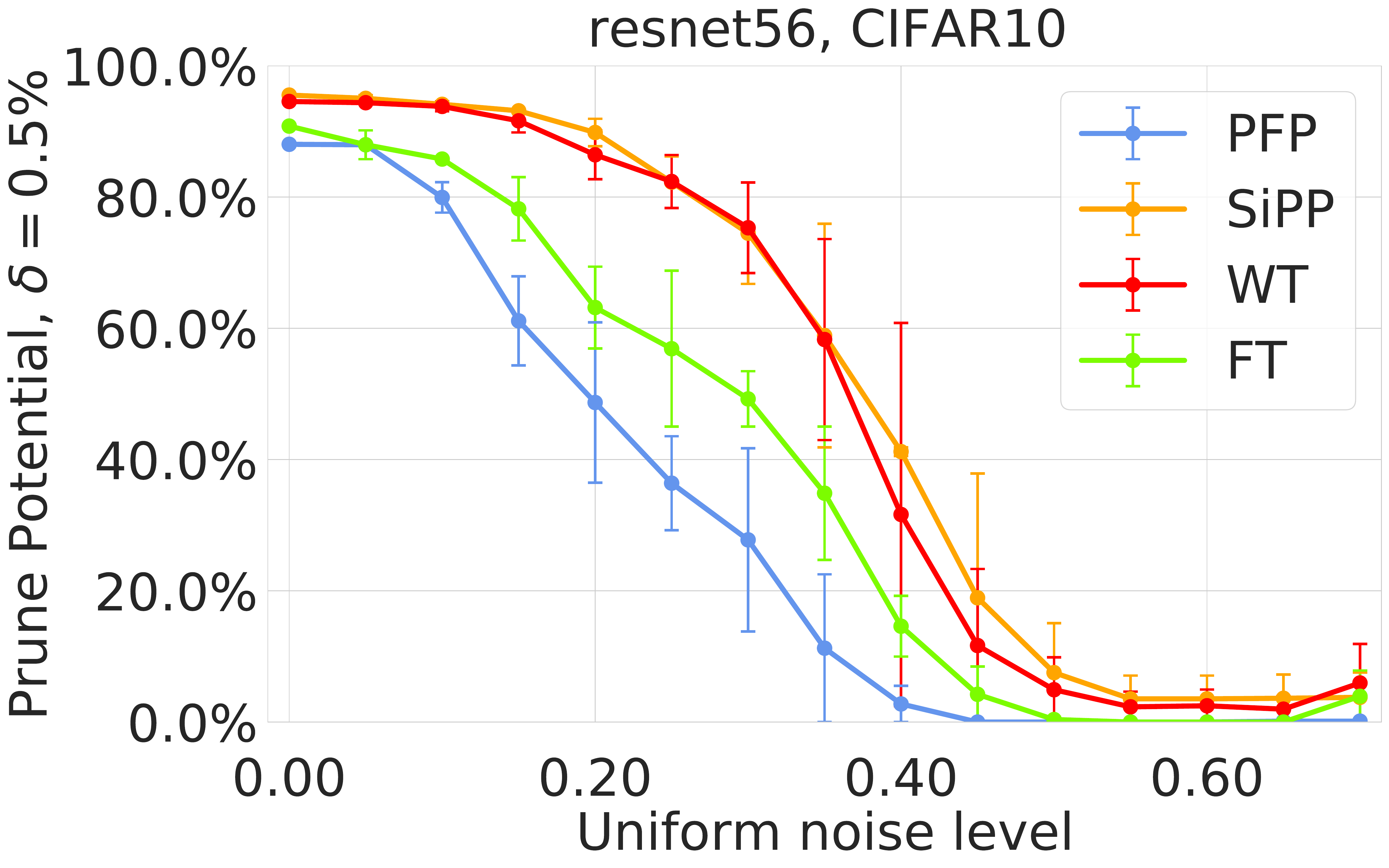}
    % \subcaption{ResNet56}
  \end{minipage}
  \begin{minipage}[t]{0.45\textwidth}
    \includegraphics[width=\textwidth]{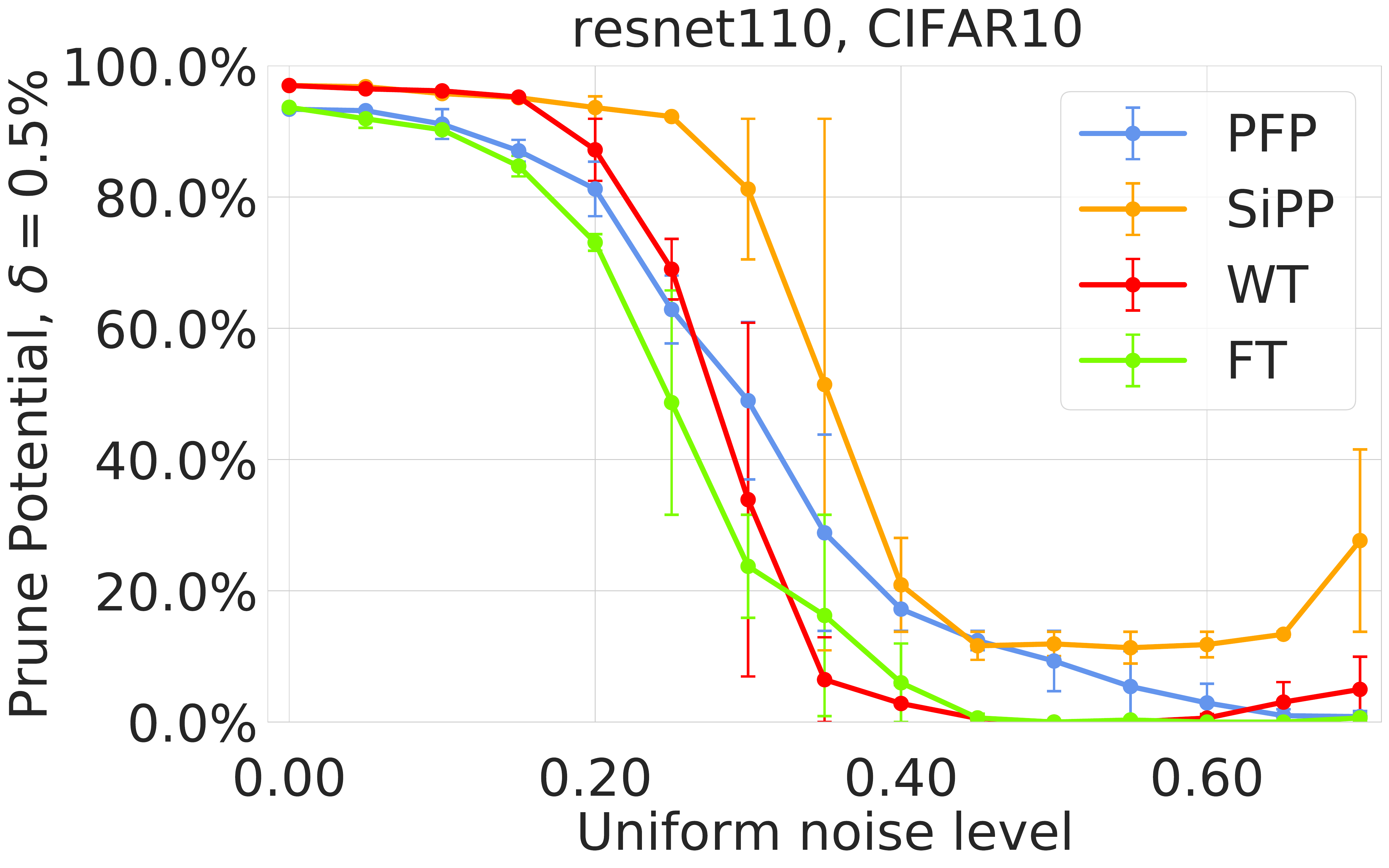}
    % \subcaption{ResNet110}
  \end{minipage}
  \begin{minipage}[t]{0.45\textwidth}
    \includegraphics[width=\textwidth]{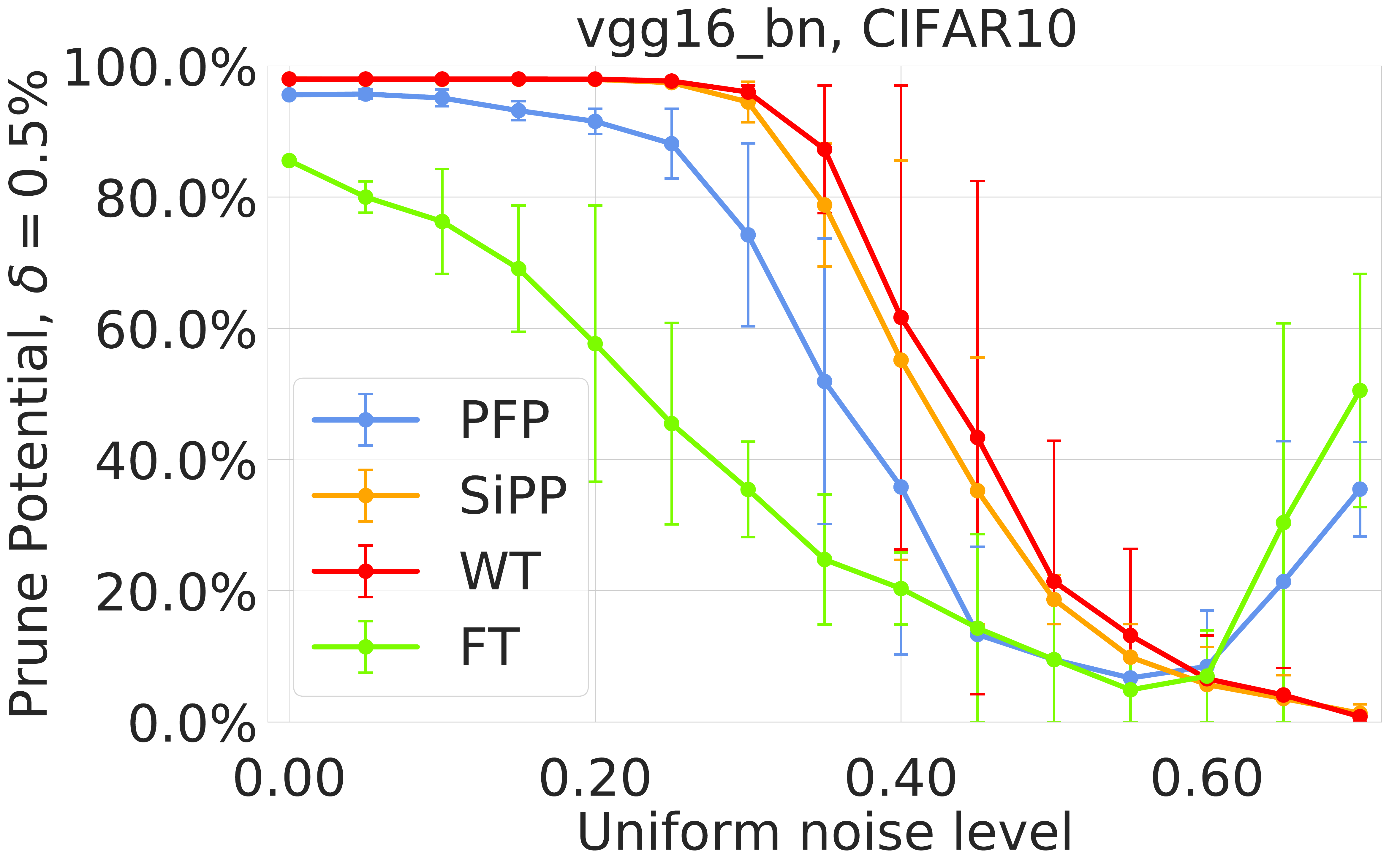}
    % \subcaption{VGG16}
  \end{minipage}
  \begin{minipage}[t]{0.45\textwidth}
    \includegraphics[width=\textwidth]{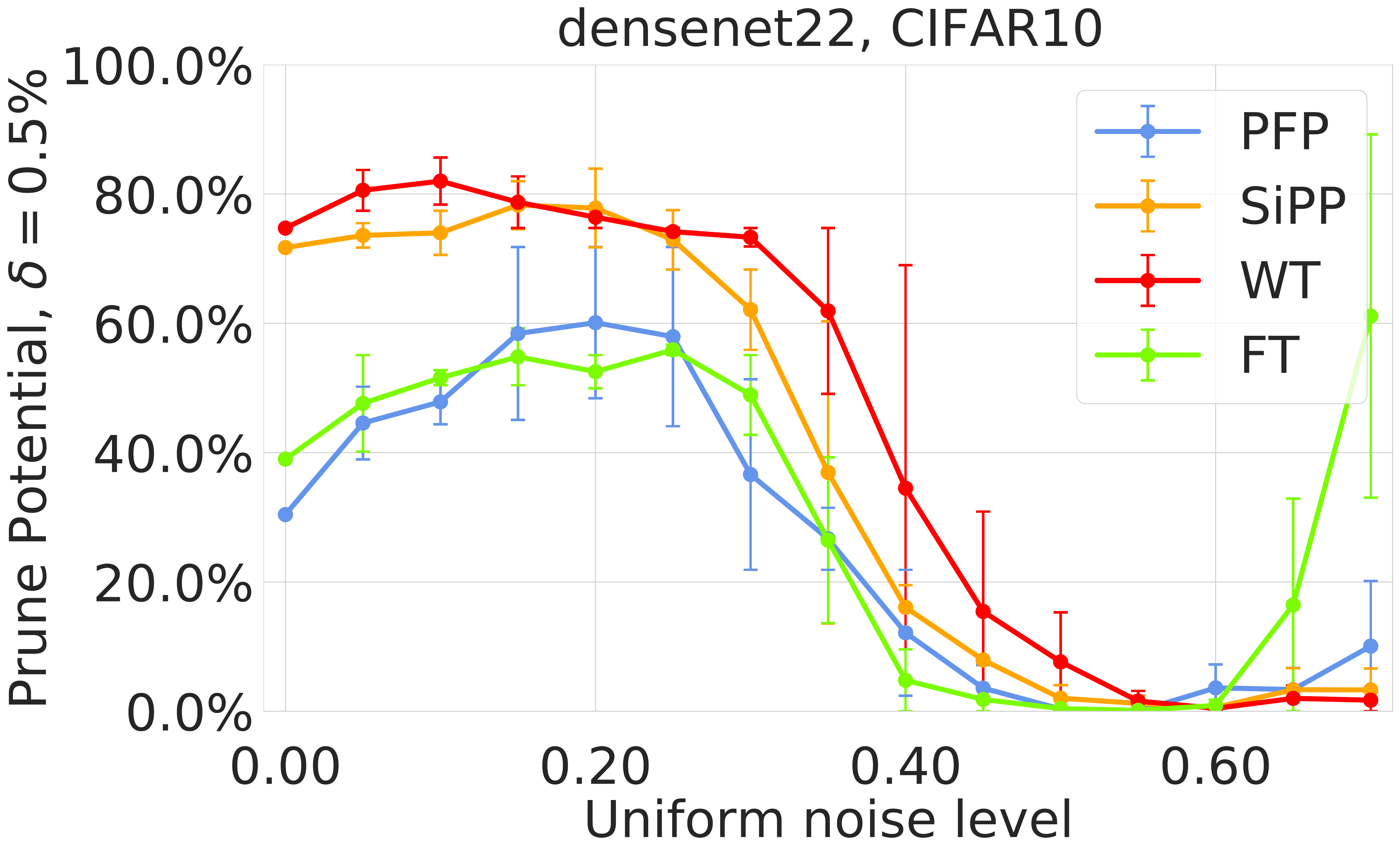}
    % \subcaption{DenseNet22}
  \end{minipage}
  \begin{minipage}[t]{0.45\textwidth}
    \includegraphics[width=\textwidth]{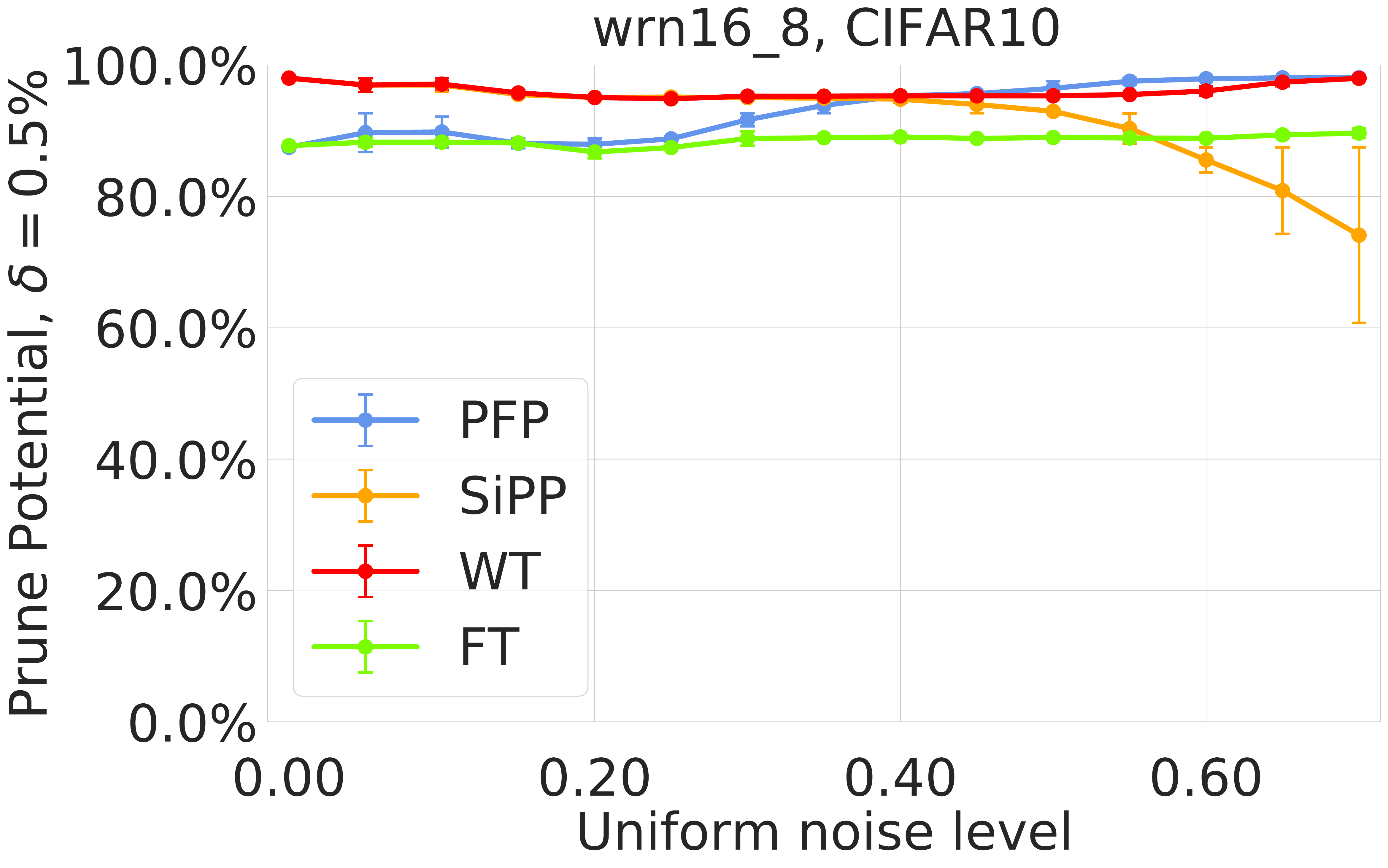}
    % \subcaption{WRN16-8}
    % \label{fig:wrn16_8_CIFAR10_noise_prune_pot}
  \end{minipage}
  \caption{The prune potential (\%) achievable over various levels of noise injected into the input.}
  \label{fig:noise_prune_pot_all}
\end{figure*}

We consider the prune potential for the neural network architectures ResNet20, ResNet56, ResNet110, VGG16, DenseNet22, and WideResNet16-8 trained on CIFAR10 as shown in Figure~\ref{fig:noise_prune_pot_all}. All networks were iteratively pruned and retrained. Overall, we can observe that the prune potential decreases with higher levels of noise and that filter-pruned networks tend to have lower prune potential than weight-pruned networks for any given target prune ratio. 
However, we note that depending on the architecture the specific prune potential may be less affected. Specifically, we note that the prune potential of WideResNets (Figure~\ref{fig:noise_prune_pot_all}, bottom right) seems to be entirely unaffected by noise across all tested prune methods. In contrast to the other networks, WideResNets are wider and less deep, which may provide a possible explanation for the observed behaviors. Due to the wide spatial dimensions of each network, the noise may be better absorbed since it may spread across the width of the layer. Moreover, the reduced depth may help in avoiding positive amplification of the noise as depth increases. From this observation, we may conclude that WideResNets are indeed overparameterized and henceforth the prune potential remains unaffected for slight perturbations in the input.

\subsection{Additional Results for Prune Potential Based on Corruptions}
\label{sec:supp-potential-corruptions}
Following Section~\ref{sec:potential} we present additional results pertaining to the prune potential of networks when generalizing to out-of-distribution test data of various kinds. 
For CIFAR10, we consider the \ood data sets by~\citet{hendrycks2019robustness} (CIFAR10-C) which are publicly available. Additionally, we also compare to CIFAR10.1 by~\citet{recht2018cifar10.1}, which is an alternative in-distribution test data set for CIFAR10.
For ImageNet, we consider the ImageNet versions of the CIFAR10-C data sets, denoted by ImageNet-C~\cite{hendrycks2019robustness}. Additionally, we compare to ObjectNet~\cite{barbu2019objectnet}, an \ood data set that exhibits large variations over the context and the pose of an object instead of image corruptions. For VOC, we consider the same set of corruptions, denoted by VOC-C, based on the generalization of the CIFAR10-C corruptions to any image data set by~\citet{michaelis2019dragon}.
For CIFAR10-C, ImageNet-C, and VOC-C, we evaluate the prune potential for severity level 3 out of 5~\cite{hendrycks2019robustness, michaelis2019dragon}.
In accordance with the results presented in Section~\ref{sec:potential} we find that the prune potential can vary substantially depending on the task. 

We consider the out-of-distribution prune potential for different CIFAR10-C data sets (severity level 3) for the network architectures ResNet20, ResNet56, ResNet110, VGG16, DenseNet22, and WideResNet16-8, see Figures~\ref{fig:resnet20_CIFAR10_generalization_prune_pot}, \ref{fig:resnet56_CIFAR10_generalization_prune_pot}, \ref{fig:resnet110_CIFAR10_generalization_prune_pot}, \ref{fig:vgg16_bn_CIFAR10_generalization_prune_pot}, \ref{fig:densenet22_CIFAR10_generalization_prune_pot}, and~\ref{fig:wrn16_8_CIFAR10_generalization_prune_pot}, respectively. Each network was weight-pruned and filter-pruned with WT and SiPP, and FT and PFP, respectively. 
We note that in general the prune potential varies across prune methods, network architectures, and task. Moreover, some of the networks seem to cope better with out-of-distribution data than other networks (e.g. WideResNet16-8 as seen from Figure~\ref{fig:wrn16_8_CIFAR10_generalization_prune_pot}). However, across all experiments the prune potential varies significantly and it seems difficult to predict clear trends highlighting the sensitivity of the prune potential w.r.t.\ out-of-distribution test data.

\subsection{Additional Results for Prune Potential Based on Corruptions on ImageNet and VOC}
Finally, we consider the prune potential for out-of-distribution test data on a ResNet18 (ImageNet), ResNet101 (ImageNet), and DeeplabV3 (VOC), see Figures~\ref{fig:resnet18_ImageNet_generalization_prune_pot},~\ref{fig:resnet101_ImageNet_generalization_prune_pot}, and~\ref{fig:deeplabv3_resnet50_VOCSegmentation2011_generalization_prune_pot}, respectively. We note that the prune potential in this case is equally sensitive to the test task emphasizing that our observations scale to larger networks and data sets as well instead of being confined to small-scale data sets such as CIFAR10. 
Moreover, considering the expansive nature of ImageNet experiments we did not prune the network to very extreme prune ratios but instead stopped at around 80\%-90\%. In light of these observations, we conjecture that the nominal prune ratio is even higher, which would result in an even larger overall gap in prune potential between in-distribution and out-of-distribution test data.

\includecifarppfigure{resnet20}{ResNet20}

\includecifarppfigure{resnet56}{ResNet56}

\includecifarppfigure{resnet110}{ResNet110}

\includecifarppfigure{vgg16_bn}{VGG16}

\includecifarppfigure{densenet22}{DenseNet22}

\includecifarppfigure{wrn16_8}{WRN16-8}

\includeimagenetppfigure{resnet18}{ResNet18}

\includeimagenetppfigure{resnet101}{ResNet101}

\includevocppfigure{deeplabv3_resnet50}{DeeplabV3}

\begin{figure*}[htb]
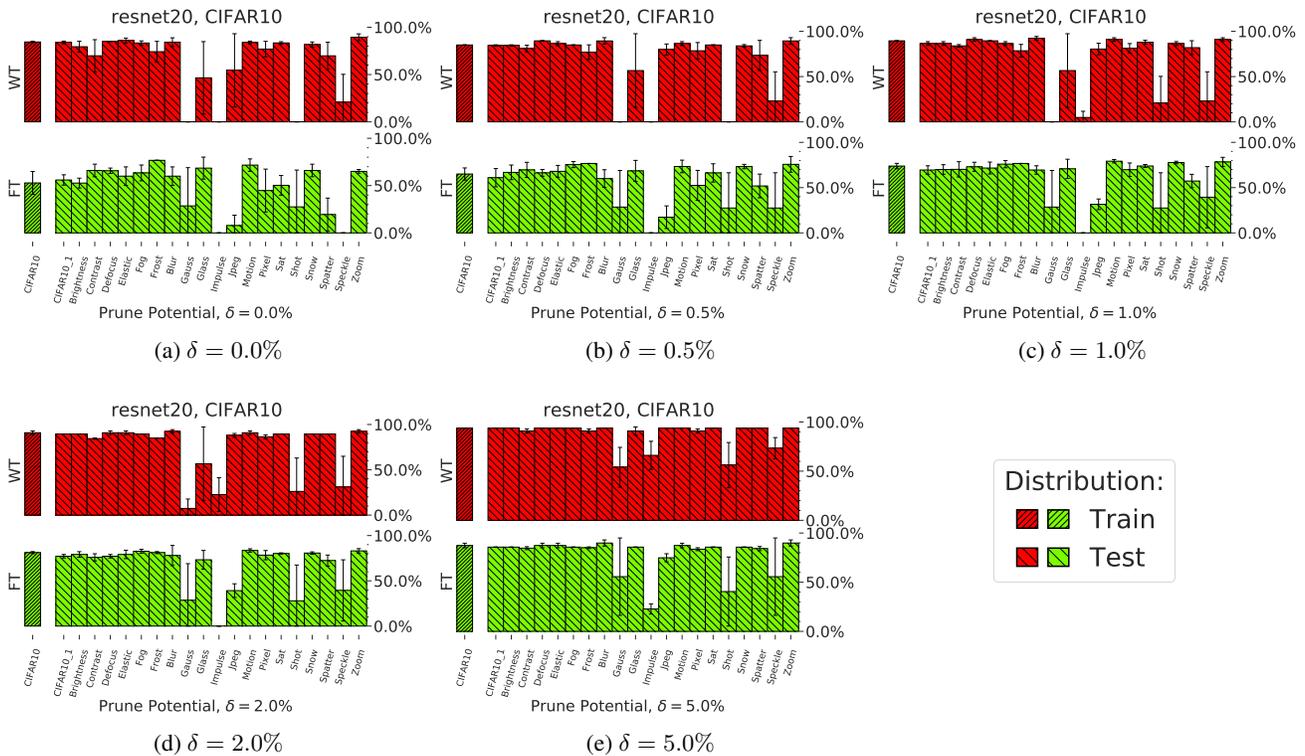

\centering
\begin{minipage}[t]{0.33\textwidth}\vspace{0pt}%
    \includeppgraphics{1.0}{delta0_resnet20_CIFAR10_WT_FT}
    \subcaption{$\delta = 0.0\%$}
\end{minipage}%
\hfill
\begin{minipage}[t]{0.33\textwidth}\vspace{0pt}%
    \includeppgraphics{1.0}{delta1_resnet20_CIFAR10_WT_FT}
    \subcaption{$\delta = 0.5\%$}
\end{minipage}%
\hfill
\begin{minipage}[t]{0.33\textwidth}\vspace{0pt}%
    \includeppgraphics{1.0}{delta2_resnet20_CIFAR10_WT_FT}
    \subcaption{$\delta = 1.0\%$}
\end{minipage}
\begin{minipage}[t]{1.0\textwidth}\vspace{0pt}%
\tiny
\vphantom{3}
\end{minipage}
\begin{minipage}[t]{0.33\textwidth}\vspace{0pt}%
    \includeppgraphics{1.0}{delta3_resnet20_CIFAR10_WT_FT}
    \subcaption{$\delta = 2.0\%$}
\end{minipage}%
\hfill
\begin{minipage}[t]{0.33\textwidth}\vspace{0pt}%
    \includeppgraphics{1.0}{delta4_resnet20_CIFAR10_WT_FT}
    \subcaption{$\delta = 5.0\%$}
\end{minipage}%
\hfill
\begin{minipage}[t]{0.33\textwidth}\vspace{0pt}%
    \centering
    \begin{minipage}[t]{0.45\textwidth}
        \vspace{5ex}
        \includepplegend{delta5_resnet20_CIFAR10_WT_FT}
    \end{minipage}
\end{minipage}
\caption{The prune potential of a ResNet20 trained on CIFAR10 for WT and FT, respectively. In each figure the same experiment is repeated with different values of $\delta$ ranging from $0\%$ to $5\%$.}
\label{fig:supp-potential-delta}
\vspace{-2ex}
\end{figure*}

\FloatBarrier
\subsection{Choice of $\delta$}
\label{sec:supp-potential-delta}
We evaluated the prune potential for a ResNet20 trained on CIFAR10 for a range of possible values for $\delta$ to ensure that our observations hold independent of the specific choice of $\delta$. Recall that $\delta$ denotes the amount of ``slack'' when evaluating the prune potential, i.e., the difference in test accuracy between the pruned and unpruned network for which the pruned network's performance is considered commensurate. Specifically, as shown in Figure~\ref{fig:supp-potential-delta}, we consider a range of $\delta$ between $0\%$ and $5\%$. Naturally, the prune potential is higher overall for larger values of $\delta$. However, we can see that our main observations essentially remain unchanged, that is the prune potential still significantly varies across different tasks and distributions. Overall, these results confirm our previous findings. 

\subsection{Additional Results for Excess Error Based on Corruptions}
\label{sec:supp-potential-excess}
Recall that the excess error is defined as the additional error incurred on the test distribution compared to the error on the train distribution. The difference in excess error between pruned and unpruned networks thus quantifies the \emph{additional error} incurred by the pruned network on the test distribution \emph{on top of} the error increase incurred by the pruned network compared to the unpruned network according to the prune-accuracy curve on the train distribution.

We used ordinary least squares (linear regression) to compute the prediction of the relationship between prune ratio and difference in excess error. The $y$-intercept is set to $0$ since by the definition the difference in excess error is $0\%$ for a prune ratio of $0\%$. The shaded regions describe the $95\%$ confidence intervals, which were computed based on bootstrapping.

The results for the CIFAR10 network architectures ResNet20, ResNet56, ResNet110, VGG16, DenseNet22, and WRN16-8 are shown in Figures~\ref{fig:resnet20_CIFAR10_excess_error},~\ref{fig:resnet56_CIFAR10_excess_error},~\ref{fig:resnet110_CIFAR10_excess_error},~\ref{fig:vgg16_bn_CIFAR10_excess_error},~\ref{fig:densenet22_CIFAR10_excess_error}, and~\ref{fig:wrn16_8_CIFAR10_excess_error}, respectively.
The results for ImageNet network architectures ResNet18 and ResNet101 are shown in Figures~\ref{fig:resnet18_ImageNet_excess_error} and~\ref{fig:resnet101_ImageNet_excess_error}, 
respectively. 
The results for the VOC network architecture DeeplabV3 is shown in Figure~\ref{fig:deeplabv3_resnet50_VOCSegmentation2011_excess_error}. 
Note that ideally the slope would be zero indicating that the prune-accuracy curve on nominal data is predictive of \ood data. However, as shown most pruned networks exhibit a significant increase in excess error that increases with higher prune ratios. 
These results further corroborate our understanding that networks cannot be pruned to full extent when faced with \ood data. 

Notable exceptions include WRN16-8 (CIFAR10) and ResNet101 (ImageNet) with little correlation between the prune ratio and the difference in excess error indicating that those networks may be genuinely overparameterized in a robust sense confirming our findings from previous sections. 

The negative delta in excess error for FT on DeeplabV3 (Figure~\ref{fig:deeplabv3_resnet50_VOCSegmentation2011_excess_error}), on the other hand, is a spurious consequence of the prune potential of FT for nominal data already being zero rather than a consequence of the amount of overparameterization in the network.

\includecifarexcessfigure{resnet20}{ResNet20}

\includecifarexcessfigure{resnet56}{ResNet56}

\includecifarexcessfigure{resnet110}{ResNet110}

\includecifarexcessfigure{vgg16_bn}{VGG16}

\includecifarexcessfigure{densenet22}{DenseNet22}

\includecifarexcessfigure{wrn16_8}{WRN16-8}

\includeimagenetexcessfigure{resnet18}{ResNet18}

\includeimagenetexcessfigure{resnet101}{ResNet101}

\includevocexcessfigure{deeplabv3_resnet50}{DeeplabV3}

\FloatBarrier
\subsection{Results for Overparameterization}
\label{sec:supp-potential-overparameterization}

We summarize our results pertaining to using the prune potential as a way to gauge the amount of overparameterization. Specifically, for each network and prune method, we evaluate the \emph{average} and \emph{minimum} prune potential for both the train and test distribution. The average and minimum are hereby computed over the corruptions/variations that are included in each distribution. Note that for these experiments the train distribution only contains the nominal data; thus the average and minimum coincides. For the test distribution we take the average and minimum over all the respective corruptions. The mean and standard deviation reported are computed over three repetitions of the same experiment. 

The resulting prune potentials are listed in Tables~\ref{tab:supp-overparameterization-weight} and~\ref{tab:supp-overparameterization-filter} for weight pruning (WT, SiPP) and filter pruning (FT, PFP), respectively. Note that for most networks we can observe around $20\%$ drop in average prune potential between train and test distribution while most networks have $0\%$ (!) minimum prune potential for data from the test distribution. As previously observed some networks may be considered \emph{genuinely} overparameterized in the robust sense including WRN16-8, ResNet101, which manifests itself with a very stable prune potential across both train and test distribution.

\begin{table*}[h!]
\centering
\small
\begin{tabular}{|c||cc|cc||cc|cc|}
\hline
\multirow{3}{*}{Model} 
& \multicolumn{4}{c||}{WT - Prune Potential ($\%$)}
& \multicolumn{4}{c|}{SiPP - Prune Potential ($\%$)}\\ 
\cline{2-9}
& \multicolumn{2}{c|}{Average} 
& \multicolumn{2}{c||}{Minimum}
& \multicolumn{2}{c|}{Average} 
& \multicolumn{2}{c|}{Minimum} \\
& Train Dist. & Test Dist. & Train Dist. & Test Dist.
& Train Dist. & Test Dist. & Train Dist. & Test Dist. \\ 
\hline\hline
ResNet20
& 84.9 $\pm$ 0.0 & \textbf{66.7 $\pm$ 3.3}
& 84.9 $\pm$ 0.0 & \textbf{0.0 $\pm$ 0.0}
& 86.4 $\pm$ 2.2 & \textbf{70.4 $\pm$ 4.3}
& 86.4 $\pm$ 2.2 & \textbf{0.0 $\pm$ 0.0} \\
Resnet56
& 94.6 $\pm$ 0.0 & \textbf{82.3 $\pm$ 4.2}
& 94.6 $\pm$ 0.0 & \textbf{4.6 $\pm$ 6.5}
& 94.5 $\pm$ 0.9 & \textbf{78.5 $\pm$ 1.0}
& 94.5 $\pm$ 0.9 & \textbf{0.0 $\pm$ 0.0} \\
ResNet110
& 96.3 $\pm$ 0.6 & \textbf{77.9 $\pm$ 1.4}
& 96.3 $\pm$ 0.6 & \textbf{0.0 $\pm$ 0.0}
& 96.5 $\pm$ 0.7 & \textbf{78.8 $\pm$ 2.8}
& 96.5 $\pm$ 0.7 & \textbf{0.0 $\pm$ 0.0} \\
VGG16
& 98.0 $\pm$ 0.0 & \textbf{80.9 $\pm$ 2.2}
& 98.0 $\pm$ 0.0 & \textbf{0.0 $\pm$ 0.0}
& 98.0 $\pm$ 0.0 & \textbf{80.7 $\pm$ 1.9}
& 98.0 $\pm$ 0.0 & \textbf{0.0 $\pm$ 0.0} \\
DenseNet22
& 79.8 $\pm$ 1.9 & \textbf{76.0 $\pm$ 6.7}
& 79.8 $\pm$ 1.9 & \textbf{24.9 $\pm$ 35.2}
& 79.8 $\pm$ 1.9 & \textbf{74.1 $\pm$ 9.1}
& 79.8 $\pm$ 1.9 & \textbf{21.5 $\pm$ 30.4} \\
WRN16-8
& 98.0 $\pm$ 0.0 & \textbf{95.7 $\pm$ 0.7}
& 98.0 $\pm$ 0.0 & \textbf{90.0 $\pm$ 2.1}
& 95.3 $\pm$ 0.0 & \textbf{94.1 $\pm$ 0.8}
& 95.3 $\pm$ 0.0 & \textbf{78.1 $\pm$ 15.1} \\
\hline\hline
ResNet18
& 85.8 $\pm$ 0.0 & \textbf{63.6 $\pm$ 0.0}
& 85.8 $\pm$ 0.0 & \textbf{0.0 $\pm$ 0.0}
& 81.6 $\pm$ 0.0 & \textbf{57.8 $\pm$ 0.0}
& 81.6 $\pm$ 0.0 & \textbf{0.0 $\pm$ 0.0} \\
ResNet101
& 81.6 $\pm$ 0.0 & \textbf{76.8 $\pm$ 0.0}
& 81.6 $\pm$ 0.0 & \textbf{0.0 $\pm$ 0.0}
& 81.6 $\pm$ 0.0 & \textbf{70.7 $\pm$ 0.0}
& 81.6 $\pm$ 0.0 & \textbf{0.0 $\pm$ 0.0} \\
\hline\hline
DeeplabV3
& 58.9 $\pm$ 9.3 & \textbf{11.6 $\pm$ 2.7}
& 58.9 $\pm$ 9.3 & \textbf{0.0 $\pm$ 0.0}
& 43.0 $\pm$ 6.6 & \textbf{11.5 $\pm$ 3.1}
& 43.0 $\pm$ 6.6 & \textbf{0.0 $\pm$ 0.0} \\
\hline
\end{tabular}
\caption{The average and minimum prune potential computed on the train and test distribution, respectively, for weight prune methods (WT, SiPP). The train distribution hereby consists of nomimal data, while the test distribution consists of the CIFAR10-C, ImageNet-C, VOC-C corruptions.}
\label{tab:supp-overparameterization-weight}
\end{table*}

\begin{table*}[h!]
\centering
\small
\begin{tabular}{|c||cc|cc||cc|cc|}
\hline
& \multicolumn{4}{c||}{FT - Prune Potential ($\%$)}
& \multicolumn{4}{c|}{PFP - Prune Potential ($\%$)}\\ \hline 
\multirow{2}{*}{Model} 
& \multicolumn{2}{c|}{Average} 
& \multicolumn{2}{c||}{Minimum}
& \multicolumn{2}{c|}{Average} 
& \multicolumn{2}{c|}{Minimum} \\
& Train Dist. & Test Dist. & Train Dist. & Test Dist.
& Train Dist. & Test Dist. & Train Dist. & Test Dist. \\ 
\hline\hline
ResNet20
& 65.0 $\pm$ 6.7 & \textbf{55.3 $\pm$ 4.8}
& 65.0 $\pm$ 6.7 & \textbf{0.0 $\pm$ 0.0}
& 66.5 $\pm$ 3.0 & \textbf{53.9 $\pm$ 4.4}
& 66.5 $\pm$ 3.0 & \textbf{0.0 $\pm$ 0.0} \\
ResNet56
& 86.6 $\pm$ 1.2 & \textbf{64.8 $\pm$ 3.7}
& 86.6 $\pm$ 1.2 & \textbf{0.0 $\pm$ 0.0}
& 88.1 $\pm$ 0.0 & \textbf{64.9 $\pm$ 4.0}
& 88.1 $\pm$ 0.0 & \textbf{0.0 $\pm$ 0.0} \\
ResNet110
& 88.1 $\pm$ 3.5 & \textbf{68.5 $\pm$ 3.2}
& 88.1 $\pm$ 3.5 & \textbf{0.0 $\pm$ 0.0}
& 92.2 $\pm$ 1.8 & \textbf{71.6 $\pm$ 1.8}
& 92.2 $\pm$ 1.8 & \textbf{0.0 $\pm$ 0.0} \\
VGG16
& 85.4 $\pm$ 2.4 & \textbf{66.3 $\pm$ 0.5}
& 85.4 $\pm$ 2.4 & \textbf{0.0 $\pm$ 0.0}
& 95.0 $\pm$ 0.5 & \textbf{77.9 $\pm$ 2.1}
& 95.0 $\pm$ 0.5 & \textbf{0.0 $\pm$ 0.0} \\
DenseNet22
& \textbf{47.4 $\pm$ 2.2} & 58.2 $\pm$ 5.3
& 47.4 $\pm$ 2.2 & \textbf{9.6 $\pm$ 13.6}
& \textbf{51.8 $\pm$ 5.3} & 59.6 $\pm$ 6.0
& 51.8 $\pm$ 5.3 & \textbf{12.6 $\pm$ 17.8} \\
WRN16-8
& 86.2 $\pm$ 1.3 & \textbf{75.7 $\pm$ 4.1}
& 86.2 $\pm$ 1.3 & \textbf{36.6 $\pm$ 28.6}
& 86.9 $\pm$ 1.9 & \textbf{86.6 $\pm$ 1.0}
& 86.9 $\pm$ 1.9 & \textbf{66.7 $\pm$ 1.7} \\ 
\hline\hline
ResNet18
& 13.7 $\pm$ 0.0 & \textbf{13.5 $\pm$ 0.0}
& 13.7 $\pm$ 0.0 & \textbf{0.0 $\pm$ 0.0}
& 30.4 $\pm$ 0.0 & \textbf{22.5 $\pm$ 0.0}
& 30.4 $\pm$ 0.0 & \textbf{0.0 $\pm$ 0.0} \\
ResNet101
& 53.1 $\pm$ 0.0 & \textbf{33.5 $\pm$ 0.0}
& 53.1 $\pm$ 0.0 & \textbf{0.0 $\pm$ 0.0}
& 50.3 $\pm$ 0.0 & \textbf{43.0 $\pm$ 0.0}
& 50.3 $\pm$ 0.0 & \textbf{0.0 $\pm$ 0.0} \\ 
\hline\hline
DeeplabV3
& \textbf{0.0 $\pm$ 0.0} & 0.5 $\pm$ 0.5
& \textbf{0.0 $\pm$ 0.0} & \textbf{0.0 $\pm$ 0.0}
& 20.2 $\pm$ 0.1 & \textbf{6.8 $\pm$ 2.6}
& 20.2 $\pm$ 0.1 & \textbf{0.0 $\pm$ 0.0} \\
\hline
\end{tabular}
\caption{The average and minimum prune potential computed on the train and test distribution, respectively, for filter prune methods (FT, PFP). The train distribution hereby consists of nomimal data, while the test distribution consists of the CIFAR10-C, ImageNet-C, VOC-C corruptions.}
\label{tab:supp-overparameterization-filter}
\end{table*}

\FloatBarrier
\section{Additional Details for Prune Potential with robust training}
\label{sec:supp-robustness}
In this section, we consider whether including additional data augmentation techniques derived from the corruptions of CIFAR10-C can boost and/or stabilize the prune potential of a network. 
Specifically, we incorporate a subset of the corruptions into the training pipeline to train and retrain the pruned network in a robust manner.
Below, we list details pertaining to the experimental setup as well as report the results on the conducted experiments. 

\subsection{Experimental Setup and Prune Results}

\begin{figure}[H]
\centering
\begin{minipage}[t]{0.3\textwidth}
    \includegraphics[width=\textwidth]{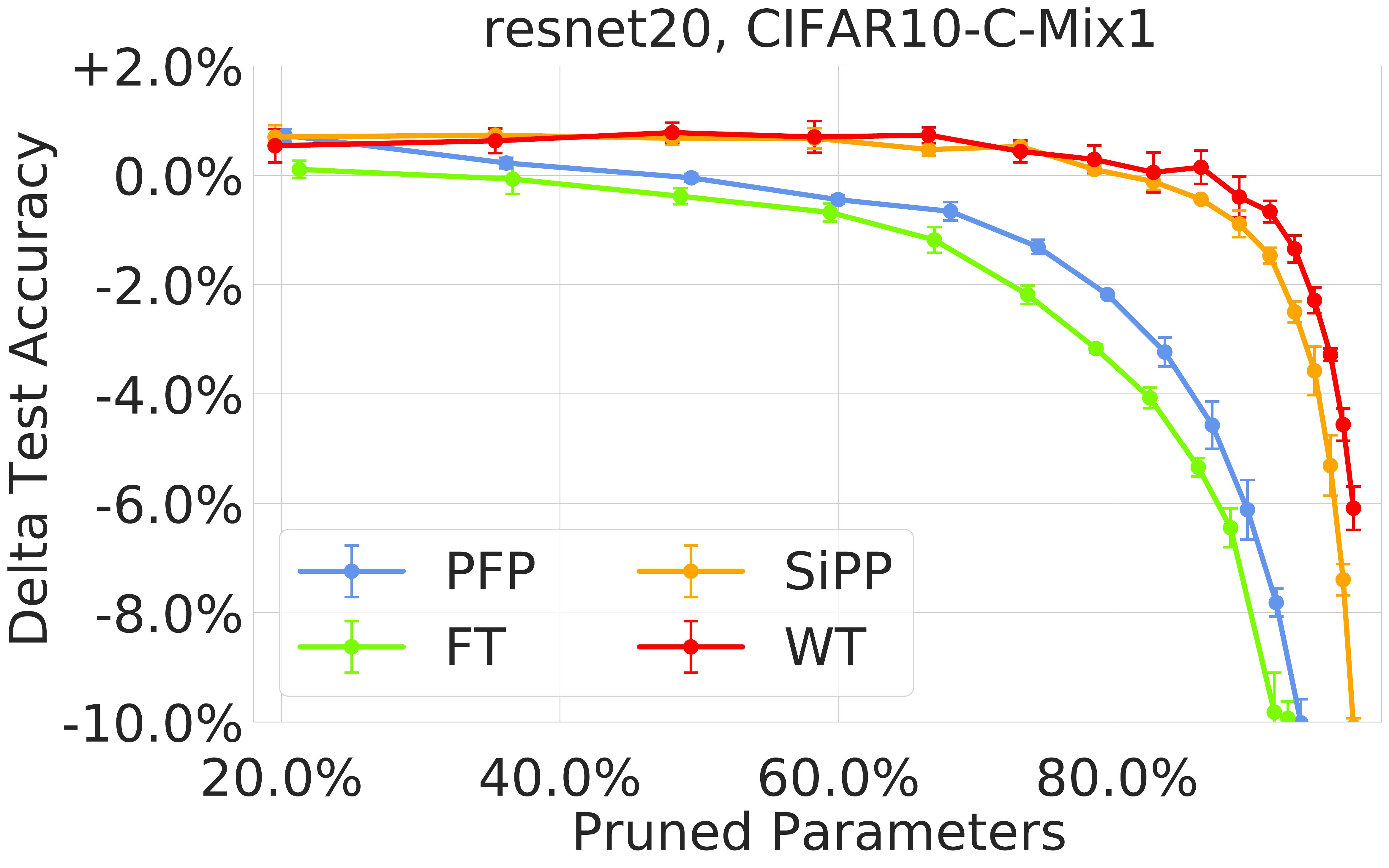}
    \subcaption{Resnet20}
\end{minipage}%
\hfill
\begin{minipage}[t]{0.3\textwidth}
    \includegraphics[width=\textwidth]{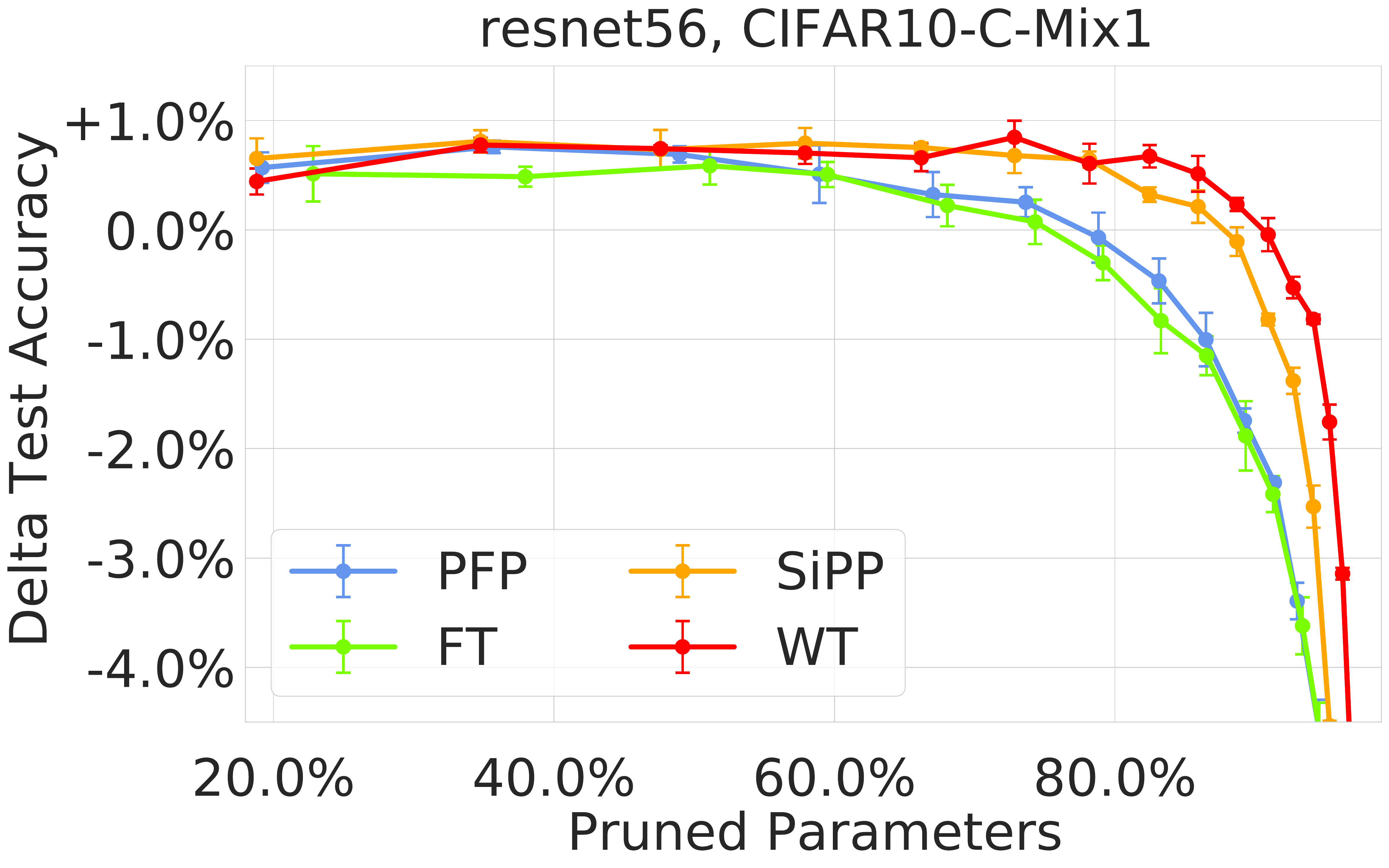}
    \subcaption{Resnet56}
\end{minipage}%
\hfill
\begin{minipage}[t]{0.3\textwidth}
    \includegraphics[width=\textwidth]{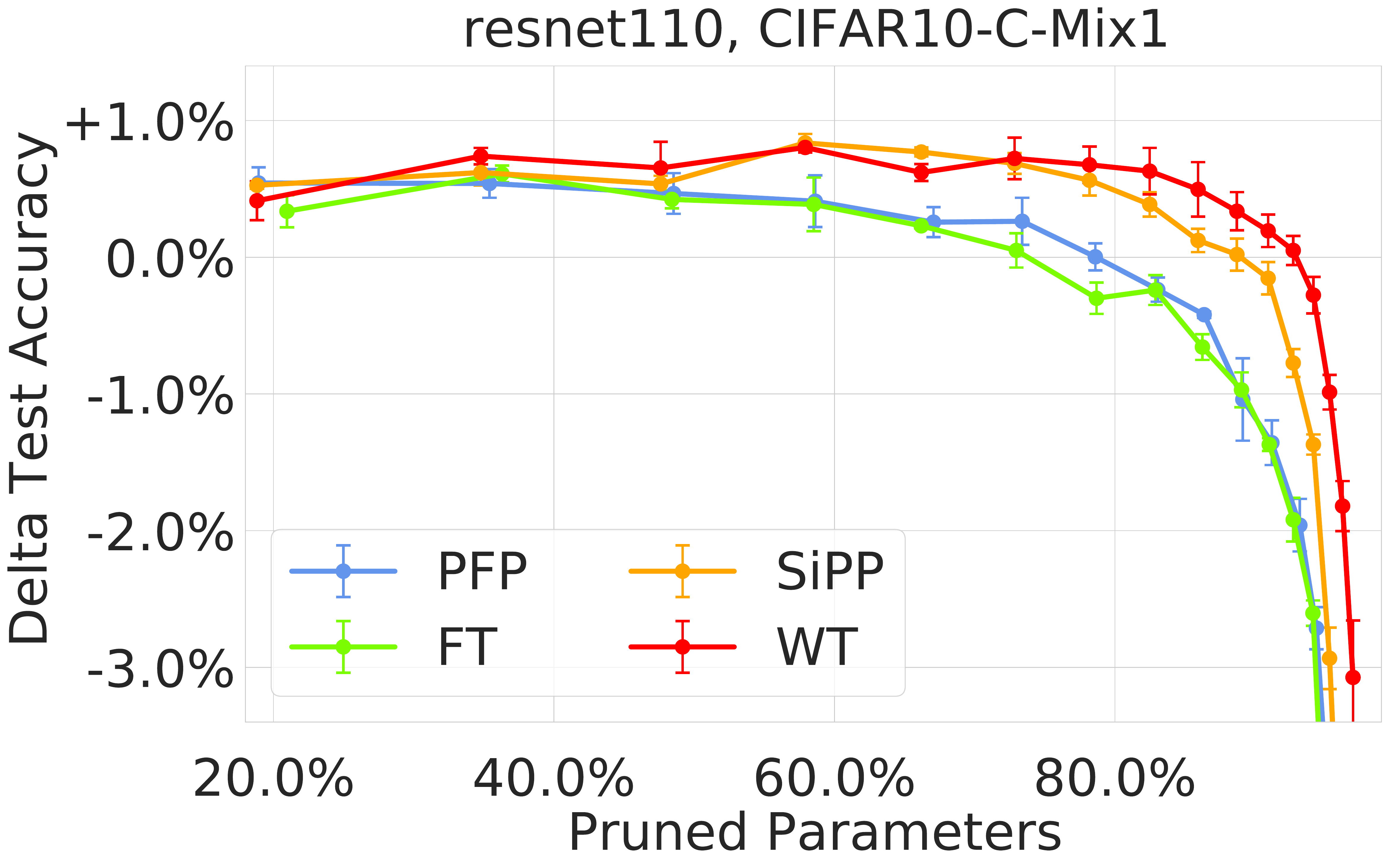}
    \subcaption{Resnet110}
\end{minipage}%
\vspace{2ex}
\begin{minipage}[t]{0.3\textwidth}
    \includegraphics[width=\textwidth]{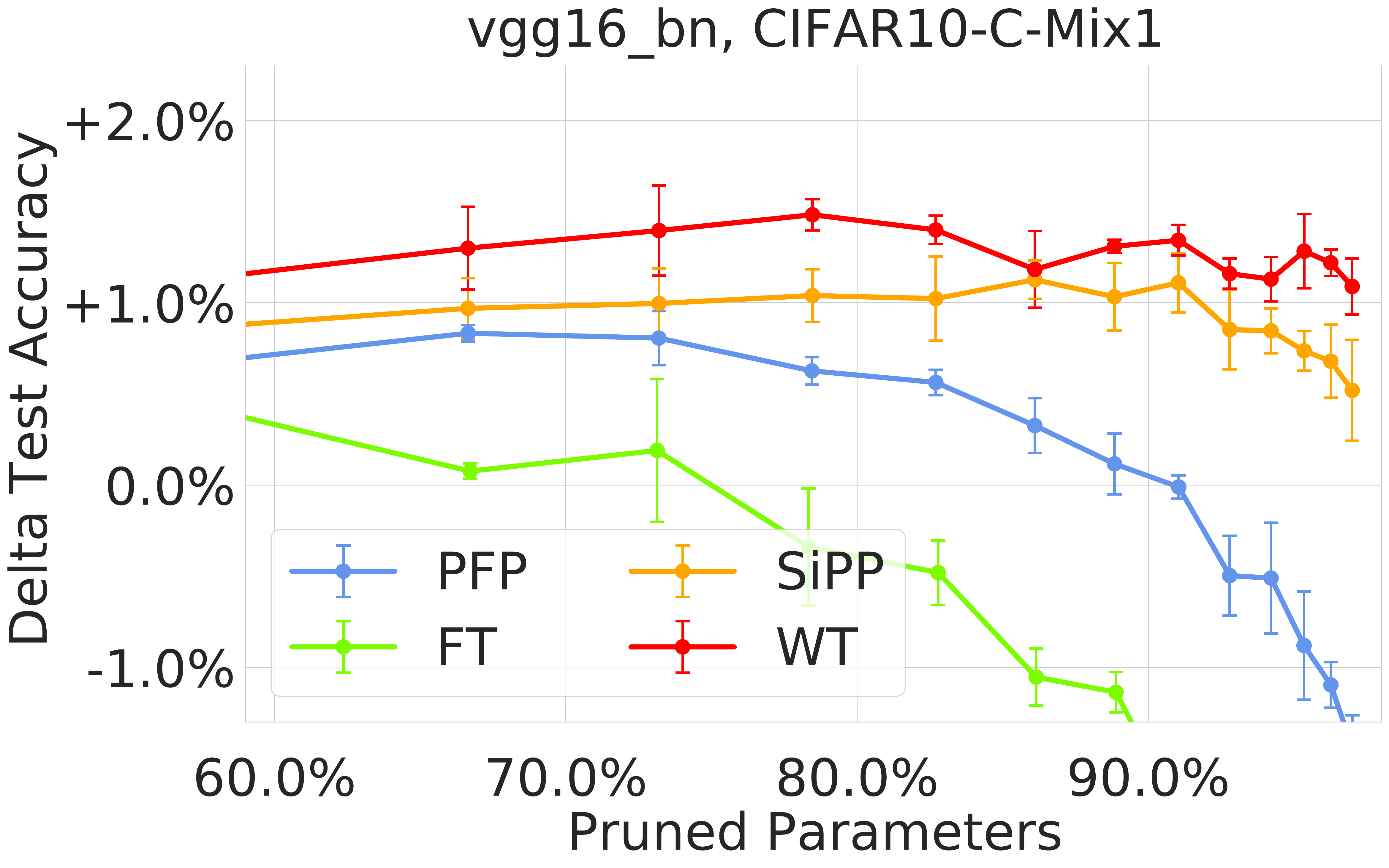}
    \subcaption{VGG16}
\end{minipage}%
\hfill
\begin{minipage}[t]{0.3\textwidth}
    \includegraphics[width=\textwidth]{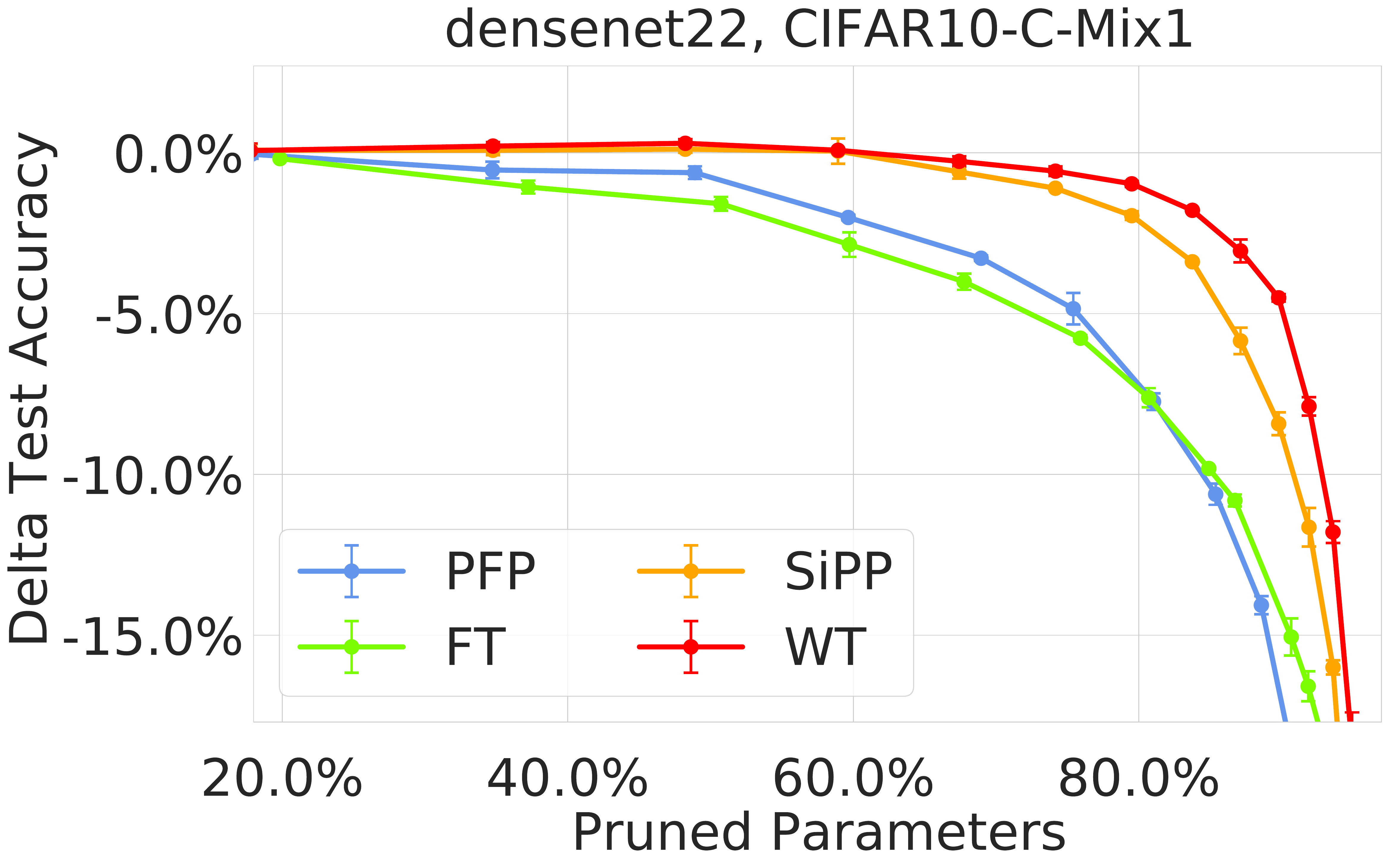}
    \subcaption{DenseNet22}
\end{minipage}%
\hfill
\begin{minipage}[t]{0.3\textwidth}
    \includegraphics[width=\textwidth]{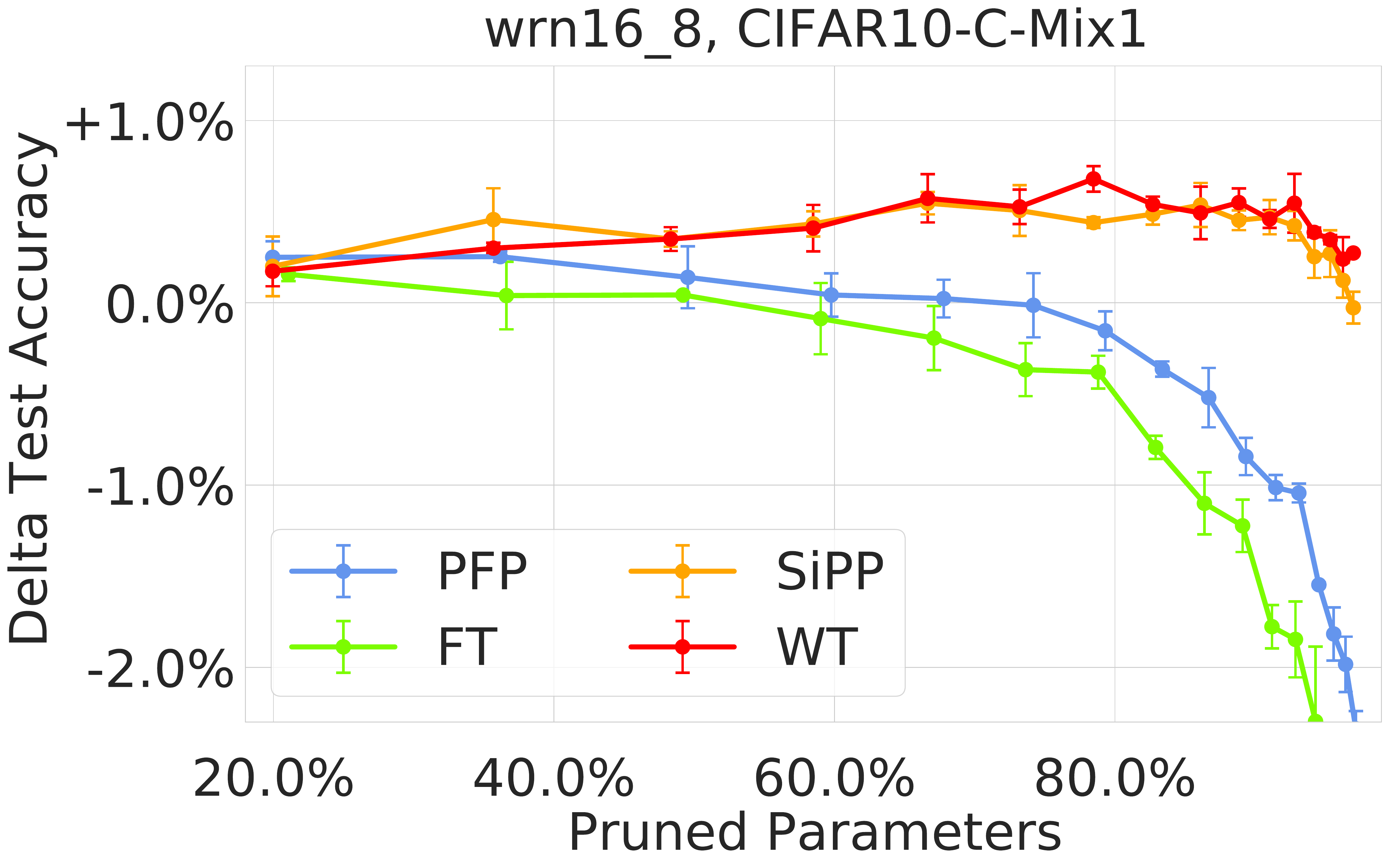}
    \subcaption{WRN16-8}
\end{minipage}%
\caption{The difference in test accuracy (nominal CIFAR10) to the uncompressed network.}
\label{fig:supp-robustness-cifar-prune}
\end{figure}

To train, prune, and retrain networks we consider the same prune pipeline and experimental setting as described in Section~\ref{sec:supp-potential-corruptions}. 
In addition, we incorporate a subset of the CIFAR10-C corruptions into the training and retraining pipeline by corrupting the training data with the respective corruption technique. That is, when sampling a batch of training data each training image is corrupted with a CIFAR10-C corruption (or no corruption) uniformly at random.
The subset of corruptions used as part of the train and test distribution are listed in Table~\ref{tab:supp-corruption-list}. Note that the train and test distribution are mutually exclusive, i.e., they do not share any of the corruptions. However, as shown in Table~\ref{tab:supp-corruption-list} each category of corruption is used in both the train and test distribution.
For each corruption, we choose severity level 3 out of 5 just as before.
The nomimal prune-accuracy curves (CIFAR10) for each of the trained and pruned networks are shown in Figure~\ref{fig:supp-robustness-cifar-prune}.

\begin{table}[t]
\centering
\begin{tabular}{r|l|l}
& Train Distribution & Test Distribution  \\ 
\hline Nominal
& CIFAR10 (no corruption) & CIFAR10.1 \\ 
\hline \multirow{1}{*}{Noise}
& Impulse, Shot
& Gauss \\
\hline \multirow{1}{*}{Blur}
& Motion, Zoom 
& Defocus, Glass \\
\hline \multirow{1}{*}{Weather}
& Snow 
& Brightness, Fog, Frost \\
\hline \multirow{1}{*}{Digital}
& Contrast, Elastic, Pixel
& Jpeg
\end{tabular}
\caption{The list of corruptions used for the train and test distribution, respectively, categorized according to type.}
\label{tab:supp-corruption-list}
\end{table}

\subsection{Results for Prune Potential}
Our results for the prune potential of the network architectures ResNet20, ResNet56, ResNet110, VGG16, DenseNet22, and WRN16-8 are shown in Figures~\ref{fig:resnet20_CIFAR10_C_Mix1_generalization_prune_pot}, \ref{fig:resnet56_CIFAR10_C_Mix1_generalization_prune_pot}, \ref{fig:resnet110_CIFAR10_C_Mix1_generalization_prune_pot}, \ref{fig:vgg16_bn_CIFAR10_C_Mix1_generalization_prune_pot}, \ref{fig:densenet22_CIFAR10_C_Mix1_generalization_prune_pot}, and~\ref{fig:wrn16_8_CIFAR10_C_Mix1_generalization_prune_pot}, respectively.
We note that overall the prune potential for corruptions from the train distribution can be well preserved since we already included the respective corruptions during training and we can predict the prune potential accurately. However, we can also observe that the prune potential improves for some of the corruptions that were not included during retraining. 
Despite training in a robust manner, however, the prune potential can still be significantly lower for corruptions from test distribution and/or exhibit high variance (low predictability). 

\includecifarmixppfigure{resnet20}{robustly pruned ResNet20}

\includecifarmixppfigure{resnet56}{robustly pruned ResNet56}

\includecifarmixppfigure{resnet110}{robustly pruned ResNet110}

\includecifarmixppfigure{vgg16_bn}{robustly pruned VGG16}

\includecifarmixppfigure{densenet22}{robustly pruned DenseNet22}

\includecifarmixppfigure{wrn16_8}{robustly pruned WRN16-8}

\FloatBarrier
\subsection{Results for Excess Error}
\label{sec:supp-robustness-excess}

Following the approach described in Section~\ref{sec:supp-potential-excess} we also evaluated the resulting difference in excess error between pruned and unpruned networks. Our results are shown in Figures~\ref{fig:resnet20_CIFAR10_C_Mix1_excess_error},~\ref{fig:resnet56_CIFAR10_C_Mix1_excess_error},~\ref{fig:resnet110_CIFAR10_C_Mix1_excess_error},~\ref{fig:vgg16_bn_CIFAR10_C_Mix1_excess_error},~\ref{fig:densenet22_CIFAR10_C_Mix1_excess_error}, and~\ref{fig:wrn16_8_CIFAR10_C_Mix1_excess_error} for ResNet20, ResNet56, ResNet110, VGG16, DenseNet22, and WRN16-8, respectively, all of which have been trained in a robust manner.
We note that for most networks, except for smaller ones like ResNet20, the correlation between prune ratio and difference in excess error almost disappears. These results encourage the use of robust pruning techniques in order to ensure that pruned networks perform reliably. However, we note that the excess error is computed as an \emph{average} over all corruptions included in the train and test distribution, respectively. Thus, it is not an appropriate measure to estimate whether particular corruptions could still impact the prune potential more significantly than others.

\includecifarmixexcessfigure{resnet20}{robustly pruned ResNet20}

\includecifarmixexcessfigure{resnet56}{robustly pruned ResNet56}

\includecifarmixexcessfigure{resnet110}{robustly pruned ResNet110}

\includecifarmixexcessfigure{vgg16_bn}{robustly pruned VGG16}

\includecifarmixexcessfigure{densenet22}{robustly pruned DenseNet22}

\includecifarmixexcessfigure{wrn16_8}{robustly pruned WRN16-8}

\FloatBarrier
\subsection{Results for Overparameterization}
\label{sec:supp-robustness-overparameterization}

Finally, we report the average and minimum prune potential across all networks and prune methods for the train and test distribution, respectively. As highlighted in Section~\ref{sec:supp-potential-overparameterization}, the prune potential is used to gauge the amount of overparameterization in the network. 

The resulting prune potentials are listed in Tables~\ref{tab:supp-robustness-overparameterization-weight} and~\ref{tab:supp-robustness-overparameterization-filter} for weight prune methods (WT, SiPP) and filter prune methods (FT, PFP), respectively. In contrast to Section~\ref{sec:supp-potential-overparameterization} the minimum and average prune potential on the train distribution differ since here the train distribution contains multiple corruptions. 
We note that with robust training the average prune potential remains almost unaffected by changes in the distribution as also apparent from the results in Section~\ref{sec:supp-robustness-excess}. In addition, for most networks even the minimum prune potential on the test distribution is nonzero. These results further encourage the use of robust training techniques when pruning neural networks.

As elaborated upon in Section~\ref{sec:robustness}, we observe that we can regain much of the prune potential by \emph{explicitly regularizing} the pruned network during retraining. In other words, the amount of overparameterization is not only a function of the data set and network architecture, but of the training procedure as well. 

However, we note that these observations hinge upon the particular choice of the train and test distribution, which share certain commonalities in this case. Potentially, it might be possible to construct test distributions that differ significantly from the train distribution, in which case pruned networks might suffer disproportionally more from the distribution change compared to unpruned networks. These results would then be analogous to the ones without robust training presented in Section~\ref{sec:supp-potential}. 

\begin{table*}[h!]
\centering
\small
\begin{tabular}{|c||cc|cc||cc|cc|}
\hline
\multirow{3}{*}{Model} 
& \multicolumn{4}{c||}{WT - Prune Potential ($\%$)}
& \multicolumn{4}{c|}{SiPP - Prune Potential ($\%$)}\\ 
\cline{2-9}
& \multicolumn{2}{c|}{Average} 
& \multicolumn{2}{c||}{Minimum}
& \multicolumn{2}{c|}{Average} 
& \multicolumn{2}{c|}{Minimum} \\
& Train Dist. & Test Dist. & Train Dist. & Test Dist.
& Train Dist. & Test Dist. & Train Dist. & Test Dist. \\ 
\hline\hline
ResNet20
& 87.5 $\pm$ 1.9 & \textbf{84.7 $\pm$ 0.9}
& 83.5 $\pm$ 3.6 & \textbf{72.6 $\pm$ 4.9}
& 84.2 $\pm$ 1.7 & \textbf{80.3 $\pm$ 1.6}
& 77.4 $\pm$ 6.1 & \textbf{65.9 $\pm$ 6.1} \\
ResNet56
& 91.8 $\pm$ 0.5 & \textbf{91.3 $\pm$ 0.4}
& 90.2 $\pm$ 1.0 & \textbf{87.8 $\pm$ 1.3}
& 88.7 $\pm$ 0.5 & \textbf{87.2 $\pm$ 0.1}
& 87.8 $\pm$ 1.3 & \textbf{81.0 $\pm$ 2.0} \\
ResNet110
& 93.8 $\pm$ 0.3 & \textbf{93.2 $\pm$ 0.4}
& 92.7 $\pm$ 0.0 & \textbf{88.7 $\pm$ 0.0}
& 90.8 $\pm$ 0.1 & \textbf{90.2 $\pm$ 0.1}
& 89.4 $\pm$ 1.0 & \textbf{83.6 $\pm$ 1.6} \\
VGG16
& 97.0 $\pm$ 0.0 & \textbf{97.0 $\pm$ 0.0}
& 97.0 $\pm$ 0.0 & \textbf{96.8 $\pm$ 0.3}
& 96.5 $\pm$ 0.1 & \textbf{95.5 $\pm$ 0.5}
& 94.6 $\pm$ 0.5 & \textbf{88.3 $\pm$ 3.9} \\
DenseNet22
& 73.4 $\pm$ 1.1 & \textbf{72.7 $\pm$ 1.9}
& 67.4 $\pm$ 0.0 & \textbf{51.8 $\pm$ 5.0}
& \textbf{64.7 $\pm$ 1.3} & 64.8 $\pm$ 0.6
& 55.4 $\pm$ 5.0 & \textbf{51.8 $\pm$ 5.0} \\
WRN16-8
& 97.0 $\pm$ 0.0 & \textbf{97.0 $\pm$ 0.0}
& 97.0 $\pm$ 0.0 & \textbf{96.8 $\pm$ 0.3}
& 97.0 $\pm$ 0.0 & \textbf{96.7 $\pm$ 0.5}
& 96.8 $\pm$ 0.3 & \textbf{94.3 $\pm$ 3.8} \\
\hline
\end{tabular}
\caption{The average and minimum prune potential computed on the train and test distribution, respectively, for weight prune methods (WT, SiPP). The train and test distribution hereby each consist of a mutually exclusive subset of corruptions as listed in Table~\ref{tab:supp-corruption-list}.}
\label{tab:supp-robustness-overparameterization-weight}
\end{table*}

\begin{table*}[h!]
\centering
\small
\begin{tabular}{|c||cc|cc||cc|cc|}
\hline
& \multicolumn{4}{c||}{FT - Prune Potential ($\%$)}
& \multicolumn{4}{c|}{PFP - Prune Potential ($\%$)}\\ \hline 
\multirow{2}{*}{Model} 
& \multicolumn{2}{c|}{Average} 
& \multicolumn{2}{c||}{Minimum}
& \multicolumn{2}{c|}{Average} 
& \multicolumn{2}{c|}{Minimum} \\
& Train Dist. & Test Dist. & Train Dist. & Test Dist.
& Train Dist. & Test Dist. & Train Dist. & Test Dist. \\ 
\hline\hline
ResNet20
& 44.0 $\pm$ 6.3 & \textbf{34.4 $\pm$ 6.1}
& 19.3 $\pm$ 15.0 & \textbf{7.1 $\pm$ 10.0}
& 55.2 $\pm$ 6.2 & \textbf{47.7 $\pm$ 6.4}
& 39.0 $\pm$ 16.1 & \textbf{13.3 $\pm$ 9.4} \\
ResNet56
& 77.1 $\pm$ 0.4 & \textbf{77.1 $\pm$ 0.9}
& 72.2 $\pm$ 2.9 & \textbf{67.3 $\pm$ 6.1}
& 80.8 $\pm$ 0.4 & \textbf{76.3 $\pm$ 0.5}
& 78.8 $\pm$ 0.2 & \textbf{58.5 $\pm$ 7.1} \\
ResNet110
& 82.1 $\pm$ 1.8 & \textbf{81.7 $\pm$ 0.8}
& 78.7 $\pm$ 0.0 & \textbf{74.9 $\pm$ 2.7}
& 84.4 $\pm$ 2.1 & \textbf{80.0 $\pm$ 2.2}
& 81.6 $\pm$ 2.1 & \textbf{58.0 $\pm$ 7.7} \\
VGG16
& 48.0 $\pm$ 3.3 & \textbf{41.5 $\pm$ 5.2}
& \textbf{0.0 $\pm$ 0.0} & \textbf{0.0 $\pm$ 0.0}
& 81.9 $\pm$ 1.8 & \textbf{79.8 $\pm$ 2.3}
& \textbf{62.7 $\pm$ 10.5} & 63.9 $\pm$ 3.9 \\
DenseNet22
& 20.3 $\pm$ 0.7 & \textbf{20.0 $\pm$ 5.7}
& 13.2 $\pm$ 9.4 & \textbf{6.6 $\pm$ 9.4}
& 34.2 $\pm$ 8.8 & \textbf{33.1 $\pm$ 6.5}
& 17.5 $\pm$ 14.2 & \textbf{0.0 $\pm$ 0.0} \\
WRN16-8
& 69.7 $\pm$ 4.1 & \textbf{64.8 $\pm$ 2.2}
& 54.2 $\pm$ 12.9 & \textbf{24.4 $\pm$ 17.3}
& 80.7 $\pm$ 0.7 & \textbf{76.9 $\pm$ 2.9}
& 72.1 $\pm$ 3.1 & \textbf{57.2 $\pm$ 14.9} \\
\hline
\end{tabular}
\caption{The average and minimum prune potential computed on the train and test distribution, respectively, for filter prune methods (FT, PFP). The train and test distribution hereby each consist of a mutually exclusive subset of corruptions as listed in Table~\ref{tab:supp-corruption-list}.}
\label{tab:supp-robustness-overparameterization-filter}
\end{table*}

\end{document}